\documentclass[10pt,journal,compsoc]{IEEEtran}
\ifCLASSOPTIONcompsoc
\else
\fi
\ifCLASSINFOpdf
\else
\fi
\usepackage{hyperref}
\usepackage{forest}
\usepackage[noadjust]{cite}
\usepackage{multirow}
\usepackage{amsmath}
\usepackage{tikz}
\usepackage{pifont}% http://ctan.org/pkg/pifont
\usepackage{caption}% http://ctan.org/pkg/caption
\begin{document}

\title{A Survey of GPT-3 Family Large Language Models Including ChatGPT and GPT-4}

\author{Katikapalli Subramanyam Kalyan \\
        Akmmus AI, Trichy, India \\
        \small{\textit{Email: kalyan@akmmusai.pro}, \textit{Website: \url{https://www.akmmusai.pro}}}
\IEEEcompsocitemizethanks{\IEEEcompsocthanksitem Katikapalli Subramanyam Kalyan is with Akmmus AI as NLP Researcher and Founder, Trichy, Tamil Nadu, India, 620015.\protect\\
E-mail: kalyan@akmmusai.pro, Website: \url{https://www.akmmusai.pro/kalyanksnlp}}% <-this % stops an unwanted space
\thanks{Preprint under review - Akmmus AI is an independent NLP-focused AI research lab from Trichy, India.}}

\IEEEtitleabstractindextext{%
\begin{abstract}
Large language models (LLMs) are a special class of pretrained language models obtained by scaling model size, pretraining corpus and computation. LLMs, because of their large size and pretraining on large volumes of text data, exhibit special abilities which allow them to achieve remarkable performances without any task-specific training in many of the natural language processing tasks. The era of LLMs started with OpenAI’s GPT-3 model, and the popularity of LLMs is increasing exponentially after the introduction of models like ChatGPT and GPT4. We refer to GPT-3 and its successor OpenAI models, including ChatGPT and GPT4, as GPT-3 family large language models (GLLMs). With the ever-rising popularity of GLLMs, especially in the research community, there is a strong need for a comprehensive survey which summarizes the recent research progress in multiple dimensions and can guide the research community with insightful future research directions. We start the survey paper with foundation concepts like transformers, transfer learning, self-supervised learning, pretrained language models and large language models. We then present a brief overview of GLLMs and discuss the performances of GLLMs in various downstream tasks, specific domains and multiple languages. We also discuss the data labelling and data augmentation abilities of GLLMs, the robustness of GLLMs, the effectiveness of GLLMs as evaluators, and finally, conclude with multiple insightful future research directions. To summarize, this comprehensive survey paper will serve as a good resource for both academic and industry people to stay updated with the latest research related to GPT-3 family large language models. 

\end{abstract}

\begin{IEEEkeywords}
Large Language Models, GPT-3, ChatGPT, GPT-4, Transformers, Survey.
\end{IEEEkeywords}}

\maketitle

\IEEEdisplaynontitleabstractindextext

% \fi
\IEEEpeerreviewmaketitle

\tableofcontents

\begin{figure*}[h!]
\begin{center}
\includegraphics[width=20cm, height=8cm]{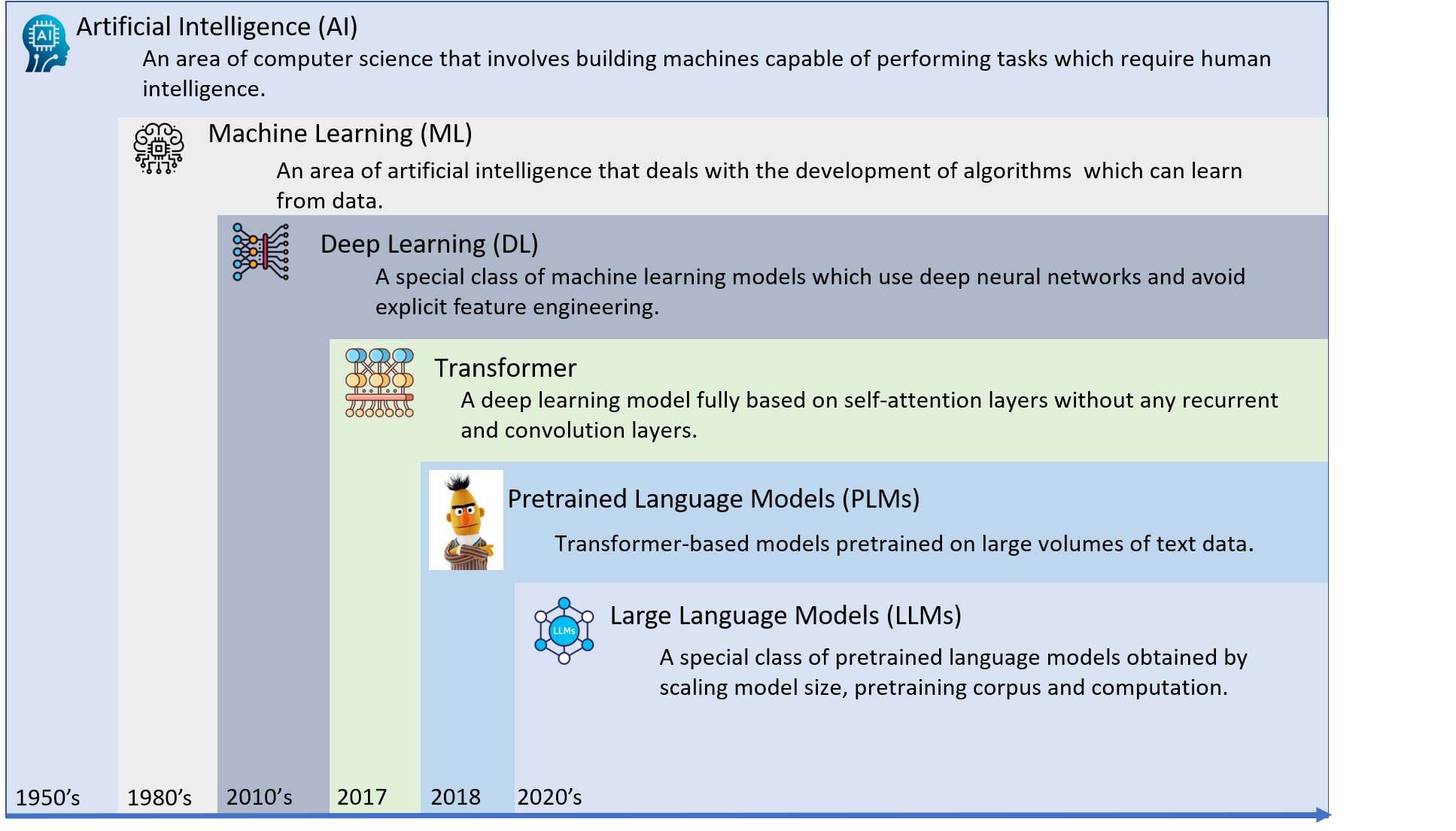}
\caption{\label{picture-1} Evolution of artificial intelligence from machine learning to large language models.} 
\end{center}
\end{figure*}

\section{Introduction}
Large Language Models (LLMs), the recent buzz in Artificial Intelligence, have garnered a lot of attention in both academic and industry circles with their remarkable performances in most of the natural language processing (NLP) tasks. These models are essentially deep learning models, specifically transformer-based, pretrained on large volumes of text data and then aligned to human preferences using meta-training. Pretraining provides universal language knowledge to the model \cite{kalyan2021ammus}, while meta-training aligns the model to act based on the user's intentions. Here user's intention includes both explicit intentions, like following instructions, and implicit intentions, like maintaining truthfulness and avoiding bias, toxicity, or any harmful behaviour \cite{ouyang2022training}. Large language models (LLMs) are a special class of pretrained language models obtained by scaling model size, pretraining corpus and computation. For downstream task usage, pretrained language models leverage supervised learning paradigm, which involves task-specific fine-tuning and hundreds or thousands of labelled instances \cite{kalyan2021ammus, kalyan2022ammu}. LLMs leverage in-context learning (ICL), a new learning paradigm which doesn’t require task-specific fine-tuning and a large number of labelled instances \cite{brown2020language}. LLMs treat any NLP task as a conditional text generation problem and generate the desired text output just by conditioning on the input prompt, which includes task description, test input and optionally, a few examples. Figure \ref{picture-1} shows the evolution of artificial intelligence from machine learning to large language models. 

In the beginning, NLP systems are predominantly rule-based. These rule-based models are built on top of domain expert-framed rules. As manual rule framing is a laborious, expensive process and also requires frequent changes, rules-based models are gradually replaced by machine models, which learn the rules automatically from the training data and completely avoid manual rule framing \cite{kalyan2021ammus}. However, machine learning models require human intervention in the form of domain experts for feature engineering. The evolution of dense text vector representation models like Word2Vec \cite{mikolov2013efficient}, Glove \cite{pennington2014glove}, FastText \cite{bojanowski2017enriching} and the advancement of computer hardware like GPUs, NLP systems are built using traditional deep learning models like CNN \cite{kalchbrenner2014convolutional}, RNN \cite{salehinejad2017recent}, LSTM  \cite{hochreiter1997long}, GRU \cite{chung2014empirical}, Seq2Seq \cite{sutskever2014sequence} and Attention-based Seq2Seq models \cite{bahdanau2015neural, luong2015effective}. However, the drawbacks of these models like the inability to (i) capture long-term dependencies and (ii) leverage GPUs fully because of sequential processing (except in the case of CNN), resulted in the evolution of advanced deep learning models like Transformers \cite{vaswani2017attention}, which are fully attention based without any recurrent and convolution layers. 

Inspired by the success of image-pretrained models \cite{krizhevsky2012imagenet, simonyan2015very, szegedy2015going} built on top of transfer learning and large convolution models, the research community focused on building pretrained language models (PLMs) like BERT \cite{devlin2018bert} and GPT-1 \cite{radford2018improving} with transformers as the backbone and pretrained based on a new learning paradigm called self-supervised learning \cite{kalyan2021ammus, liu2021self, gui2023survey}. Unlike traditional deep learning models and vanilla transformers, which require training from scratch for downstream usage, pretrained language models can be easily adapted to downstream tasks with fine-tuning. The huge success of BERT and GPT-1 models triggered the development of other pretrained language models like RoBERTa, XLNet \cite{yang2019xlnet}, ELECTRA \cite{clark2019electra}, ALBERT \cite{lan2019albert}, DeBERTa \cite{he2022debertav3, he2020deberta}, GPT-2 \cite{radford2019language}, T5 \cite{raffel2020exploring}, BART \cite{lewis2020bart} etc. 

Although PLMs have many advantages compared to traditional deep learning and vanilla transformer models, they still suffer from drawbacks like the inability to generalize to unseen tasks without task-specific training. So, the research community focused on developing more advanced models like large language models which can generalize to unseen tasks without any task-specific training. The era of LLMs started with GPT-3 \cite{brown2020language}, and the success of GPT-3 inspired the development of other LLMs like PaLM \cite{chowdhery2022palm}, Chinchilla \cite{hoffmann2022training}, GLaM \cite{du2022glam}, LaMDA \cite{thoppilan2022lamda}, Gopher \cite{rae2021scaling}, Megatron–Turing NLG \cite{smith2022using}[181], BLOOM \cite{scao2022bloom}, Galactica \cite{taylor2022galactica}, OPT \cite{zhang2022opt}, LLaMA \cite{touvron2023llama, touvron2023llama2} etc. The popularity of LLMs is increasing exponentially after the recent launch of Open AI’s models like ChatGPT and GPT-4 \cite{openai2023gpt4}.  For example, ChatGPT has garnered millions of users within a few weeks of its launch.  Because of the ability to generalize to unseen tasks based on the task description and a few examples without requiring any task-specific training, just like humans, LLMs can be considered as a baby step towards Artificial General Intelligence \cite{bubeck2023sparks}. In this survey paper, we mainly focus on Open AI LLMs like GPT-3 models, GPT-3.5 models (InstructGPT, ChatGPT etc.) and GPT-4, which we refer to as GPT-3 family large language models (GLLMs).  This survey paper provides a comprehensive review of research works related to GLLMs in multiple dimensions.

\textbf{Contributions.}  The key contributions of this survey paper are
\begin{itemize}
    \item First survey paper to present a comprehensive review of GPT-3 family large language models (GLLMs) in multiple dimensions covering more than 350 recent research papers.
    \item We discuss various foundation concepts like transformers, transfer learning, self-supervised learning, pretrained language models and large language models.
    \item We discuss  GPT-3 family large language models in detail, starting from GPT-3 to the latest ChatGPT and GPT-4.
    \item We discuss the performances of GLLMs in various downstream tasks and present a thorough discussion on the data labelling, and data augmentation abilities of GLLMs. 
    \item We discuss the robustness and the evaluation abilities of GLLMs.
    \item We present multiple insightful future research directions which will guide the research community to improve the performances of GLLMs further.    
\end{itemize}

\textbf{Comparison with existing surveys.}
The existing survey papers provide a review of large language models \cite{zhao2023survey} and the relevant concepts like in-context learning \cite{dong2022survey}, evaluation \cite{chang2023survey, zhuang2023through}, alignment with human values \cite{wang2023aligning, liu2023trustworthy}, safety and trustworthiness \cite{huang2023survey}, reasoning \cite{huang2022towards}, challenges and applications \cite{kaddour2023challenges}, LLM compression \cite{zhu2023survey} and multi-modal LLMs \cite{yin2023survey}. For example, Zhao et al. \cite{zhao2023survey} are the first to provide a comprehensive of large language models. Unlike Zhao et al. \cite{zhao2023survey}, the other existing survey papers focus on specific concepts of LLMs. For example, the survey papers written by Dong et al. \cite{dong2022survey}, Chang et al. \cite{chang2023survey}, Wang et al. \cite{wang2023aligning} and Huang et al. \cite{huang2022towards} focus on in-context learning, evaluation of LLMs, alignment of LLMs with human values and reasoning ability of LLMs respectively. Similarly, the survey papers written by Yin et al. \cite{yin2023survey} and Huan et al. \cite{huang2023survey} provide a review of multi-modal LLMs and the safety and trustworthiness of LLMs, respectively. However, there is no existing survey paper which provides a comprehensive survey of GPT-3 family large language models.  With the ever-rising popularity of GPT-3 family large language models like GPT-3, InstructGPT, ChatGPT, GPT-4 etc. and a lot of research works using these models, there is a strong need for a survey paper which focuses exclusively on GPT-3 family large language models.

\textbf{Papers collection.} For this survey paper, we gathered over 350 research papers that appeared online in the period of June 2020 to September 2023. Initially, we selected GLLMs like GPT-3, InstructGPT, Codex and GPT-4 papers as seed papers and collected all the citing papers. We also collected papers from popular venues like ACL, EMNLP, COLING, AAAI, ICML, ICLR, NeurIPS etc and popular databases like Google Scholar and ScienceDirect using the keywords GPT-3, ChatGPT, GPT-3.5, InstructGPT, Codex and GPT-4. After removing the duplicate papers, we did a manual review to arrive at a final set of over 350 relevant research papers. 

\textbf{Survey paper organization.} The survey paper is organized as follows: Section \ref{section-2} presents a brief overview of various foundation concepts like transformers, transfer learning, self-supervised learning, pretrained language models and large language models.  Section \ref{section-4} presents  GPT-3 family large language models in detail, starting from GPT-3 to the latest ChatGPT and GPT-4. Sections \ref{section-5}, \ref{section-6}, and \ref{section-7} discuss the performances of GLLMs in various downstream tasks, specific domains and multilingual scenarios, respectively. Section \ref{section-8} presents the data labelling and data augmentation abilities of GLLMs. Section \ref{section-9} discusses various research works presenting approaches to detect text generated by GLLMs. Sections \ref{section-10} and \ref{section-11} discuss the robustness and evaluation abilities of GLLMs, respectively. Section \ref{section-12} presents multiple insightful future research directions. 

\section{Foundation Concepts}
\label{section-2}
\subsection{Transformer}
\subsubsection{Traditional Deep Learning Models}
Before the evolution of the transformer model, most of the research in natural language processing involved deep learning models like multi-layer perceptron (MLP), convolutional neural network (CNN), recurrent neural network (RNN), long short-term memory (LSTM) network, gated recurrent unit (GRU), sequence-to-sequence and attention-based sequence-to-sequence \cite{young2018recent}. MLP is a feed-forward neural network with three or more layers (input layer, one or more hidden layers, and output layer), and the neurons in these layers are fully connected. MLPs are easy to understand and simple to implement. However, as MLPs ignore the sequence information and struggle to capture the semantic relationships, these models are subsequently replaced by advanced models like CNN and RNN. CNN, originally developed to process images, is also explored for natural language processing tasks by treating text as a one-dimensional image \cite{ kalchbrenner2014convolutional, kim2014convolutional}. CNNs can learn local features (n-grams) effectively using convolution layers but struggle to capture long-term dependencies. RNNs evolved as a deep learning model exclusively to process sequential data like text, time series, etc \cite{salehinejad2017recent}. RNNs can handle input with varying lengths and process sequential data by maintaining a hidden state to capture the context from previous inputs. However, RNNs suffer from vanishing gradients problems and struggle to capture long-term dependencies. LSTM \cite{hochreiter1997long} and GRU \cite{chung2014empirical, cho2014learning} evolved as advanced RNN variants to address the issues with the vanilla RNN model. The gating mechanism in these models helps to regulate the flow of information along the sequence and retain the most important information. Compared to LSTM, which includes three gates (input, forget and output gates), GRU is more parameter efficient as it includes only two gates, namely the input and the reset gates. 

RNN and its variants like LSTM and GRU expect the input and output sequences to be the same length. However, in the case of natural language generation tasks like machine translation, text summarization, etc., the input and output sequences can be of different lengths. So, the researchers introduced the sequence-to-sequence (Seq2Seq) model to handle tasks with different input and output sequence lengths \cite{sutskever2014sequence}.  The Seq2Seq model is originally developed for machine translation and later explored for other NLP tasks. The Seq2Seq  model consists of an encoder and decoder based on RNN, LSTM or GRU to process the input sequence and generate the output sequence. The encoder processes the input sequence to generate a fixed-size context vector based on which the decoder generates the output sequence. However, the fixed-size context vector fails to encode the entire information in the input sequence, especially when the input sequence is long \cite{ bahdanau2015neural}. The attention mechanism is introduced to address this issue, allowing the decoder to focus on the relevant input tokens at each decoding step \cite{ bahdanau2015neural, luong2015effective}. However, as the encoder and decoder of the Seq2Seq model are based on RNN and its variants, the Seq2Seq model suffers from vanishing gradients and struggles to capture long-term dependencies.  

\subsubsection{Drawbacks of Traditional Deep Learning Models}
Here are the drawbacks of traditional deep learning models
\begin{itemize}
    \item \textit{Lack of sequence and semantic understanding} - MLPs ignore sequence information, treating all input tokens as independent. Moreover, MLPs can learn statistical patterns but struggle to capture semantic information in the input sequence. 
    \item \textit{Computationally expensive} - CNNs require a large number of parameters to achieve good results.  Although LSTM and GRU address the limitations of vanilla RNNs to some extent, these models include a gating mechanism which significantly increases the number of model parameters. The large number of parameters makes these models computationally expensive to train and use. 
    \item \textit{Vanishing gradients} - RNN suffer from vanishing gradients problem. Although LSTM and GRU address this problem to some extent, these models also suffer from vanishing gradient problem and have difficulties in capturing long-term dependencies. 
    \item \textit{Sequential Computation} - RNN and its variants process the input sequence token by token, i.e. sequentially. This sequential computation is a bottleneck for these models to leverage parallel computing capability in advanced computing hardware like GPUs and TPUs. This sequential computation also slows down training and inference processes, especially for long sequences.
\end{itemize}

\begin{figure*}[h!]
\begin{center}
\includegraphics[width=16cm, height=5cm]{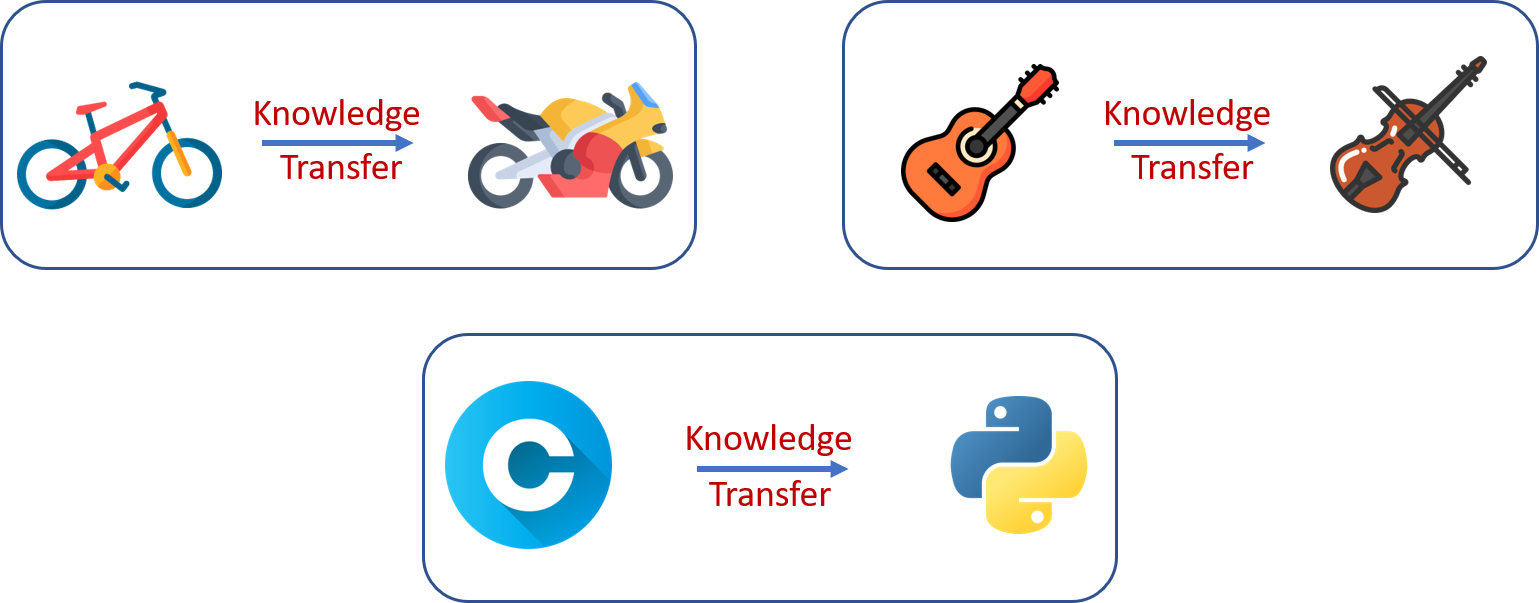}
\caption{\label{transfer} Real-life examples of knowledge transfer (transfer learning). Examples are inspired from \cite{zhuang2020comprehensive}} 
\end{center}
\end{figure*}

\subsubsection{Transformer Description}
The transformer model evolved as an effective alternative to traditional deep learning models and addressed most associated issues \cite{vaswani2017attention}. In no time, the transformer model, with its novel and efficient architecture, gained a lot of popularity and became a de facto choice for building pretrained language models and large language models using self-supervised learning paradigm \cite{kalyan2021ammus, zhao2023survey}. The key ingredient behind the massive success of the transformer model is its self-attention mechanism. The self-attention mechanism allows the transformer model to process the input sequence without using recurrent or convolution layers. This attention mechanism also allows the model to effectively capture long-range dependencies in the input sequence, making it highly effective for natural language understanding and generation tasks. 

The transformer consists of encoder and decoder components. The encoder processes the input text using a stack of encoder layers and then produces rich contextualized vector representations for each token in the input sequence, which are later used by the decoder. Each encoder layer consists of a self-attention mechanism and a feedforward neural network. The self-attention mechanism adds contextual information to the token vectors by allowing each token to attend to all other input tokens, and this helps the model to capture long-term dependencies better.  After the self-attention mechanism, the token vectors are passed through a feedforward neural network, which introduces non-linearity and further transforms the representations. In this way, each encoder layer applies self-mechanism and feed-forward network to add more contextual information to the token vector representations. 

The decoder receives the output from the last encoder layer and processes it sequentially by applying a stack of layers, with each decoder layer having masked self-attention, encoder-decoder self-attention and feed-forward neural network. The masked self-attention allows each token to attend to the previously generated tokens only and prevents the model from attending to future tokens. The encoder-decoder self-attention allows the decoder to attend to the encoded input sequence and helps the decoder focus on relevant input sequence tokens to generate the output tokens. 

The self-attention mechanism in the Transformer uses multiple attention heads, which allow the model to learn different aspects of relationships between tokens and encode more contextual information in the token representations. The encoder and decoder layers also include the embedding layer,  residual connections and layer normalization. The embedding layer transforms input tokens into vector representations where each vector representation encodes both the meaning and position information.  The residual connections and layer normalization are applied after the self-attention mechanism and feed-forward network. Residual connection avoids vanishing gradients and ensures a smooth flow of gradients, while layer normalization is applied to normalize the token representations and stabilize training. Apart from the embedding layer and stack of decoder layers, the decoder also includes an output layer. The output layer is nothing but a softmax layer that assigns probabilities to each token in the vocabulary, indicating the likelihood of each token being the next word in the generated sequence.

\subsection{Transfer Learning}
\subsubsection{Why Transfer Learning?}
Although machine learning models tasted some success, these models require feature engineering, which is a laborious and expensive process involving human intervention in the form of domain experts \cite{kalyan2021ammus}. Deep learning models, essentially a subset of machine learning, don’t require feature engineering as deep learning models learn features during training. Over the years, deep learning witnessed the evolution of various models like multi-layer perceptron (MLP), convolution neural networks (CNN), recurrent neural networks (RNN), long short-term memory networks (LSTM), gated recurrent unit networks (GRU), encoder-decoder networks, encoder-decoder with attention networks and recently transformers \cite{young2018recent, otter2020survey}.  Even though deep learning models eliminated the requirement of manual feature engineering and achieved significant progress, the main drawback with these models is the requirement of a large amount of labelled data to achieve good results. Along with developing various deep learning models, the research community also focused on developing high-quality datasets for various tasks \cite{han2021pre}. However, manual data annotation is a time-consuming, expensive and laborious process. Additionally, when there is a change in the data distribution, it is essential to re-train deep learning models with new labelled data to maintain good performances \cite{pan2009survey}. To reduce the costs, the research community focused on how to effectively train deep learning models with limited labelled data. Transfer learning evolved as one of the effective solutions to train deep learning models with limited labelled data \cite{zhuang2020comprehensive, pan2009survey}. 

\subsubsection{What is Transfer Learning?}
Transfer Learning in the context of artificial intelligence involves existing knowledge transfer from one task (or domain) to another different but related task (or domain) \cite{zhuang2020comprehensive, pan2009survey}. Transfer learning avoids training a model from scratch and helps improve the model's performance on the target task (or domain) by leveraging already existing knowledge. Transfer learning is largely based on the idea that when two tasks (or domains) are similar, the knowledge from the source task (or domain) with sufficient data can be used to enhance the performance of the target task (or domain) with limited data. For example, consider the task of sentiment analysis of reviews of different products. It is highly expensive to annotate large data separately for each product. In such cases, transfer learning helps to adapt the model trained on one product reviews to perform well on other product reviews without requiring large labelled data \cite{blitzer2007biographies}. 

Transfer learning draws inspiration from human beings, i.e., human beings can do new tasks without or with few examples just by reusing previously gained knowledge \cite{han2021pre}. Figure \ref{transfer} illustrates real-life examples of knowledge transfer (transfer learning).  For example, a person who can cycle can learn to ride a bike quickly with less effort. This is because riding a cycle and a bike involves a lot of common things like handling the balance, etc. Similarly, a person familiar with C programming language can learn Python programming language easily. This is because both C and Python are programming languages and share many common concepts. So, due to the ability to reuse the existing knowledge and train the target models with limited data, transfer learning evolved as a promising learning paradigm and eventually played a crucial role in the evolution of advanced deep learning models like pretrained language models \cite{kalyan2021ammus, kalyan2022ammu} and the recent large language models. Overall, the advantages of transfer learning are
\begin{itemize}
    \item Transfer learning helps to reduce the requirement of labelled data.  (Data efficiency)
    \item Transfer learning avoids training models from scratch by providing a good initialization from existing related models. (Faster training and development)
    \item Transfer learning helps to enhance the performance on the target task (or domain) by reusing existing knowledge. (Enhance target task performance)
    \item Transfer learning is explored across AI areas like computer vision, natural language processing, and speech processing. (Versatile)
\end{itemize}

In conclusion, transfer learning is a powerful learning paradigm in artificial intelligence that has benefits regarding data efficiency, speed, performance, adaptability, and real-world practicality.

\begin{figure*}[h!]
\begin{center}
\includegraphics[width=14cm, height=5cm]{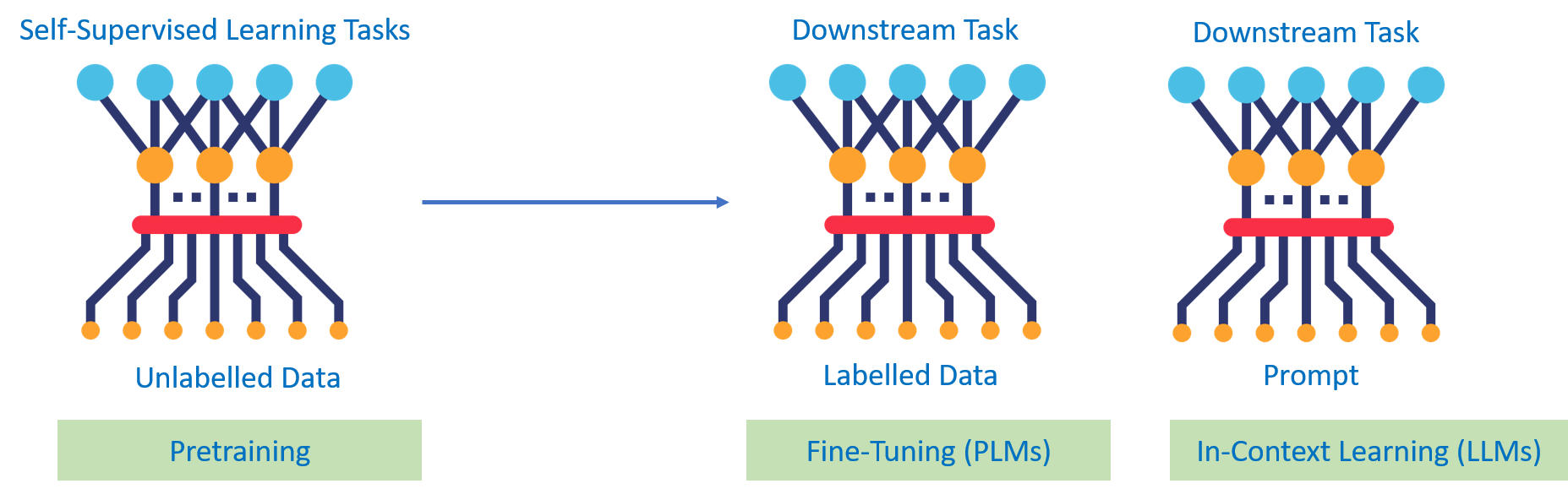}
\caption{\label{ssl1} Illustration of self-supervised learning paradigm.} 
\end{center}
\end{figure*}

\subsubsection{Transfer Learning vs Other Learning Paradigms}
Along with transfer learning, the other learning paradigms that evolved to address large labelled data requirements are semi-supervised learning \cite{van2020survey}  and multi-task learning \cite{zhang2021survey}. Semi-supervised learning is a learning paradigm in artificial intelligence that uses labelled and unlabelled data to train models \cite{van2020survey}. As semi-supervised learning uses labelled and unlabelled data, it lies between unsupervised and supervised learning paradigms. As semi-supervised learning uses only a small amount of labelled data, it reduces the amount of labelled data required, like transfer learning. However, unlike transfer learning, where the distribution of source and target tasks can be different, in semi-supervised, the distribution of labelled and unlabelled data should be the same \cite{zhuang2020comprehensive}. Multi-task learning is a learning paradigm which focuses on enhancing the performance of a group of tasks by leveraging the interconnections between the tasks and learning them simultaneously \cite{van2020survey}. Unlike multi-task learning, which simultaneously learns all the tasks, transfer learning first learns the source task and then transfers the knowledge to the target task. In multi-task learning, the focus is generally on all the tasks, while transfer learning focuses more on the target task \cite{pan2009survey}.

\subsection{Self-Supervised Learning}
\subsubsection{Why Self-Supervised Learning?}
The main drawback with traditional deep learning models like CNN is the requirement of training from scratch. Training from scratch requires a large amount of labelled data. Data labelling is not only expensive but also a time-consuming and laborious process, which eventually makes the model development expensive. To reduce the requirement of labelled data and make the model development process less expensive, the computer vision research community focused on developing models like VGGNet \cite{simonyan2015very}, AlexNet \cite{krizhevsky2012imagenet}  and GoogleNet \cite{szegedy2015going} on top of large CNNs, transfer learning and supervised learning. These models are pretrained on a large number of labelled images from ImageNet dataset \cite{deng2009imagenet} using supervised learning, and then adapted to downstream tasks. These pretrained models avoid training downstream models from scratch by providing a good initialization. Moreover, downstream models initialized from pretrained models converge faster and achieve good results even with limited labelled data \cite{han2021pre}.

\begin{figure*}[h!]
\begin{center}
\includegraphics[width=16cm, height=5cm]{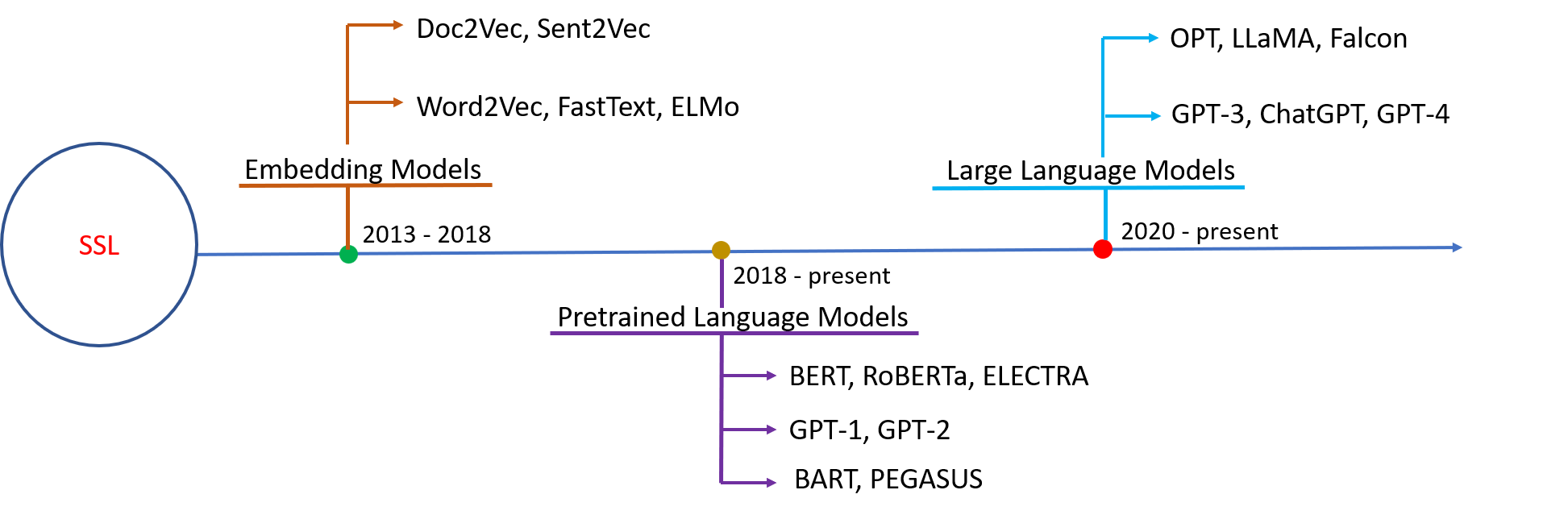}
\caption{\label{ssl2} Evolution of self-supervised learning in natural language processing.} 
\end{center}
\end{figure*}

Inspired by the huge success of pretrained image models, the NLP research community focused on developing pretrained language models \cite{han2021pre, kalyan2021ammus, kalyan2022ammu}. However, the main challenge here is the use of supervised learning at scale to pretrain language models. This is because supervised learning at scale requires huge volumes of labelled data, which is almost impossible to obtain in many cases because of highly expensive annotation costs. Besides high annotation costs, supervised learning also suffers from generalization errors and spurious correlations \cite{ kalyan2021ammus, gui2023survey}. Self-supervised learning with the ability to automatically generate the labels and make use of unlabelled data evolved as an effective alternative to supervised learning to pretrain language models at scale \cite{liu2021self, gui2023survey, kalyan2021ammus}.

\subsubsection{What is Self-Supervised Learning?}
Self-supervised learning, a promising learning paradigm in artificial intelligence, helps models from different modalities like language, speech or image to learn background knowledge from large volumes of unlabeled data \cite{liu2021self, gui2023survey}. Unlike supervised learning, which relies on large volumes of labelled data, self-supervised learning pretrains the models at scale based on the pseudo supervision offered by one or more pretraining tasks. Here, the pseudo supervision stems from the labels, which are automatically generated without human intervention based on the description of the pretraining task. In general, self-supervised learning involves one or more pretraining tasks  \cite{kalyan2021ammus, kalyan2022ammu}. Moreover, the efficiency of self-supervised learning is heavily influenced by the choice of pretraining task \cite{kalyan2021ammus, clark2019electra,  he2022debertav3}. 

Figure \ref{ssl1} presents the self-supervised learning paradigm. In the pretraining phase, the labels are automatically generated based on the description of pretraining tasks, and the models learn universal knowledge using the pseudo supervision offered by one or more pretraining tasks. Pretraining helps the models to gain strong background knowledge, which allows the models to provide a good initialization to downstream models. The initialization from pretrained models enhances the downstream models in terms of generalization, performance,  and robustness and makes them data efficient. After pretraining, pretrained language models can be easily adapted to downstream tasks with limited labelled data, and large language models can be used to solve downstream tasks using in-context learning without any task-specific fine-tuning.

\subsubsection{Evolution of Self-Supervised Learning}
Figure \ref{ssl2} shows the evolution of self-supervised learning in natural language processing from embedding models to the recent large language models. The evolution of self-supervised learning in natural language processing happened in three stages, namely embedding models, pretrained language models and large language models. Initially, self-supervised learning is explored to develop non-contextual embedding models (e.g. Word2Vec \cite{mikolov2013efficient}, FastText \cite{bojanowski2017enriching}), followed by sentence embedding (e.g. Sent2Vec \cite{pagliardini2018unsupervised})  and contextual embedding models (e.g. ELMo \cite{peters-etal-2018-deep}).  The quest to develop pretrained models motivated NLP researchers to explore self-supervised learning to develop pretrained language models \cite{kalyan2021ammus, kalyan2022ammu, han2021pre}. As pretrained language models cannot generalize to NLP tasks without fine-tuning, the NLP research community focused on developing large language models using self-supervised learning at a large scale \cite{brown2020language, touvron2023llama,  touvron2023llama2, anil2023palm, openai2023gpt4}. To summarize, self-supervised is undergoing a rapid evolution and is also treated as a significant element in achieving near human-level intelligence \cite{gui2023survey}. 

\subsubsection{Self-Supervised Learning vs Other Learning Paradigms}
Self-supervised learning, with its exceptional ability to make use of unlabelled data at scale, evolved as an alternative to supervised learning to pretrain models. However, self-supervised learning has similarities and dissimilarities with supervised learning \cite{kalyan2021ammus}. Both self-supervised and supervised provide supervision. However, unlike supervised learning, which offers supervision based on human-labelled data, self-supervised learning offers supervision based on automatically generated data. Supervised learning is mostly used to train downstream models with task-specific data, while self-supervised learning is used to train pretrained models to offer good initialization to downstream models.  Similarly, self-supervised learning has similarities and dissimilarities with unsupervised learning \cite{kalyan2021ammus}. Both self-supervised learning and unsupervised learning make use of unlabelled data without requiring any labelled data. However, unlike self-supervised learning, which focuses on learning rich data representations using pseudo supervision,  the main focus of unsupervised learning is to identify the hidden patterns in the data without any supervision.

\begin{figure*}[h!]
\begin{center}
\includegraphics[width=16cm, height=4cm]{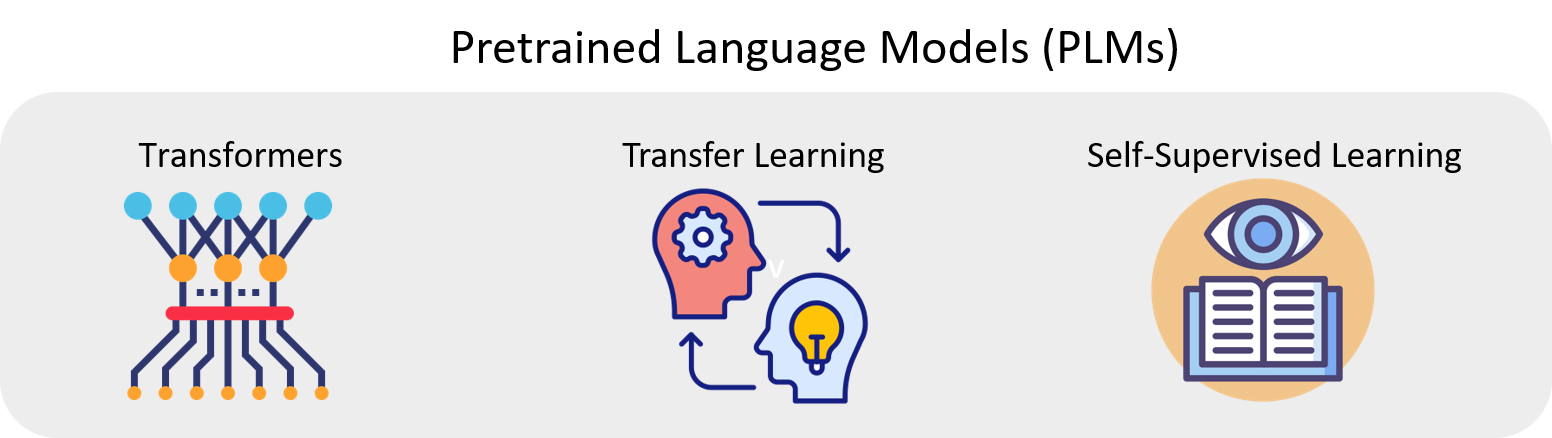}
\caption{\label{plms} Key ingredients in the evolution and success of pretrained language models.} 
\end{center}
\end{figure*}

\subsection{Pretrained Language Models}
\subsubsection{Overview} 
Deep learning witnessed the evolution of several models, from convolution neural networks to the latest transformers \cite{ young2018recent, otter2020survey}. Transformer addressed drawbacks of traditional deep learning models like convolutional neural network, recurrent neural network and its variants and achieved significant progress \cite{vaswani2017attention, lin2022survey}. However, transformer and traditional deep learning models suffer from one major drawback: training from scratch, which requires large volumes of labelled data and makes model development expensive. Inspired by the success of pretrained image models like VGGNet \cite{simonyan2015very}, AlexNet \cite{krizhevsky2012imagenet} and GoogleNet \cite{ szegedy2015going} in computer vision, NLP researchers focused on developing pretrained models for natural language processing based on transformers and self-supervised learning \cite{kalyan2021ammus, kalyan2022ammu, han2021pre, qiu2020pre}. Pretrained language models are advanced deep learning models essentially transformer-based, pretrained on large volumes of text data and can be adapted to downstream tasks with limited labelled data. Along with transformer model, self-supervised learning and transfer learning are key concepts which make pretrained language models possible \cite{kalyan2021ammus} (refer Figure \ref{plms}).  The era of pretrained language models started with GPT-1 \cite{ radford2018improving} and BERT \cite{devlin2018bert} models. The massive success of BERT and GPT-1 models triggered the development of other pretrained language models like RoBERTa \cite{liu2019roberta}, XLNet \cite{yang2019xlnet}, ELECTRA \cite{clark2019electra}, ALBERT \cite{ lan2019albert}, DeBERTa \cite{he2022debertav3, he2020deberta}, GPT-2 \cite{radford2019language}, T5 \cite{raffel2020exploring}, BART \cite{lewis2020bart}, PEGASUS \cite{zhang2020pegasus}  etc.

\subsubsection{Evolution of Pretrained Language Models}
The evolution of pretrained language models happened along three dimensions: encoder-based models, decoder-based models and encoder-decoder based models \cite{kalyan2021ammus}. Encoder-based models consist of an embedding layer and stack of encoder layers, with each encoder layer having self-attention and feed-forward networks. Encoder-based models are primarily used for natural language understanding tasks like text classification, entity extraction, relation extraction, etc. Some of the popular encoder-based pretrained language models are BERT, RoBERTa, XLNet, ALBERT, ELECTRA, DeBERTa, etc. Decoder-based models consist of an embedding layer and a stack of decoder layers, with each decoder layer having self-attention, masked self-attention and feed-forward networks. Decoder-based models are used for both natural language understanding and generation tasks. Some of the popular decoder-based pretrained language models are GPT-1, GPT-2 etc. Encoder-decoder based models consist of both encoder and decoder modules. In general, encoder-decoder based models are used for natural language generation tasks like machine translation, text summarization, etc., while some are explored for both natural language understanding and generation tasks. Some of the popular encoder-decoder based models are T5, BART, PEGASUS, M2M100, NLLB,  etc.  

After the massive success of pretrained language models in the English language, the research community started to develop multilingual pretrained language models \cite{doddapaneni2021primer} and pretrained language models for non-English languages \cite{kalyan2021ammus}. Some of the popular multilingual pretrained language models are mBERT \cite{devlin2018bert}, mT5 \cite{xue2021mt5}, mBART \cite{ liu2020multilingual}, IndicBERT \cite{kakwani2020indicnlpsuite},  XLM \cite{conneau2019cross}, XLM-R \cite{conneau2020unsupervised}, mDeBERTa \cite{he2022debertav3} etc. As the performance of general domain pretrained language models is limited in domain-specific tasks \cite{kalyan2021ammus,kalyan2022ammu}, the research community focused on developing pretrained language models for specific domains like social media \cite{nguyen2020bertweet,barbieri2020tweeteval}, finance \cite{yang2020finbert, araci2019finbert, liu2021finbert}, legal \cite{chalkidis2020legal, leivaditi2020benchmark}, coding \cite{feng2020codebert, wang2021codet5, wang2023codet5+}, healthcare \cite{lee2020biobert, gu2020domain, raj2021bioelectra} etc., As pretrained language models have millions of parameters which make model fine-tuning and deployment expensive, compact pretrained language models like DistilBERT \cite{sanh2019distilbert}, TinyBERT \cite{jiao2020tinybert}, MobileBERT \cite{sun2020mobilebert}, MiniLM \cite{wang2020minilm}etc., are developed. As pretrained language models have a limited context length which limits the performance on long sequences, long-sequence pretrained language models like LongFormer \cite{beltagy2020longformer}, BigBird \cite{zaheer2020big} etc., are developed. Pretrained language models encode only the universal language knowledge available in the pretraining corpus and lack valuable knowledge available in ontologies. So, the research community developed ontology-enriched models like SapBERT \cite{liu2021selfbert}, UmlsBERT \cite{michalopoulos2020umlsbert}, etc.  

\begin{figure*}[h!]
\begin{center}
\includegraphics[width=16cm, height=5cm]{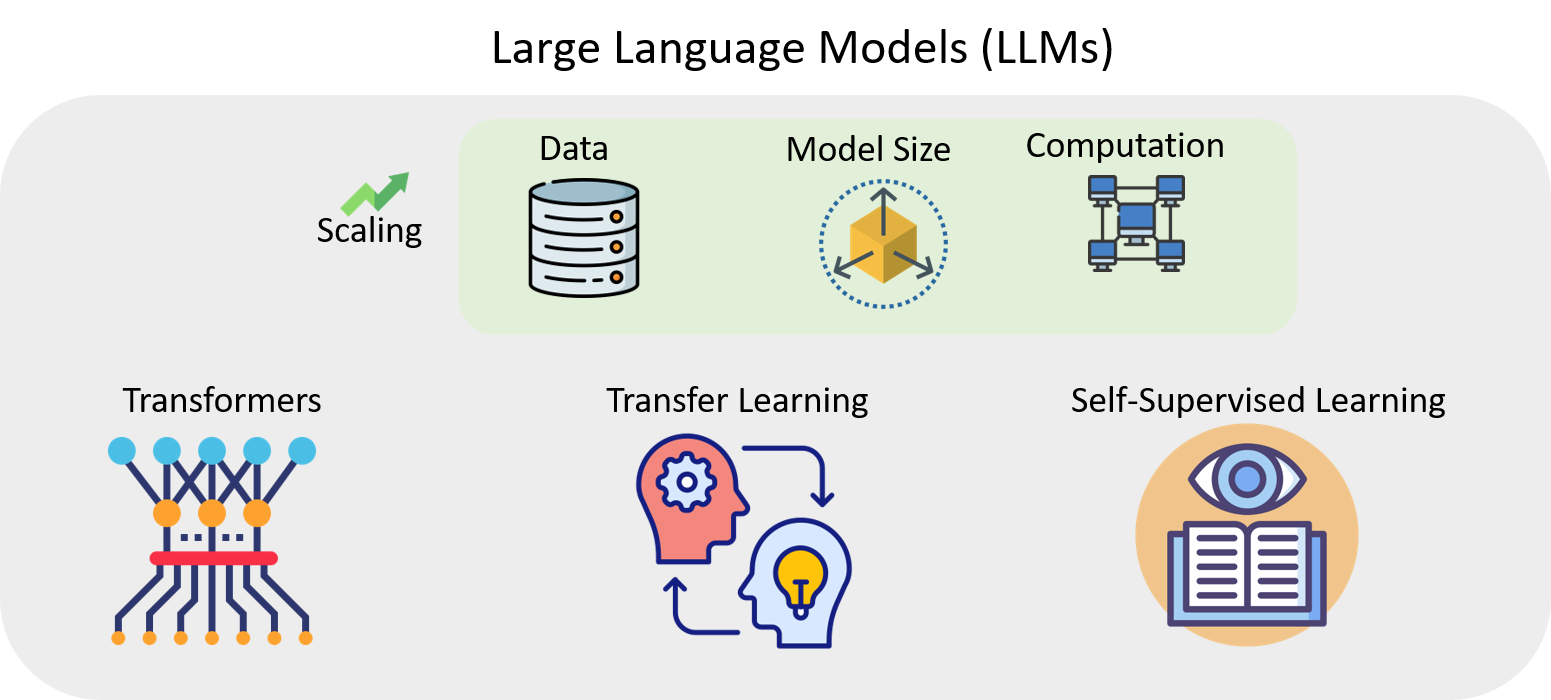}
\caption{\label{llms} Key ingredients in the evolution and success of large language models.} 
\end{center}
\end{figure*}

\subsection{Large Language Models}

\subsubsection{Overview}
The pretrained language models, starting from GPT-1 \cite{radford2018improving}, BERT \cite{devlin2018bert} models to the latest DeBERTa \cite{he2022debertav3, he2020deberta}, achieved significant progress and also reduced the amount of labelled data required to train the task-specific models \cite{kalyan2021ammus,kalyan2022ammu}. Pretrained language models follow the paradigm “pretrain then fine-tune”, i.e., the model is pretrained first and then adapted to downstream tasks by fine-tuning. As task-specific fine-tuning is mandatory to adapt the pretrained language model to downstream tasks, pretrained language models cannot generalize to unseen downstream tasks without task-specific fine-tuning. Moreover, task-specific fine-tuning requires labelled data and creates a separate copy of the pretrained language model for each downstream NLP task, increasing the model development and deployment costs \cite{kalyan2021ammus}. 

Pretrained language models are treated as narrow AI systems as they are adapted through fine-tuning and then used for specific downstream tasks. However, the main focus of the research community is to develop artificial general intelligence systems \cite{goertzel2014artificial,bubeck2023sparks}  which are not narrowly focused on specific tasks but have the ability for general problem-solving and can handle even the unseen tasks by utilizing the existing knowledge like human beings. The NLP researchers observed that the performance of pretrained language models can be enhanced further through scaling along three dimensions: pretraining computation, pretraining data and model size \cite{liu2019roberta, radford2019language, raffel2020exploring}. Large size allows the models to capture more nuanced language patterns, which in turn enhances their ability to understand and generate text, while large pretraining data helps the model to learn from a wider range of text. The promising results from scaling and the quest to build artificial general intelligence systems motivated NLP researchers to build much bigger and bigger models, which eventually resulted in the evolution of GPT-3 and its successor models \cite{ brown2020language, chowdhery2022palm,  hoffmann2022training, du2022glam}. Learning paradigms like transfer learning and self-supervised learning make large language models possible, but scaling makes these models powerful.

The research community coined a new phrase, “large language models”, to refer to GPT-3 and its successor large models to differentiate these models from small pretrained language models \cite{zhao2023survey}. Large language models (LLMs) are a special class of pretrained language models obtained by scaling model size, pretraining corpus and computation as showin in Figure \ref{llms}.  Large language models (LLMs) are essentially deep learning models, specifically transformer-based, pretrained on large volumes of text data and aligned to human preferences using meta-training. Pretraining provides universal language knowledge to the model \cite{kalyan2021ammus}, while meta-training aligns the model to act based on the user's intentions. Here, the user's intention includes explicit intentions, like following instructions, and implicit intentions, like maintaining truthfulness and avoiding bias, toxicity, or harmful behaviour \cite{ouyang2022training}. 

Because of their large size and pretraining on large volumes of text data, LLMs exhibit special abilities referred to as emerging abilities \cite{wei2022emergent, schaeffer2023emergent}, allowing them to achieve remarkable performances without any task-specific training in many natural language processing tasks. For downstream task usage, pretrained language models leverage supervised learning paradigm, which involves task-specific fine-tuning and hundreds or thousands of labelled instances \cite{kalyan2021ammus, kalyan2022ammu}. LLMs leverage in-context learning (ICL), a new learning paradigm that doesn’t require task-specific fine-tuning and many labelled instances \cite{brown2020language, dong2022survey}. LLMs treat any NLP task as a conditional text generation problem and generate the desired text output by conditioning on the input prompt, including task description, test input and optionally, a few examples.

\begin{figure*}[h!]
\begin{center}
\includegraphics[width=15.5cm, height=5cm]{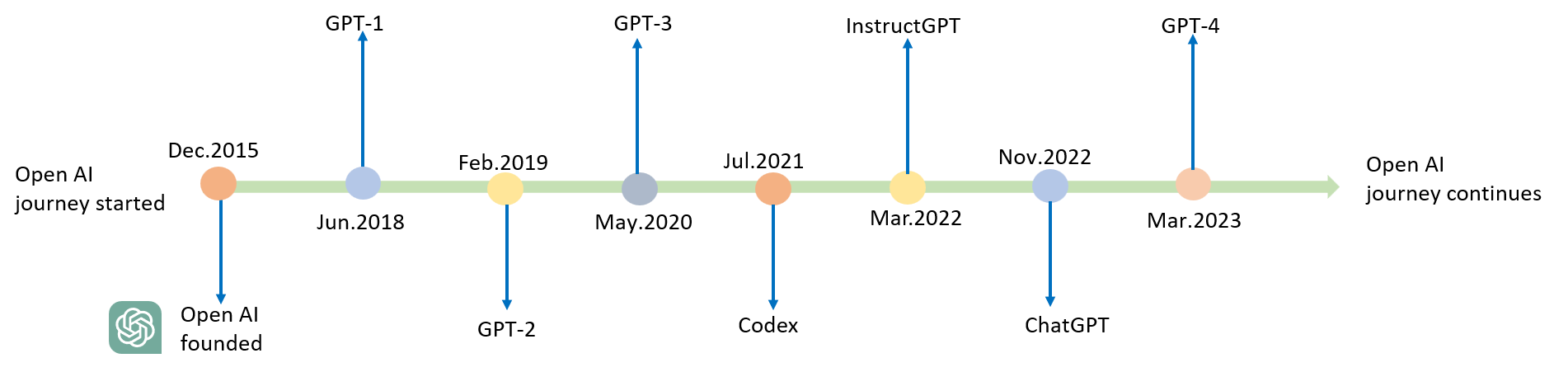}
\caption{\label{gllms-evolution} Open AI journey starting from GPT-1 to the latest GPT-4.} 
\end{center}
\end{figure*}

\subsubsection{Evolution of Large Language Models}
The evolution of large language models happened along two dimensions: closed-source LLMs and open-source LLMs. The era of LLMs roughly started with GPT-3. Following the success of GPT-3, Open AI developed successor models like InstructGPT \cite{ouyang2022training}, Codex \cite{chen2021evaluating}, ChatGPT and GPT-4 \cite{openai2023gpt4}. Google introduced models like GLaM \cite{du2022glam}, PaLM \cite{chowdhery2022palm}, PaLM2 \cite{anil2023palm}, LaMDA \cite{thoppilan2022lamda} and Bard. DeepMind developed models like Gopher \cite{rae2021scaling}, Chinchilla \cite{hoffmann2022training}, AlphaCode \cite{li2022competition} and Sparrow \cite{glaese2022improving}. Companies like Baidu, AI21 labs and Amazon developed the models Ernie 3.0 Titan \cite{wang2021ernie}, Jurassic-1 \cite{lieber2021jurassic} and AlexaTM \cite{soltan2022alexatm}, respectively. Although the performances of closed-source LLMs are impressive, the main drawback with these models is that they are behind the paywalls, i.e., their weights are not publicly available, only some of them are accessible only through the APIs offered by the respective companies, and the model usage is charged based on the tokens processed and generated. 

To address this issue, the research community focused on developing open-source LLMs with publicly available weights. Some of the popular open-source LLMs are OPT \cite{zhang2022opt}, OPT-IML \cite{iyer2022opt}, Galactica \cite{taylor2022galactica}, LLaMA \cite{touvron2023llama}, LLaMA2 \cite{touvron2023llama2} and Falcon. The performances of these open-source LLMs are on par with closed-source LLMs. Moreover, in some cases, open-source LLMs outperform closed-source LLMs. For example, Galactica beats closed-source LLMs like GPT-3, Chinchilla and PaLM. Inspired by the success of open-source LLMs in the English language, the research community focused on developing multilingual and bilingual LLMs. BLOOM \cite{scao2022bloom} and BLOOMZ \cite{muennighoff2022crosslingual} are examples of multilingual LLMs, JAIS  \cite{sengupta2023jais} (English and Arabic), GLM \cite{zeng2022glm} (English and Chinese)  and FLM-101B \cite{li2023flm} (English and Chinese) are examples of bilingual LLMs. 

\begin{figure*}[h!]
\begin{center}
\includegraphics[width=18cm, height=8cm]{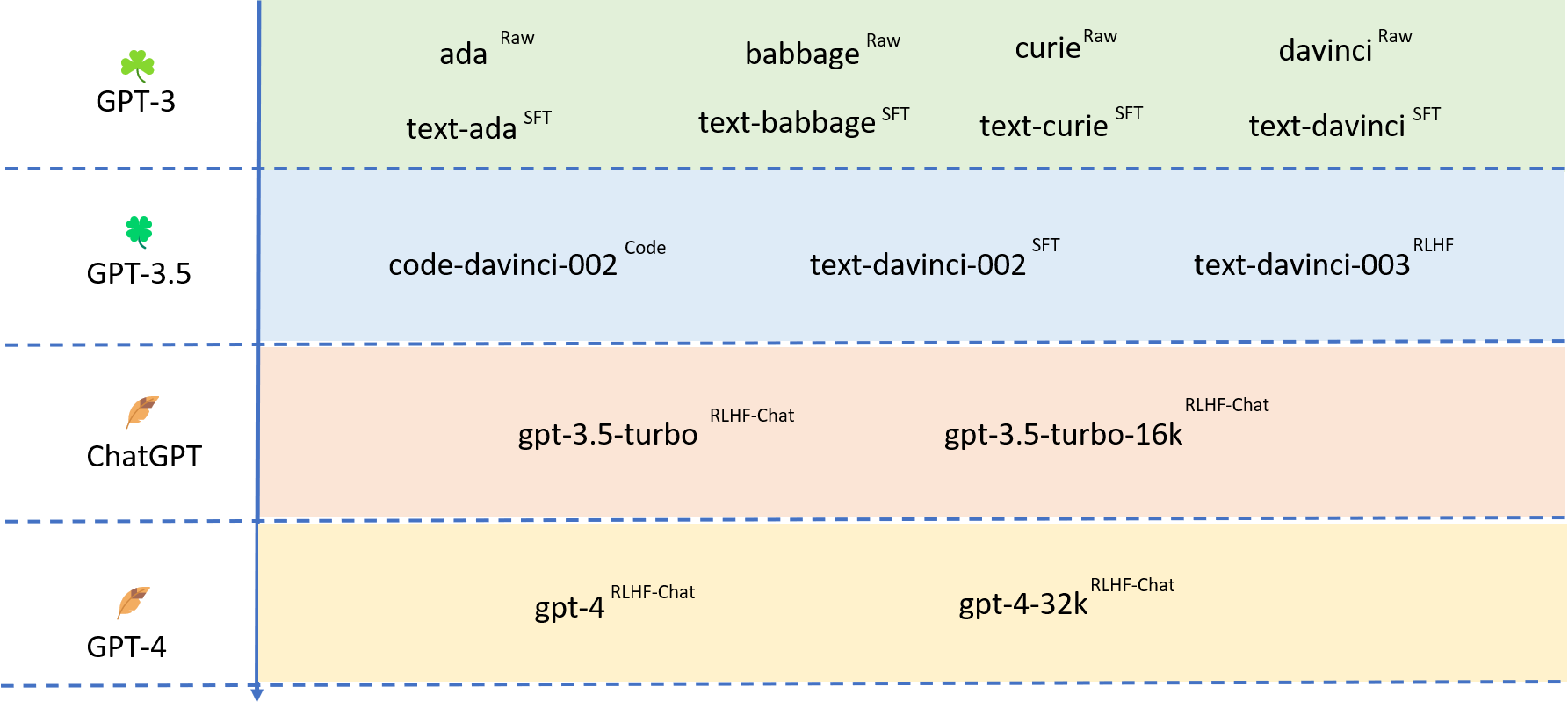}
\caption{\label{gllms-family} GPT-3 family large language models (GLLMs) starting from GPT-3 series to the latest GPT-4. Here, SFT stands for supervised fine-tuning, and RLHF stands for reinforcement learning from human feedback. Here, raw represents that the model is just pretrained and is not aligned using SFT or RLHF. Here, RLHF-Chat represents that the model is aligned using RLHF and optimized for chat.} 
\end{center}
\end{figure*}

The success of closed and open-source LLMs in the general domain triggered the development of domain-specific LLMs like FinGPT \cite{yang2023fingpt} and BloombergGPT \cite{wu2023bloomberggpt} in the finance domain, MedPaLM \cite{singhal2023large} and  MedPaLM2 \cite{singhal2023towards} in the healthcare domain and StarCoder \cite{ li2023starcoder}, CodeLlaMa \cite{roziere2023code}, CodeGen \cite{nijkamp2022codegen} and CodeGen2 \cite{nijkamp2023codegen2} in the coding domains. For example, Bloomberg developed BloombergGPT, an exclusive LLM for the finance domain. Similarly, Google developed MedPaLM and MedPaLM2 LLMs exclusively for the healthcare domain based on PaLM and PaLM2 models respectively. Similarly, HuggingFace developed StarCoder, MetaAI developed Code LlaMA, and SalesForce developed CodeGen and CodeGen2 LLMs exclusively for coding tasks.

\section{GPT-3 Family Large Language Models}
\label{section-4}
\subsection{Overview} 
Open AI, an AI company established in 2015, focused on building generative models. The Open AI researchers initially explored RNNs for developing generative language models \cite{radford2017learning}. Inspired by the huge success of the transformer model and its ability to capture long-term dependencies, Open AI researchers leveraged the transformer decoder to build GPT-1 (117M parameters), the first-ever transformer-based pretrained language model \cite{radford2018improving}. GPT-1 introduced a new paradigm, “pretrain and fine-tune”, to develop downstream task models effectively. Originally, the “pretrain and fine-tune” paradigm was introduced by  Dai et al. \cite{dai2015semi} and then explored by Howard and Ruder \cite{howard2018universal} to build language models for text classification. However, unlike Radford et al. \cite{radford2018improving} work, these research works build language models based on LSTM, which lacks parallelization ability and has difficulties in capturing long-term dependencies.  Radford et al. \cite{radford2018improving} used casual language modeling as a pretraining task to pretrain the GPT-1 model. The casual language modeling pretraining task involves generating the next token based on the previous tokens. GPT-1 achieved SOTA results in 9 out of 12 NLP tasks \cite{radford2018improving}. 

Inspired by the success of GPT-1, Open AI researchers introduced the GPT-2 model to push the results further \cite{radford2019language}. The GPT-2 model is pretrained on the WebText corpus (40B text), which is much larger than the Books corpus used to pretrain the GPT-1 model. The authors developed four versions of the GPT-2 model with varying parameters: 117M, 345M, 762M and 1.5B. The authors observed that the perplexity decreases with an increase in the model's size, and even for the largest version of 1.5B, the decrease in perplexity did not exhibit saturation. This revealed that GPT-2 underfitted the pretraining dataset, and extending the training duration could have further reduced perplexity. This observation triggered the insight that “developing even larger language models will decrease the perplexity further and enhance natural language understanding and generation capabilities”. The insights gained from the GPT-1 and GPT-2 models laid a strong foundation for the evolution of the GPT-3 family large language models, including the latest models like ChatGPT and GPT-4. Figure \ref{gllms-evolution} shows the journey of Open AI starting from GPT-1 to the latest GPT-4 and Figure \ref{gllms-family} shows the GPT-3 family large language models starting from GPT-3 series to the latest GPT-4.

\subsection{GPT-3 Models}
The experiment results of GPT-2 showed that increasing the model size further reduces the perplexity, and the model with more parameters achieves better results than the models with fewer parameters. This observation motivated Open AI researchers to train much bigger GPT models, which eventually resulted in the introduction of the GPT-3 model \cite{brown2020language}. GPT-3 model contains 175B parameters and is 100 times bigger than its predecessor model, GPT-2. Moreover, the GPT-3 model is trained over a corpus with the text from multiple sources like webpages, Wikipedia and books, unlike GPT-1 and GPT-2 models, which are pretrained over corpora with the text from books and webpages, respectively. Scaling in three dimensions: pretraining data, model size, and pretraining computation allows the GPT-3 model to learn more from large volumes of texts from different sources, which eventually empowers the model to handle unseen tasks without any task-specific training. Unlike GPT-1 and GPT-2 models, which leverage supervised learning to do downstream tasks, GPT-3 leverages training-free in-context learning. In-context learning is a new learning paradigm that is training-free and solves the downstream tasks by using knowledge encoded in the model parameters \cite{dong2022survey}.  In-context learning accepts prompts as input where the input prompt consists of task descriptions, optimally few examples and other instructions. 

\subsection{GPT-3.5 Models}
Two main drawbacks of the GPT-3 model are (i) GPT-3 is not trained over code data, and hence, it lacks complex reasoning abilities like solving math problems \cite{zhao2023survey}, and (ii) GPT-3 model struggles to follow user instructions and sometimes generate harmful text \cite{ouyang2022training}. These two drawbacks are addressed by GPT-3.5 models.  Brown et al. \cite{brown2020language} observed that GPT-3 can generate simple programs, although it is not specifically trained for generating code.  The Open AI researchers triggered by this observation introduced Codex \cite{chen2021evaluating}, an exclusive GLLM for coding tasks. Codex is developed by fine-tuning a GPT model with 12B parameters over publicly available Github code. Moreover, it is observed that GPT models explicitly trained over code data exhibit better reasoning capabilities.  

During pretraining, the GPT-3 model is optimized based on the casual language modeling objective, which involves predicting the next word based on the previous words. In-context learning during inference can be viewed as conditional text generation, where the model generates the output by conditioning on the given prompt. The model performs text generation during pretraining and inference, but it does vanilla text generation during pretraining and conditional text generation during inference. During pretraining, the model conditions on the previous words and generates the next word, i.e., vanilla text generation. However, during in-context learning, the model conditions on the prompt and generates the answer rather than generating the next words, i.e., conditional text generation. So, there is a gap between pretraining and in-context learning at inference. Due to this, in many cases during inference, the GPT-3 model fails to understand the given prompt and tends to generate the next words. 

The pretraining corpus of the GPT-3 model includes some amount of text with undesired qualities like misinformation, abuse, hate, sexism, etc., due to which the model sometimes generates harmful text. To enhance complex reasoning ability,  the instruction following ability and reduce the harmful text generation, GPT-3.5 models are developed by fine-tuning GPT-3 models over code data and then aligned using supervised fine-tuning (SFT) or reinforcement learning from human feedback (RLHF) \cite{ouyang2022training}. For example, the text-davinci-002 model is developed by fine-tuning the GPT-3 model (text-davinci) over code data to get code-davinci-002, which is further aligned using SFT.

\subsection{ChatGPT and GPT-4}
GPT-3 models are capable of understanding and generating natural language, while GPT-3.5 models are capable of understanding and generating both natural language and code.  However, both GPT-3 and GPT-3.5 models are not chat optimized. This drawback is addressed by ChatGPT (GPT-3.5-turbo) and GPT-4 \cite{openai2023gpt4} models.  Open AI introduced ChatGPT in November 2022. With extraordinary conversational abilities, ChatGPT, ChatGPT has garnered millions of users within a few weeks of its launch. Following ChatGPT,  Open AI released the GPT-4 model in March 2023, which can handle both text and image inputs. Apart from generating text with human-like fluency,  these models further pushed the results in many natural language processing tasks. The performance of these models in downstream tasks and specific domains is discussed in detail in Sections \ref{section-5} and \ref{section-6}.

 \begin{table*}[h!]
\begin{center}
{\renewcommand{\arraystretch}{1.5}% for the vertical padding
\begin{tabular}{|p{0.6cm}|p{5.6cm}|p{3cm}|p{1cm}|p{1.6cm}|p{2cm}|p{0.9cm}|}
\hline

\scriptsize{\textbf{Paper}} &  \scriptsize{\textbf{Task(s)}} & \scriptsize{\textbf{GLLMs Explored}} & \scriptsize{\textbf{Prompt Settings}} & \scriptsize{\textbf{Domain(s)}} & \scriptsize{\textbf{Language(s)}} & \scriptsize{\textbf{SOTA Results}}\\ \hline
  
\scriptsize{\cite{zhang2023investigating}} & \scriptsize{Stance Detection} &	\scriptsize{ChatGPT}	 & \scriptsize{ZS, FS} &	\scriptsize{Social Media} &	\scriptsize{English} & \scriptsize{No} \\ \hline  

\scriptsize{\cite{lamichhane2023evaluation}} & 	\scriptsize{Stress Detection, Depression Detection , Suicidal Detection} &	\scriptsize{ChatGPT}	 &	\scriptsize{ZS}	& \scriptsize{Social Media}  &	\scriptsize{English} & No \\ \hline 

\scriptsize{\cite{ yang2023evaluations}}	& \scriptsize{ Mental Health Analysis Tasks} &	\scriptsize{ChatGPT}	&	\scriptsize{ZS}	& \scriptsize{Social Media} &	\scriptsize{English} &  No \\ \hline

\scriptsize{\cite{wang2023chatgpt}} &	\scriptsize{Sentiment Analysis}	& \scriptsize{ChatGPT}	&	\scriptsize{ZS, FS}	& \scriptsize{Social Media}	& \scriptsize{English, Chinese} & \scriptsize{No} \\ \hline

\scriptsize{\cite{lopez2023can}} &	\scriptsize{Stock Prediction based on Sentiment Analysis}	& \scriptsize{ChatGPT}	& \scriptsize{ZS}	& \scriptsize{Finance}	& \scriptsize{English} & \scriptsize{No} \\ \hline

\scriptsize{\cite{ziems2023can}} &	\scriptsize{Computational Social Science Tasks}	& \scriptsize{GPT-3, ChatGPT}  &	\scriptsize{ZS}	& \scriptsize{Social Media} &	\scriptsize{English} & \scriptsize{No} \\ \hline

\scriptsize{\cite{kuzman2023chatgpt}} &	\scriptsize{Genre Identification} &	\scriptsize{ChatGPT}	&	\scriptsize{ZS}	& \scriptsize{General}	& \scriptsize{English, Slovenian} & \scriptsize{No} \\ \hline

\scriptsize{\cite{bang2023multitask}} &	\scriptsize{Sentiment Analysis, Misinformation Detection} &	\scriptsize{ChatGPT}	&	\scriptsize{ZS}	& \scriptsize{Social Media}	& \scriptsize{English, Indonesian, Javanese, Buginese} & \scriptsize{No} \\ \hline

\scriptsize{\cite{kocon2023chatgpt}} & \scriptsize{Nine NLU tasks including Sentiment Analysis and Natural Language Inference} & \scriptsize{ChatGPT} & \scriptsize{ZS} & \scriptsize{General, Social Media} & \scriptsize{English} & \scriptsize{No} \\ \hline 

\scriptsize{\cite{zhong2023can}} 	& \scriptsize{Paraphrase Detection, Sentiment Analysis, Natural Language Inference}	& \scriptsize{ChatGPT}	&	\scriptsize{ZS,FS}	& \scriptsize{General}	& \scriptsize{English} & \scriptsize{No} \\ \hline

 \scriptsize{\cite{ye2023comprehensive}} & \scriptsize{ Sentiment Analysis, Natural Language Inference} & \scriptsize{GPT-3, GPT-3.5, ChatGPT} & \scriptsize{ZS, FS} & \scriptsize{General, Social Media} & \scriptsize{English} & \scriptsize{No} \\ \hline 

\scriptsize{\cite{li2023chatgpt}} 	& \scriptsize{Financial News Classification , Sentiment Analysis}	& \scriptsize{ChatGPT, GPT-4}	&	\scriptsize{ZS}	& \scriptsize{Finance}	& \scriptsize{English} & \scriptsize{No} \\ \hline

\scriptsize{\cite{wu2023exploring}} 	& \scriptsize{Natural Language Inference}	& \scriptsize{ChatGPT, GPT4}	&	\scriptsize{ZS,FS}	& \scriptsize{Healthcare}	& \scriptsize{English} & \scriptsize{No} \\ \hline 

\scriptsize{\cite{wang2023large}}	& \scriptsize{Natural Language Inference, Document Classification} &	\scriptsize{GPT3.5, GPT4, Bard}	&	\scriptsize{ZS, FS}	& \scriptsize{Healthcare}	& \scriptsize{English} & \scriptsize{No} \\ \hline

\scriptsize{\cite{chiu2021detecting}} 	& \scriptsize{Hate Speech Detection}	& \scriptsize{GPT-3}	& \scriptsize{ZS, FS}	& \scriptsize{Social Media} &	\scriptsize{English} & \scriptsize{No} \\ \hline

\scriptsize{\cite{huang2023chatgpt}}	& \scriptsize{Implicit Hate Speech Detection} &	\scriptsize{ChatGPT}	&	\scriptsize{ZS}	& \scriptsize{Social Media}	& \scriptsize{English} & \scriptsize{No} \\ \hline

\scriptsize{\cite{chen2023evaluation}} 	& \scriptsize{Clinical Text Classification}	& \scriptsize{GPT-3, ChatGPT, GPT-4}	&	\scriptsize{ZS, FS}	& \scriptsize{Healthcare}	& \scriptsize{English} & \scriptsize{No} \\ \hline

\scriptsize{\cite{amin38will}}	& \scriptsize{Sentiment Analysis, Suicide Tendency Detection, Personality Prediction} &	\scriptsize{ChatGPT}	&	\scriptsize{ZS}	& \scriptsize{Social Media}	& \scriptsize{English} & \scriptsize{No} \\ \hline 

\scriptsize{\cite{parikh2023exploring}}	& \scriptsize{Intent Classification}	& \scriptsize{GPT-3}	&	\scriptsize{ZS}	& \scriptsize{Social Media} &	\scriptsize{English} & \scriptsize{No} \\ \hline

\scriptsize{\cite{sun2023text}} 	& \scriptsize{News Classification, Sentiment Analysis}	& \scriptsize{InstructGPT}	&	\scriptsize{ZS, FS}	& \scriptsize{General, Social Media} &	\scriptsize{English} & \scriptsize{Yes} \\ \hline

\end{tabular}}
\end{center}
\caption{ \label{gllms-tc}  Summary of research works exploring GLLMs for various text classification problems. Here ZS represents zero-shot, and FS represents few-shot.} 
\end{table*}

\section{Performance of  GLLMs in Downstream Tasks}
\label{section-5}
\subsection{Text Classification}
\textbf{Overview.} Text Classification is one of the fundamental tasks in natural language processing \cite{li2022survey}. It involves assigning label(s) from a predefined set of labels to a given piece of text. Here, the piece of text can be a phrase, sentence, paragraph or even a document. Many of the natural language processing problems, like offensive language identification, stance detection, sentiment analysis, hate speech detection, etc., are approached as text classification. Text Classification can be binary, multi-class or multi-label. 

In the case of text classification, the large language model is prompted with a task description, a predefined set of labels, examples (optional) and the test input. Here, task description, a predefined set of labels and examples constitute the context. The model understands what actually the task is from the context and then assigns the most appropriate label(s) to the given test input. The additional inputs, like examples in the context, enrich the prompt with more information which allows the model to understand the task better and then perform better. \\ 

\textbf{Research works exploring GLLMs for text classification.} The recent works explored GLLMs like GPT-3, GPT-3.5 ChatGPT and GPT-4 for various text classification problems like sentiment analysis \cite{wang2023chatgpt, lopez2023can, bang2023multitask, zhong2023can, li2023chatgpt, amin38will, sun2023text}, stance detection \cite{zhang2023investigating}, intent classification \cite{parikh2023exploring}, mental health analysis \cite{lamichhane2023evaluation, yang2023evaluations}, hate speech detection \cite{chiu2021detecting, huang2023chatgpt}, misinformation detection \cite{bang2023multitask}, paraphrase detection \cite{zhong2023can}, news classification \cite{li2023chatgpt}, natural language inference \cite{zhong2023can, wu2023exploring, wang2023large}etc. The evaluation is done in zero and few-shot settings using different prompting strategies like chain-of-thought (CoT) \cite{zhang2023investigating, yang2023evaluations, zhong2023can, wu2023exploring, wang2023large, chen2023evaluation, sun2023text}, self-question prompting (SQP) \cite{wang2023large}, clue and reasoning prompting (CARP) \cite{sun2023text} etc. Most of the research works focused on English datasets, except a few research works focused on other languages like Chinese \cite{wang2023chatgpt}, Slovenian \cite{kuzman2023chatgpt}, Indonesian \cite{bang2023multitask}, Javanese \cite{bang2023multitask}, and Buginese \cite{bang2023multitask}. A brief summary of research works exploring GLLMs for various text classification problems is presented in Table \ref{gllms-tc}.

Most of the research works showed that compared to direct prompting, advanced prompting strategies help the model to achieve better results. This is because advanced prompting involves generating intermediate outputs, which in turn guide the model in generating the correct final output. Zhang et al. \cite{zhang2023investigating} explored the ChatGPT model with direct and chain-of-thought prompting for stance detection in tweets in zero and few-shot settings. Experiment results on three datasets showed that one-shot chain of thought prompting outperforms zero-shot direct prompting and also achieves near state-of-the-art results. Yang et al. \cite{yang2023evaluations} designed emotion-enhanced CoT prompting to combine emotion information with the power of CoT prompting for mental health analysis tasks. Experiments on five different mental health analysis tasks showed that ChatGPT with emotion-enhanced CoT outperforms other prompting strategies. Overall, ChatGPT outperforms traditional deep learning models like CNN and RNN but still lags behind task-specific fine-tuned models. Wu et al. \cite{wu2023exploring} explored models like GPT-4 and ChatGPT for radiology natural language inference task. The authors reported that GPT-4 with IRSA prompting strategy outperforms ChatGPT in both zero and few-shot settings. IRSA stands for Instruction Response Semantic Alignment. IRSA prompting strategy is almost the same as direct prompting except that in the case of IRSA prompting, the model is instructed to give the labels “contain” and “not contain” instead of “entailment” and “not entailment”, just to reduce the complexity. Wang et al. \cite{wang2023large} evaluated the performances of the latest LLMs like GPT-3.5, GPT-4, and Bard models on text classification tasks like natural language inference and document classification in the healthcare domain. The GPT-4 model with the newly designed self-question prompting (SQP) outperforms other models in both zero and few-shot settings. The SQP strategy involves identifying the key elements of input, generating questions and answers related to the key elements, and then using them to generate the final output. Parikh et al. \cite{parikh2023exploring} showed that the performance of the GPT-3 model for intent classification in zero-shot settings can be enhanced by including intent class descriptions in the prompt. 

Some of the research works demonstrated that GPT-3 family large language models can outperform task-specific fine-tuned models \cite{kuzman2023chatgpt, zhong2023can} and domain-specific LLMs \cite{li2023chatgpt}. Kuzman et al. \cite{kuzman2023chatgpt} showed that ChatGPT outperforms fine-tuned XLM-R model in the task of automatic genre identification in the English language. Zhong et al. \cite{zhong2023can} compared the performances of ChatGPT and fine-tuned models based on base and large versions of BERT and RoBERTa models on tasks like natural language inference, sentiment analysis and paraphrase identification. The results showed that ChatGPT outperforms both base and large fine-tuned models by a large margin in the case of natural language inference task. Li et al. \cite{li2023chatgpt} evaluated the performances of general LLMs like ChatGPT and GPT-4 and domain-specific LLMs like BloombergGPT on tasks like finance news classification and sentiment analysis. In the case of finance news classification, GPT-4 outperforms all other LLMs, including the domain-specific BloombergGPT model. 

In all the above discussed research works, the performance of GLLMs is impressive but still lags behind SOTA results. Sun et al. \cite{sun2023text} showed that it is possible to achieve SOTA results in text classification tasks with the newly designed clue And reasoning prompting (CARP) prompting strategy. CARP involves a progressive reasoning approach for handling complex linguistic phenomena, and it involves three steps: finding clues based on input, generating reasoning steps based on the input and the generated clues, and then arriving at the final output based on the input, generated clues and reasoning steps. Experiment results showed that the results are impressive as InstructGPT with CARP prompting strategy using just 16 examples achieves SOTA results on four text classification datasets.

 \begin{table*}[h!]
\begin{center}
{\renewcommand{\arraystretch}{1.5}% for the vertical padding
\begin{tabular}{|p{0.6cm}|p{5.6cm}|p{3cm}|p{1cm}|p{1.6cm}|p{2cm}|p{0.9cm}|}
\hline

\scriptsize{\textbf{Paper}} &  \scriptsize{\textbf{Task(s)}} & \scriptsize{\textbf{GLLMs Explored}} & \scriptsize{\textbf{Prompt Settings}} & \scriptsize{\textbf{Domain(s)}} & \scriptsize{\textbf{Language(s)}} & \scriptsize{\textbf{SOTA Results}}\\ \hline

\scriptsize{\cite{gonzalez2023yes}} & \scriptsize{Entity Extraction} & \scriptsize{ChatGPT} & \scriptsize{ZS} & \scriptsize{General} & \scriptsize{English} & \scriptsize{No} \\ \hline 

\scriptsize{\cite{hu2023zero}} & \scriptsize{Entity Extraction} & \scriptsize{	GPT-3, ChatGPT} & \scriptsize{ZS} & \scriptsize{Healthcare} & \scriptsize{English} & \scriptsize{ No } \\ \hline 

\scriptsize{\cite{wei2023zero}} 	& \scriptsize{Entity Extraction, Event Extraction, Relation Extraction}	& \scriptsize{ChatGPT} & \scriptsize{ZS}	& \scriptsize{General} & \scriptsize{	English, Chinese} & \scriptsize{No} \\ \hline 

\scriptsize{\cite{gutierrez2022thinking}} & \scriptsize{Entity Extraction, Relation Classification} & \scriptsize{GPT-3} & \scriptsize{FS} & \scriptsize{Healthcare} & \scriptsize{English} & \scriptsize{No} \\ \hline 

\scriptsize{\cite{gao2023exploring}}  & \scriptsize{Event Extraction} & \scriptsize{ChatGPT} & \scriptsize{FS} & \scriptsize{General} & \scriptsize{English} & \scriptsize{No} \\ \hline 

\scriptsize{\cite{rehana2023evaluation}} & \scriptsize{Protein-Protein Interaction Extraction} & \scriptsize{GPT-3, ChatGPT and GPT-4} & \scriptsize{ZS} & \scriptsize{Healthcare} & \scriptsize{English} & \scriptsize{No} \\ \hline 

\scriptsize{\cite{yuan2023zero}} & \scriptsize{Temporal Relation Extraction} & \scriptsize{ChatGPT} & \scriptsize{ZS} & \scriptsize{General} & \scriptsize{English} & \scriptsize{No} \\ \hline 

\scriptsize{\cite{li2023evaluating}} & \scriptsize{Entity Typing, Entity Extraction, Relation Classification, Relation Extraction,  Event Detection, Event Argument Extraction, Event Extraction} & \scriptsize{ChatGPT} & \scriptsize{ZS} & \scriptsize{General} & \scriptsize{English} & \scriptsize{No} \\ \hline 

\scriptsize{\cite{chan2023chatgpt}} & \scriptsize{Temporal Relation Classification, Causal Relation Classification, Discourse Relation Classification} & \scriptsize{ChatGPT} & \scriptsize{ZS, FS} & \scriptsize{General} & \scriptsize{English} & \scriptsize{No} \\ \hline 

\scriptsize{\cite{xu2023unleash}} & \scriptsize{Relation Classification} & \scriptsize{GPT-3.5} & \scriptsize{FS} & \scriptsize{General, Scientific Literature} & \scriptsize{English} & \scriptsize{Yes} \\ \hline 

\scriptsize{\cite{wan2023gpt}} & \scriptsize{Relation Classification} & \scriptsize{GPT-3.5} & \scriptsize{FS} & \scriptsize{General, Scientific Literature} & \scriptsize{English} & \scriptsize{Yes} \\ \hline 

\scriptsize{\cite{qin2023chatgpt}} & \scriptsize{Entity Extraction} & \scriptsize{GPT-3.5, ChatGPT} & \scriptsize{ZS} & \scriptsize{General} & \scriptsize{English} & \scriptsize{No} \\ \hline 

\scriptsize{\cite{ye2023comprehensive}} & \scriptsize{Entity Extraction, Relation Extraction} & \scriptsize{	GPT-3, GPT-3.5, ChatGPT}& \scriptsize{ZS, FS} & \scriptsize{General, Social Eedia} & \scriptsize{English} & \scriptsize{No} \\ \hline 

\scriptsize{\cite{ma2023large}} & \scriptsize{Entity Extraction, Relation Extraction and Event Detection} & \scriptsize{InstructGPT} & \scriptsize{FS} & \scriptsize{General} & \scriptsize{English} & \scriptsize{Yes} \\ \hline 

\scriptsize{\cite{wang2023gpt}} & \scriptsize{Entity Extraction} & \scriptsize{GPT-3	} & \scriptsize{FS} & \scriptsize{General} & \scriptsize{English} & \scriptsize{No} \\ \hline 

\scriptsize{\cite{wang2023large}} & \scriptsize{Entity Extraction, Relation Classification} & \scriptsize{GPT-3.5, GPT-4} & \scriptsize{ZS, FS} & \scriptsize{Healthcare} & \scriptsize{English} & \scriptsize{No} \\ \hline 

\scriptsize{\cite{stammbach2022heroes}} & \scriptsize{Entity Extraction} & \scriptsize{GPT-3} & \scriptsize{ZS} & \scriptsize{General} & \scriptsize{English} & \scriptsize{No} \\ \hline 

\scriptsize{\cite{wadhwa2023revisiting}} & \scriptsize{Relation Extraction} & \scriptsize{GPT-3} & \scriptsize{FS} & \scriptsize{General, Healthcare} & \scriptsize{English} & \scriptsize{No} \\ \hline 

\scriptsize{\cite{li2023chatgpt}} & \scriptsize{Entity extraction} & \scriptsize{ChatGPT, GPT-4} & \scriptsize{FS} & \scriptsize{Finance} & \scriptsize{English} & \scriptsize{ No} \\ \hline 

\scriptsize{\cite{ li2023codeie}} & \scriptsize{Entity Extraction, Relation Extraction} & \scriptsize{GPT-3, Codex} & \scriptsize{FS} & \scriptsize{General, Scientific Literature} & \scriptsize{English} & \scriptsize{No} \\ \hline 

\scriptsize{\cite{zhang2023aligning}} & \scriptsize{Relation Classification} & \scriptsize{GPT-3.5, ChatGPT} & \scriptsize{ZS} & \scriptsize{General} & \scriptsize{English} & \scriptsize{No} \\ \hline 

\end{tabular}}
\end{center}
\caption{ \label{gllms-ie}  Summary of research works exploring GLLMs for information extraction tasks. Here ZS represents zero-shot, and FS represents few-shot.} 
\end{table*}

\subsection{Information Extraction}

\textbf{Overview.} Information Extraction (IE) in natural language processing involves extracting structured data like entities, relationships and events from unstructured text data \cite{lu2022unified}. Transforming unstructured text data into structured data enables efficient data processing, knowledge discovery, decision making and enhances information retrieval and search. Information extraction involves a number of tasks like entity typing, entity extraction, relation classification, relation extraction, event detection, event argument extraction and event extraction \cite{li2023evaluating}. Entity typing (ET) involves classifying identified named entity mentions into one of the predefined entity types \cite{chen2022learning}. Named Entity Recognition (NER) or Entity Extraction (EE) involves identifying entity mentions and then assigning them to appropriate entity types \cite{das2022container}. Relation classification (RC) involves identifying the semantic relationship between the given two target entities in a sentence \cite{wu2019enriching}. Relation Extraction (RE) involves extracting the entities and then classifying the semantic relationship between the two target entities, i.e., involves entity extraction followed by relation classification \cite{ye2022packed}. Event Detection (ED) aims to identify and categorize words or phrases that trigger events \cite{zhao2022knowledge}. Event Argument Extraction (EAE) involves identifying event arguments, i.e., entities involved in the event and then classifying their roles \cite{ma2022prompt}.  Event Extraction (EE) aims to extract both the events and the involved entities, i.e., it involves event detection followed by event argument extraction \cite{du2020event}. 

\textbf{Research works expoloring GLLMs for information extraction tasks} The recent works explored GPT-3 family large language models for various information extraction tasks like entity typing \cite{li2023evaluating}, entity extraction \cite{gonzalez2023yes, hu2023zero, wei2023zero, gutierrez2022thinking, li2023evaluating, ma2023large, wang2023gpt, wang2023large, stammbach2022heroes, li2023chatgpt, li2023codeie}, relation classification \cite{gutierrez2022thinking, li2023evaluating, chan2023chatgpt, xu2023unleash, wan2023gpt, wang2023large, zhang2023aligning}, relation extraction  \cite{wei2023zero,rehana2023evaluation, yuan2023zero, li2023evaluating, ma2023large, wadhwa2023revisiting, li2023codeie}, event classification \cite{li2023evaluating}, event argument extraction \cite{li2023evaluating} and event extraction  \cite{wei2023zero, gao2023exploring, li2023evaluating, ma2023large}. The evaluation is done in zero and few-shot settings using different prompting strategies like chain-of-thought (CoT) \cite{yuan2023zero, wan2023gpt, wang2023large, wadhwa2023revisiting}, self-verification \cite{wang2023gpt},  self-question prompting (SQP) \cite{wang2023large}, event ranking (ER) \cite{yuan2023zero} etc. Most of the research works focused on English datasets, except a few research works focused on other languages like Chinese \cite{wei2023zero}. A brief summary of research works exploring GLLMs for various information extraction tasks is presented in Table \ref{gllms-ie}.

Hu et al. \cite{hu2023zero} demonstrated the performance of ChatGPT in extracting clinical entities like problem, treatment, and test can be enhanced by including additional information about entity types like synonyms and subtypes in the prompt. Wei et al. \cite{wei2023zero} proposed ChatIE, a two-stage framework for information extraction, with each stage implemented as a multi-turn question answering. This two-stage framework helps the model break complex IE tasks into sub-tasks which allows the model to perform better. Results showed that ChatGPT used with the ChatIE framework outperforms vanilla ChatGPT by a large margin of more than 18 points. Gutierrez et al. \cite{gutierrez2022thinking} enhanced the performance of the GPT-3 model for entity extraction and relation classification by using techniques like contextual calibration \cite{zhao2021calibrate} to reduce bias and kNN-based demonstration selection. Gao et al. \cite{gao2023exploring} examined the performance of ChatGPT for event extraction in few-shot settings. The model is prompted with task descriptions, definitions of event types, positive and negative examples, and test input. The authors reported that including negative examples decreases the performance of the model, which is in line with other existing works \cite{wang2022super}. The possible reason for this is that the model misunderstands negative examples as positive examples. Rehana et al. \cite{rehana2023evaluation} explored GPT-3 family models like GPT-3, ChatGPT and GPT-4 for protein-protein interaction extraction. It is reported that including normalized protein names in the prompt enhances the performance of the model. However, fine-tuned PubMedBERT model outperforms GPT-4 model with an F1-score of 86.47. 

Yuan et al. \cite{ yuan2023zero} demonstrated that advanced prompting strategies like event ranking and chain-of-thought improve the performance of ChatGPT compared to vanilla prompting in temporal relation extraction. However, ChatGPT lags behind traditional neural networks like LSTM and fine-tuned pre-trained language models, which indicates the toughness of the temporal relation extraction task. Wang et al. \cite{wang2023large} evaluated the performances of the latest LLMs like GPT-3.5, GPT-4, and Bard models on entity extraction and relation classification in the clinical domain. Experiment results showed that GPT-4 with self-question prompting outperforms other LLMs on most of the datasets. Li et al. \cite{li2023codeie} compared  the performances of both natural language and code LLMs like GPT-3 and Codex using natural language and code style prompts. Experiment results showed that (i) Codex outperforms GPT-3 model and moderately sized fine-tuned models and (ii) Codex model with natural language or code style prompt outperforms GPT-3 model (iii) Code style prompts achieves better results in case of both Codex and GPT-3 models. The possible explanation for this is Codex which is pretrained over large volumes of code, encode structured code information which is useful for IE tasks as IE tasks involve structured outputs. Zhang et al. \cite{zhang2023aligning} proposed the QA4RE framework, which frames relation extraction as a question-answering problem. In the QA4RE framework, the sentence serves as context, and the relation types serve as options from which the LLMs choose. Experiment results showed that the proposed approach improves the performance of ChatGPT and GPT-3.5 models by a good margin in relation extraction.

Some of the research works \cite{xu2023unleash, wan2023gpt, ma2023large} demonstrated that GPT-3 family models can achieve SOTA results in information extraction tasks. Wan et al. \cite{wan2023gpt} achieved SOTA results in relation extraction with the GPT-RE framework. GPT-RE framework overcomes the drawbacks in existing works using entity-aware demonstration retrieval based on fine-tuned model and gold label-induced reasoning. The use of representations from fine-tuned relation model for demonstration selection is more effective as they naturally include entity and relation information. Ma et al. \cite{ma2023large}  proposed a “filter then rerank” approach to use both fine-tuned models and LLMs to take advantage of the strengths of both models for few-shot information extraction. Here fine-tuned model acts as a filter while LLM acts as a re-ranker. The proposed approach achieves SOTA results with an average improvement of over 2 points in the F1 score. 

\subsection{Question Answering}
\textbf{Overview.} Question Answering (QA) is an important natural language processing task which deals with the development of algorithms to understand and interpret user queries in natural language and then deliver accurate responses \cite{zaib2022conversational, chali2011improving}. The main aim of question answering systems is to enhance human-computer interaction, i.e., QA systems avoid the use of complex commands and allow the user to interact with machines in a more natural way through natural language queries. For example, popular AI assistants like Amazon Alexa\footnote{https://alexa.amazon.com}, Google Assistant\footnote{https://assistant.google.com} and Apple Siri\footnote{https://www.apple.com/in/siri/} rely on QA to provide accurate answers to user queries. The option of interaction through natural language queries enhances the reach of technology to a broader audience.  QA can be treated as a fine-grained version of information retrieval \cite{torfi2020natural}, and the demand for QA systems is increasing day by day because of the ability to generate answers which are accurate, relevant and short. 

 \begin{table*}[h!]
\begin{center}
{\renewcommand{\arraystretch}{1.5}% for the vertical padding
\begin{tabular}{|p{0.6cm}|p{5.6cm}|p{3cm}|p{1cm}|p{1.6cm}|p{2cm}|p{0.9cm}|}
\hline

\scriptsize{\textbf{Paper}} &  \scriptsize{\textbf{Task(s)}} & \scriptsize{\textbf{GLLMs Explored}} & \scriptsize{\textbf{Prompt Settings}} & \scriptsize{\textbf{Domain(s)}} & \scriptsize{\textbf{Language(s)}} & \scriptsize{\textbf{SOTA Results}}\\ \hline

\scriptsize{\cite{nunes2023evaluating}} &  \scriptsize{	Admission Exam Question Answering} &  \scriptsize{	GPT-3.5, ChatGPT, GPT-4} &  \scriptsize{	ZS, FS} &  \scriptsize{Education} &  \scriptsize{Brazilian Portuguese} &  \scriptsize{No} \\ \hline 

\scriptsize{\cite{tan2023evaluation}} &  \scriptsize{Knowledge-based Complex Question Answering} &  \scriptsize{GPT-3, GPT-3.5, ChatGPT} &  \scriptsize{ZS} &  \scriptsize{General} &  \scriptsize{Multiple languages} &  \scriptsize{No} \\ \hline 

\scriptsize{\cite{yang2022empirical}} &  \scriptsize{	Knowledge-based Visual Question Answering} &  \scriptsize{GPT-3} &  \scriptsize{	ZS} &  \scriptsize{General} &  \scriptsize{English} &  \scriptsize{Yes}  \\ \hline 

\scriptsize{\cite{ srivastava2022towards}} &  \scriptsize{	Tabular Question Answering} &  \scriptsize{GPT-3} &  \scriptsize{ZS, FS} &  \scriptsize{News} &  \scriptsize{English} &  \scriptsize{ No} \\ \hline 

 \scriptsize{\cite{ zheng2023does}} &  \scriptsize{Open Domain Question Answering} &  \scriptsize{ChatGPT} &  \scriptsize{ZS} &  \scriptsize{General} &  \scriptsize{English} &  \scriptsize{No} \\ \hline 

\scriptsize{\cite{samaan2023assessing}} &  \scriptsize{	Bariatric Surgery Question Answering} &  \scriptsize{ChatGPT} &  \scriptsize{ZS} &  \scriptsize{Healthcare} &  \scriptsize{English} &  \scriptsize{No} \\ \hline 

\scriptsize{\cite{holmes13evaluating}} &  \scriptsize{Radiation Oncology Physics Question Answering} &  \scriptsize{ChatGPT, GPT-4} &  \scriptsize{ZS} &  \scriptsize{Healthcare} &  \scriptsize{English} &  \scriptsize{No} \\ \hline 

 \scriptsize{\cite{ joshi2023chatgpt}} &  \scriptsize{Computer Science Question Answering} &  \scriptsize{ChatGPT} &  \scriptsize{ZS} &  \scriptsize{Education} &  \scriptsize{English} &  \scriptsize{No} \\ \hline 

\scriptsize{\cite{nori2023capabilities}} &  \scriptsize{Medical Question Answering} &  \scriptsize{GPT-3.5, GPT-4} &  \scriptsize{ZS, FS} &  \scriptsize{Healthcare} &  \scriptsize{English} &  \scriptsize{No} \\ \hline 

\scriptsize{\cite{hamidi2023evaluation}} &  \scriptsize{Patient-specific Question Answering} &  \scriptsize{ChatGPT} &  \scriptsize{	ZS} &  \scriptsize{Healthcare} &  \scriptsize{English} &  \scriptsize{No} \\ \hline 

\scriptsize{\cite{bang2023multitask}} &  \scriptsize{Question Answering} &  \scriptsize{ChatGPT} &  \scriptsize{ZS} &  \scriptsize{General} &  \scriptsize{English} &  \scriptsize{Yes} \\ \hline 

\scriptsize{\cite{qin2023chatgpt}} &  \scriptsize{Boolean Question Answering} &  \scriptsize{ChatGPT} &  \scriptsize{ZS} &  \scriptsize{General} &  \scriptsize{English} &  \scriptsize{No} \\ \hline 

\scriptsize{\cite{kocon2023chatgpt}} &  \scriptsize{Multiple Choice Question Answering} &  \scriptsize{ChatGPT} &  \scriptsize{ZS} &  \scriptsize{General, Social Media} &  \scriptsize{English} &  \scriptsize{No} \\ \hline 

\scriptsize{\cite{ye2023comprehensive}} &  \scriptsize{Question Answering} &  \scriptsize{	GPT-3, GPT-3.5, ChatGPT} &  \scriptsize{ZS, FS} &  \scriptsize{General} &  \scriptsize{English} &  \scriptsize{No} \\ \hline 

\scriptsize{\cite{savelka2023large}} &  \scriptsize{Multiple Choice Code Question Answering} &  \scriptsize{GPT-3.5} &  \scriptsize{ZS} &  \scriptsize{Coding} &  \scriptsize{English} &  \scriptsize{No} \\ \hline 

\scriptsize{\cite{bommarito2022gpt}} &  \scriptsize{Bar Exam  Question Answering} &  \scriptsize{GPT-3.5} &  \scriptsize{ZS} &  \scriptsize{Legal} &  \scriptsize{English} &  \scriptsize{No} \\ \hline 

\scriptsize{\cite{pereira2023visconde}} &  \scriptsize{Multi-Document Question Answering} &  \scriptsize{GPT-3.5} &  \scriptsize{ FS} &  \scriptsize{General, Scientific Literature} &  \scriptsize{English} &  \scriptsize{No} \\ \hline

\scriptsize{\cite{gupta2023performance}} &  \scriptsize{Plastic Survey Exam Question Answering} &  \scriptsize{ChatGPT} &  \scriptsize{ZS} &  \scriptsize{Healthcare} &  \scriptsize{English} &  \scriptsize{No} \\ \hline 

\scriptsize{\cite{tanaka2023performance}} &  \scriptsize{Japanese Medical Exam Question Answering} &  \scriptsize{GPT-3.5, GPT-4} &  \scriptsize{FS} &  \scriptsize{Healthcare} &  \scriptsize{Japanese} &  \scriptsize{No} \\ \hline 

\scriptsize{\cite{ li2023chatgpt}} &  \scriptsize{Financial Question Answering} &  \scriptsize{ChatGPT, GPT-4} &  \scriptsize{	ZS} &  \scriptsize{Finance} &  \scriptsize{English} &  \scriptsize{No} \\ \hline 

\scriptsize{\cite{wang2023large}} &  \scriptsize{Medical Question Answering} &  \scriptsize{GPT-3.5, GPT4} &  \scriptsize{ZS, FS} &  \scriptsize{Healthcare} &  \scriptsize{English} &  \scriptsize{No} \\ \hline 

\scriptsize{\cite{robinson2022leveraging}} &  \scriptsize{Multiple Choice Question Answering} &  \scriptsize{GPT-3, Codex, InstructGPT} &  \scriptsize{ZS} &  \scriptsize{General} &  \scriptsize{English}&  \scriptsize{No} \\ \hline 

\scriptsize{\cite{weng2023large}} &  \scriptsize{Medical Conversational Question Answering} &  \scriptsize{GPT-3, InstructGPT} &  \scriptsize{ZS} &  \scriptsize{Healthcare} &  \scriptsize{English, Chinese}&  \scriptsize{No} \\ \hline 

\scriptsize{\cite{lin2022truthfulqa}} &  \scriptsize{Question Answering} &  \scriptsize{GPT-3} &  \scriptsize{ZS} &  \scriptsize{Multiple domains including Legal and Health} &  \scriptsize{English} &  \scriptsize{No} \\ \hline 

\scriptsize{\cite{kasai2023evaluating}} &  \scriptsize{Japanese Medical Exam Question Answering} &  \scriptsize{GPT-3, ChatGPT, GPT-4} &  \scriptsize{FS} &  \scriptsize{Healthcare} &  \scriptsize{Japanese} &  \scriptsize{No} \\ \hline

\end{tabular}}
\end{center}
\caption{ \label{gllms-qa}  Summary of research works exploring GLLMs for question answering tasks. Here ZS represents zero-shot, and FS represents few-shot.} 
\end{table*}
	
\textbf{Research works exploring GLLMs for question answering tasks.} The NLP research community explored GLLMs for question answering in various domains like education \cite{ nunes2023evaluating, joshi2023chatgpt}, news \cite{srivastava2022towards}, healthcare \cite{samaan2023assessing, holmes13evaluating, nori2023capabilities, hamidi2023evaluation, gupta2023performance, tanaka2023performance, wang2023large, weng2023large,kasai2023evaluating }, social media \cite{ye2023comprehensive}, coding \cite{savelka2023large}, legal \cite{bommarito2022gpt, lin2022truthfulqa}, finance \cite{li2023chatgpt} and scientific literature \cite{pereira2023visconde}. Most of the research works focused on the English language, except a few research works focusing on languages like Portuguese \cite{nunes2023evaluating}, Japanese \cite{tanaka2023performance, kasai2023evaluating} and Chinese \cite{weng2023large}. As advanced prompting methods allow GLLMs to perform well, some of the research works investigated the effectiveness of advanced prompting strategies like chain-of-thought \cite{nunes2023evaluating, tan2023evaluation, holmes13evaluating, pereira2023visconde, wang2023large, kasai2023evaluating}, self-question prompting \cite{wang2023large, weng2023large} and holistically thought \cite{weng2023large} for question answering. Table \ref{gllms-qa} presents a summary of research works exploring GLLMs for question answering across various domains and languages. 

Zheng et al. \cite{zheng2023does} studied the shortcomings of ChatGPT in answering complex open-domain questions and found errors related to understanding, factual accuracy, specificity, and logical reasoning. They also analyzed the importance of knowledge memorization, recall, and reasoning abilities in addressing these failures. The authors demonstrated that providing the model with external knowledge, cues for knowledge recall, and guidance for logical reasoning can enhance its ability to provide more accurate answers. Samaan et al. \cite{samaan2023assessing} examined the accuracy of ChatGPT in answering questions related to Bariatric surgery. The authors reported that ChatGPT correctly answered 131 questions from 151 questions, i.e., ChatGPT achieves an accuracy of 86.8\%. The impressive performance of ChatGPT shows that it can serve as an additional information resource in addition to healthcare professionals and reduce their burden in answering patient questions.  Holmes et al. \cite{holmes13evaluating} compared the performances of GLLMs like ChatGPT, GPT-4 with other LLMs like Bard, BLOOMZ and medical physicists in answering related questions to Radiation Oncology Physics. The performance of GPT-4 is very impressive as the model outperforms medical physicists and other LLMs like ChatGPT, Bard and BLOOMZ. The performance of GPT-4 is further enhanced using CoT prompting, i.e., the model is prompted to arrive at the answer after step-by-step reasoning. Nori et al. \cite{nori2023capabilities} performed a comprehensive evaluation of the GPT-4 model on medical question answering in zero and few-shot settings. For evaluation, the authors used six datasets: two related to the United States Medical License Examination (USMLE) exam and four from the MultiMedQA benchmark \cite{singhal2023large}. The performance of GPT-4 is very impressive as it outperforms not only general LLM like GPT-3.5 but also medical domain-specific LLM like Med-PaLM \cite{singhal2023large}. Moreover, on USMLE exam datasets, GPT-4 model score is 20 points more than the passing score. 

Hamidi et al. \cite{hamidi2023evaluation} evaluated ChatGPT and Claude in answering patient-specific medical questions from MIMIC-III clinical notes. Experiment results demonstrated that the performances of both models are promising as these models display significant levels of coherence, accuracy, coverage and relevance in their answers. Li  et al. \cite{li2023chatgpt} demonstrated that GPT4 achieves the best results for question answering in the finance domain  and outperforms ChatGPT, domain-specific models like BloombergGPT, FinQANet and general LLMs like OPT (66B), and BLOOM (176B). Although the performance of GLLMs is impressive in zero and few-shot settings in multiple choice question answering, these models still lag behind SOTA results. The main reason for this is the use of cloze prompts. In cloze prompts, the model is prompted with only question without answer options, so the model generates the answers just by conditioning on the question. Robinson et al. \cite{robinson2022leveraging} proposed a new prompting strategy called multiple choice prompt which prompts the model with question and answer options so that the model generates the answer by conditioning on both question and answer options. Evaluation on 20 datasets showed that multiple-choice prompt helps GLLMs to achieve near SOTA results. 

Some of the research works explored the effectiveness of GLLMs in answering exam questions from various domains. Nunes et al. \cite{nunes2023evaluating} investigated the performances of GLLMs like GPT-3.5, ChatGPT and GPT-4 in answering questions from the Brazilian university admission exam. Here all the questions are in Brazilian Portuguese language. The authors explored different prompting strategies like vanilla (zero-shot and few-shot) and CoT (few-shot). The authors observed that GPT-4 outperforms all other models by a large margin of over 11 points and achieves the best results with CoT prompting in few-shot settings. Joshi et al. \cite{joshi2023chatgpt} evaluated ChatGPT in answering undergraduate-level computer science exam questions. For the evaluation, the authors gathered (i) questions from various computer science subjects like data structures, operating systems, machine learning and database management systems, (ii) questions from the GATE exam and (iii) programming questions from the Leetcode website. The results showed that ChatGPT is inconsistent in answering the questions, so students are not advised to rely on ChatGPT completely for their assignments and exams. Bommarito et al. \cite{bommarito2022gpt} examined the ability of OpenAI's text-davinci-003 (GPT-3.5) model in answering multiple choice questions from the Bar Exam. Interestingly, human participants with extensive education and specialized training achieved a 68\% accuracy rate, while the GPT-3.5 model achieved a lower accuracy rate of 50.3\%. Gupta et al. \cite{gupta2023performance} evaluated how effective ChatGPT is in answering questions from plastic surgery inservice training examination. The authors reported that ChatGPT achieves an accuracy of 54.96\% by correctly answering 242 questions. Tanaka et al. \cite{tanaka2023performance} evaluated the performances of GLLMs like GPT-3.5 and GPT-4 in answering questions from the Japanese National Medical Licensing Examination (NMLE). Here the input includes sample examples, instructions to translate the question into English, and then summarizing the question before answering. The authors reported that GPT-4 achieves a score better than the minimum passing score, and further analysis showed that the incorrect answers are due to insufficient medical knowledge and insufficient information about the Japanese-specific medical system. Kasai et al. \cite{kasai2023evaluating} reported that GPT-4 outperforms other models and passes the Japanese national medical licensing exam in the last six years. Moreover, ChatGPT with English-translated prompts achieves better results than ChatGPT with Japanese prompts. This is because ChatGPT is predominantly trained over the English text corpus. 

\begin{table*}[h!]
\begin{center}
{\renewcommand{\arraystretch}{1.5}% for the vertical padding
\begin{tabular}{|p{0.6cm}|p{2.2cm}|p{2cm}|p{3.5cm}|p{3cm}|p{1.5cm}|p{2cm}|}
\hline

\scriptsize{\textbf{Paper}} & \scriptsize{\textbf{GLLMs Explored}} & \scriptsize{\textbf{Prompt Settings}} & \scriptsize{\textbf{Domain(s)}} & \scriptsize{\textbf{Language(s)}} & \scriptsize{\textbf{Granularity}}  & \scriptsize{\textbf{Outperforms Commercial Systems}} \\ \hline

\scriptsize{\cite{gu2023linguistically}} & \scriptsize{ChatGPT} & \scriptsize{ZS} & \scriptsize{General} & \scriptsize{	Japanese, Chinese} & \scriptsize{Sentence} & \scriptsize{No} \\ \hline 

\scriptsize{\cite{peng2023towards}} & \scriptsize{ChatGPT} & \scriptsize{	ZS} & \scriptsize{General, News, Healthcare} & \scriptsize{English, Chinese, German, Romanian} & \scriptsize{Sentence} & \scriptsize{No} \\ \hline 

\scriptsize{\cite{jiao2023chatgpt}} & \scriptsize{ChatGPT, GPT-4} & \scriptsize{ZS} & \scriptsize{General, Healthcare ,Social Media} & \scriptsize{English, Chinese, German, Romanian} & \scriptsize{Sentence} & \scriptsize{Yes} \\ \hline 

\scriptsize{\cite{hendy2023good}} & \scriptsize{InstructGPT, ChatGPT, GPT-4} & \scriptsize{ZS,FS} & \scriptsize{News, Social Media, E-Commerce, Dialogue} & \scriptsize{English, German, Chinese} & \scriptsize{Sentence, Document} & \scriptsize{Yes} \\ \hline 

\scriptsize{\cite{gao2023design}} & \scriptsize{ChatGPT} & \scriptsize{ZS, FS} & \scriptsize{General, News, Social Media, Dialogue, E-Commerce} & \scriptsize{English, French, Spanish} & \scriptsize{Sentence} & \scriptsize{Yes}  \\ \hline 

\scriptsize{\cite{wang2023document}} & \scriptsize{ChatGPT, GPT-4} & \scriptsize{ZS} & \scriptsize{General, Social Media, News, Dialogue} & \scriptsize{English, German, Russian} & \scriptsize{Document} & \scriptsize{Yes} \\ \hline 

\scriptsize{\cite{zhu2023multilingual}} & \scriptsize{ChatGPT} & \scriptsize{ZS, FS} & \scriptsize{General} & \scriptsize{102 Languages in 202 directions} & \scriptsize{ Sentence} & \scriptsize{No}  \\ \hline 

\scriptsize{\cite{lyu2023new}} & \scriptsize{ChatGPT} & \scriptsize{ZS} & \scriptsize{General} & \scriptsize{English, Chinese, French} & \scriptsize{Paragraph} & \scriptsize{No}  \\ \hline 

\scriptsize{\cite{bang2023multitask}} & \scriptsize{ChatGPT} & \scriptsize{ZS} & \scriptsize{General} & \scriptsize{	Twelve languages, including four low-resource languages} & \scriptsize{ Sentence} & \scriptsize{No}  \\ \hline 

\scriptsize{\cite{karpinska2023large}} & \scriptsize{GPT-3.5} & \scriptsize{ZS} & \scriptsize{General} 	& \scriptsize{18 language Pairs, including Japanese, English and Polish} & \scriptsize{Sentence, Paragraph} & \scriptsize{Yes}  \\ \hline  

\scriptsize{\cite{moslem2023adaptive}} & \scriptsize{GPT-3.5} & \scriptsize{ZS, FS} & \scriptsize{General} & \scriptsize{English, Arabic, Chinese, German, Spanish} & \scriptsize{Sentence} & \scriptsize{Yes} \\ \hline 

 \scriptsize{\cite{he2023exploring}} & \scriptsize{GPT-3.5} & \scriptsize{ZS, FS} & \scriptsize{General} & \scriptsize{English, Chinese, Japanese, German, French} & \scriptsize{Sentence} & \scriptsize{No}  \\ \hline 

\scriptsize{\cite{raunak2023leveraging}} & \scriptsize{GPT-3.5, GPT-4} & \scriptsize{ZS} & \scriptsize{General} & \scriptsize{	English, German, Chinese} & \scriptsize{Sentence} & \scriptsize{Yes} \\ \hline 

\scriptsize{\cite{raunak2023gpts}} & \scriptsize{GPT-3.5} & \scriptsize{ZS} & \scriptsize{General} & \scriptsize{English, German, Russian} & \scriptsize{Sentence} & \scriptsize{Yes} \\ \hline 

\end{tabular}}
\end{center}
\caption{ \label{gllms-mt}  Summary of research works exploring GLLMs for machine translation. Here ZS represents zero-shot, and FS represents few-shot.} 
\end{table*}

Some of the research works explored GLLMs for more challenging tasks in question answering like  tabular question answering \cite{srivastava2022towards}, knowledge-based complex question answering \cite{tan2023evaluation}, multiple choice code question answering \cite{savelka2023large}, multi-document question answering \cite{pereira2023visconde} and conversational question answering  \cite{weng2023large}. Srivastava et al. \cite{srivastava2022towards} evaluated the effectiveness of GPT-3 for question answering on tabular data in zero and few-shot settings. Here the model is prompted with unstructured passage text, tabular data in JSON format, examples (in the case of few-shot) and the question. The authors reported that GPT-3 displayed its ability to successfully locate the table, comprehend its structure, and accurately access the relevant cells or passages of text in order to provide answers to the given questions. Savelka et al. \cite{savelka2023large} evaluated the effectiveness of GPT-3.5 models in answering multiple-choice questions (MCQs), particularly those involving code snippets from programming courses. Experiment results showed that MCQs with code snippets have lower success rates compared to those without code, indicating a challenge in answering multiple-choice questions with code snippets.  Pereira et al.\cite{pereira2023visconde}  presented Visconde, a novel framework based on the GPT-3.5 model to tackle multi-document question answering. Visconde follows a three-step process involving decomposition, retrieval, and aggregation. The decomposition phase uses the GPT-3.5 model in few-shot settings for question simplification, the retrieval stage uses the SOTA model to select the relevant text chunks, and the final aggregation phase uses the GPT-3.5 with few-shot CoT prompting to get the answer. The authors observed that CoT prompting, i.e., generating reasoning steps before generating the final answer, enhances the performance. Weng et al. \cite{weng2023large} enhanced the performance of GLLMs in answering medical conversational questions in English and Chinese using a novel prompt strategy called Holistically Thought (HoT). The HoT prompting strategy involves diffused thinking and focused thinking strategies to generate high-quality responses. Diffused thinking helps to generate various responses through diversified decoding, focused thinking generates a concise medical summary based on the dialogues and the final response is generated based on the dialogues, outputs of diffused thinking and focused thinking.

Unlike all the above discussed research works where the performances of GLLMs are just satisfactory but not SOTA, some of the research works \cite{yang2022empirical, bang2023multitask} demonstrated that it is possible to achieve SOTA results for question answering task using GLLMs. For example, Yang et al. \cite{yang2022empirical}  explored GPT-3 model for knowledge-based visual question answering. Knowledge-based visual question answering involves answering questions which require information which is not available in the input images. The authors propose a novel approach which uses GPT-3 as a knowledge source which is implicit and unstructured. Experiment results showed that the proposed approach achieves new SOTA results by outperforming existing approaches with a large margin of over 8 points.

\subsection{Machine Translation}
\textbf{Overview. } Machine Translation (MT), an important task of natural language processing, deals with the development of models which can translate input text from the source language to the target language \cite{stahlberg2020neural, Yang2020ASO, Tan2020NeuralMT}. MT models receive the input text in the source language, understand the syntax and semantics of the input text and then generate the translation in the target language. So, a good machine translation model should possess strong natural language understanding and generation skills to generate quality translations. The main objective of MT systems is to enhance cross-lingual communication by reducing the gap between individuals from different linguistic communities.  The evolution of MT systems started with rule-based models followed by statistical and neural models \cite{Tan2020NeuralMT}. Rule-based MT systems are built on top of manually crafted syntactic and grammatical rules. As manually framing rules is heavily laborious and expensive, these systems are later replaced by statistical MT systems. Statistical MT systems use statistical models trained on bilingual data. With the evolution of deep learning models, the research community started to build neural machine translation (NMT) systems with the help of neural models \cite{sutskever2014sequence, Bahdanau2014NeuralMT, luong2015effective}. These neural models are essentially based on the encoder-decoder architecture, where the encoder understands the input sequence and encodes it into a vector, and the decoder, based on the encoder output, generates the output sequence auto-regressively. Some of the recent  neural models used for translation are mBART-50 \cite{tang2020multilingual}, M2M100 \cite{Fan2020BeyondEM}, NLLB200 \cite{costa2022no} etc.

\textbf{Research works exploring GLLMs for machine translation.} In recent times, GLLMs like ChatGPT and GPT-4 demonstrated remarkable performances in both natural language understanding and generation tasks. A good machine translation system requires strong natural language understanding and generation skills. As ChatGPT and GPT-4 possess strong natural language understanding and generation skills, the research community investigated the effectiveness of these models for machine translation across various domains like news \cite{peng2023towards, hendy2023good, gao2023design, wang2023document}, healthcare \cite{peng2023towards, jiao2023chatgpt}, social media \cite{jiao2023chatgpt, hendy2023good, gao2023design, wang2023document}, dialogue \cite{hendy2023good, wang2023document,gao2023design } and e-commerce \cite{hendy2023good, gao2023design}. Most of the research works focused on sentence-level machine translation \cite{gu2023linguistically, peng2023towards, jiao2023chatgpt, hendy2023good, gao2023design, zhu2023multilingual, bang2023multitask, karpinska2023large, moslem2023adaptive, he2023exploring, raunak2023leveraging, raunak2023gpts}, except a few research works focused on paragraph-level machine translation \cite{lyu2023new, karpinska2023large}  and document-level machine translation \cite{hendy2023good, wang2023document}. As advanced prompting methods allow GLLMs to perform well, some of the research works investigated the effectiveness of advanced prompting strategies like pivot \cite{jiao2023chatgpt}, chain-of-thought \cite{raunak2023leveraging} and multi-aspect prompting and selection \cite{he2023exploring}. Table \ref{gllms-mt} presents a summary of research works exploring GLLMs for machine translation across various domains and languages. 

Gu et al. \cite{gu2023linguistically} proposed a novel approach based on ChatGPT to enhance the quality of translation from Japanese to Chinese by effectively handling attribute clauses using a pre-edit scheme. The proposed approach, which integrates the pre-edit scheme with a novel two-step prompting strategy, enhances the translation quality by more than 35\%.  Peng et al. \cite{peng2023towards} explored the impact of temperature, task and domain information on the translation performance of ChatGPT. The authors showed that (i) ChatGPT performance degrades with an increase in temperature, and hence it is recommended to use a lower temperature (recommended is 0). and (ii) including task and domain information in the prompt enhances the performance of ChatGPT consistently for both high and low language translations. Zhu et al. \cite{zhu2023multilingual} evaluated the performance of ChatGPT and other LLMs like OPT, BLOOM and XGLM on 102 languages in 202 translation directions. The authors reported that ChatGPT comprehensively outperforms other LLMs but still lags behind neural machine translation models like NLLB in the majority of the translation directions. Further analysis showed three errors, namely hallucination, monotonic translation and off-target translation.  Lyu et al. \cite{lyu2023new} presented some interesting research directions with respect to using LLMs for machine translation. The presented interesting research directions include stylized machine translation, interactive machine translation and translation memory-based machine translation. Neural machine translation systems just focus on source-target text mapping, which results in a lot of errors. Unlike neural machine translation systems, the human translation process involves intermediate steps to ensure high translation quality. Inspired by the human translation process, He et al. \cite{he2023exploring}  proposed MAPS, which involves three steps: knowledge mining, knowledge integration and knowledge selection to generate quality translations. Extension evaluation of the WMT22 test set shows that MAPS improves the performance of models like GPT-3.5 and Alpaca and also addresses the hallucination issue by resolving 59\% of hallucination errors. 

 \begin{table*}[h!]
\begin{center}
{\renewcommand{\arraystretch}{1.5}% for the vertical padding
\begin{tabular}{|p{0.6cm}|p{3cm}|p{1cm}|p{3cm}|p{2cm}|p{0.9cm}|}
\hline

\scriptsize{\textbf{Paper}} & \scriptsize{\textbf{GLLMs Explored}} & \scriptsize{\textbf{Prompt Settings}} & \scriptsize{\textbf{Domain(s)}} & \scriptsize{\textbf{Language(s)}} & \scriptsize{\textbf{SOTA Results}}\\ \hline

\scriptsize{\cite{martinez2023chatgpt}} & \scriptsize{ ChatGPT} & \scriptsize{ZS} & \scriptsize{News, Scientific Literature} & \scriptsize{English} & \scriptsize{Yes} \\ \hline 

\scriptsize{\cite{song2023chatgpt} } & \scriptsize{ ChatGPT} & \scriptsize{ZS} & \scriptsize{Scientific Literature} & \scriptsize{English} & \scriptsize{No} \\ \hline 

\end{tabular}}
\end{center}
\caption{ \label{gllms-kpg}  Summary of research works exploring GLLMs for keyphrase generation task. Here ZS represents zero-shot, and FS represents few-shot.} 
\end{table*}

In all the above discussed research works, the performances of GLLMs are just satisfactory but not on par or beyond the performances of commercial machine translation systems. Some of the research works \cite{jiao2023chatgpt, hendy2023good, gao2023design,wang2023document, karpinska2023large, moslem2023adaptive, raunak2023leveraging, raunak2023gpts} showed that it is possible to outperform commercial machine translation systems using GLLMs. For example, Jiao et al. \cite{jiao2023chatgpt} investigated the translation capabilities of GLLMs like ChatGPT and GPT-4  and compared the performance with commercial systems like Google Translate, DeepL Translate and Tencent TranSmart. Extensive evaluation of multiple datasets showed that (i) the performance of GLLMs is on par with commercial systems in the case of high resources languages only, and (ii) the translation quality of low-resource languages can be enhanced using a novel pivot prompting strategy, which involves translating into high resource language before translating into the target low resource language. The naive prompts are unable to elicit the translation ability of ChatGPT fully. So, Gao et al. \cite{gao2023design}  focused on developing advanced prompting strategies by including additional information like task information, domain information and syntactic information like PoS (parts of speech) tags. The authors showed that ChatGPT, with the proposed advanced prompting strategy, achieves promising results and even outperforms commercial systems like Google Translate and DeepL Translate.  Wang et al. \cite{wang2023document} examined the performances of ChatGPT and GPT-4 for document-level machine translation and also compared the results with commercial systems from Google, DeepL and Tencent. The authors reported that GLLMs do well when the sentences in the document are combined and given at once to the model. Moreover, with this prompting strategy, both the GLLMs exhibit better performances than commercial machine translation systems according to human evaluation and also outperform most document-level neural machine translation methods in terms of d-BLEU scores. Karpinska et al. \cite{karpinska2023large} explored the GPT-3.5 model for paragraph-level machine translation. The authors experimented with three different prompting strategies, namely translating sentence by sentence in isolation, translating sentence by sentence in the presence of the rest of the paragraph and translating the entire paragraph at once. After extensive evaluation of 18 language pairs, including English and Japanese, the authors report that translating the entire paragraph at once outperforms other strategies and commercial systems like Google Translate. Raunak et al. \cite{raunak2023gpts} examined the differences between the translations generated by GLLMs like GPT-3.5 and NMT systems like Microsoft Translator. The authors reported that GLLM generated translations are less literal, with better scores.

 \begin{table*}[h!]
\begin{center}
{\renewcommand{\arraystretch}{1.5}% for the vertical padding
\begin{tabular}{|p{0.6cm}|p{5.5cm}|p{2cm}|p{1cm}|p{1.8cm}|p{1.3cm}|}
\hline

\scriptsize{\textbf{Paper}} & \scriptsize{\textbf{Task(s)}} & \scriptsize{\textbf{GLLMs Explored}} & \scriptsize{\textbf{Prompt Settings}} & \scriptsize{\textbf{Domain(s)}} & \scriptsize{\textbf{Language(s)}} \\ \hline

\scriptsize{\cite{pan2023preliminary}} & \scriptsize{Spoken Language Understanding and Dialogue State Tracking} & \scriptsize{GPT-3.5, ChatGPT} & \scriptsize{ZS} & \scriptsize{General} & \scriptsize{English}  \\ \hline 

\scriptsize{\cite{zhao2023chatgpt}} & \scriptsize{Emotion Dialogue Understanding and Generation Tasks} & \scriptsize{ChatGPT} & \scriptsize{ZS, FS} & \scriptsize{General} & \scriptsize{English} \\ \hline 

\scriptsize{\cite{chintagunta2021medically}} & \scriptsize{Dialogue Summarization} & \scriptsize{GPT-3} & \scriptsize{ZS} & \scriptsize{Healthcare} & \scriptsize{English} \\ \hline 

\scriptsize{\cite{bang2023multitask}} & \scriptsize{Dialogue Generation} & \scriptsize{ChatGPT} & \scriptsize{ZS} & \scriptsize{General} & \scriptsize{English} \\ \hline 

\scriptsize{\cite{qin2023chatgpt}} & \scriptsize{Dialogue Summarization} & \scriptsize{ChatGPT} & \scriptsize{ZS} & \scriptsize{General} & \scriptsize{English}  \\ \hline 

\scriptsize{\cite{prodan2022prompt}} & \scriptsize{Dialogue Summarization} & \scriptsize{GPT-3} & \scriptsize{FS} & \scriptsize{General} & \scriptsize{English} \\ \hline 

\scriptsize{\cite{huynh2023understanding}} & \scriptsize{Dialog Evaluation} & \scriptsize{GPT-3} & \scriptsize{FS} & \scriptsize{General} & \scriptsize{English} \\ \hline 

\scriptsize{\cite{fan2023uncovering}} & \scriptsize{Dialogue Discourse Analysis} & \scriptsize{ChatGPT} & \scriptsize{ZS, FS} & \scriptsize{General} & \scriptsize{English, Chinese} \\ \hline 

\scriptsize{\cite{wang2023chain}} & \scriptsize{Dialogue Question Answering} & \scriptsize{ChatGPT} & \scriptsize{ZS, FS} & \scriptsize{General} & \scriptsize{English, Chinese} \\ \hline

\end{tabular}}
\end{center}
\caption{ \label{gllms-dt}  Summary of research works exploring GLLMs for various dialogue tasks. Here ZS represents zero-shot, and FS represents few-shot.} 
\end{table*}

\subsection{Keyphrase Generation}
\textbf{Overview.} Keyphrase generation (KPG) involves generating a set of phrases that capture the main ideas of a document \cite{meng2021empirical}. The primary advantage of KPG over keyphrase extraction is the ability to generate both extractive and abstractive keyphrases. Keyphrase generation is approached as a sequence-to-sequence generation task \cite{sutskever2014sequence, yuan2020one, kulkarni-etal-2022-learning} in the existing works.  The current state-of-the-art model for keyphrase generation is, KeyBART \cite{kulkarni-etal-2022-learning}, which is based on BART and trained using the text-to-text generation paradigm. Table \ref{gllms-kpg} presents a summary of research works exploring GLLMs for keyphrase generation.

\textbf{Research works exploring GLLMs for keyphrase generation.} Martinez et al. \cite{martinez2023chatgpt} performed a  comprehensive evaluation of ChatGPT as a keyphrase generator by evaluating its performance on six datasets using six candidate prompts. The authors reported that the results are promising, but ChatGPT struggles in the case of generating absent keyphrases. Song et al. \cite{song2023chatgpt} evaluated ChatGPT on multiple datasets from news and scientific literature domains having both short and long documents. Experiment results showed that ChatGPT outperforms KeyBART \cite{kulkarni-etal-2022-learning}, the SOTA model, on all the datasets.

\subsection{Dialogue Tasks}
\textbf{Overview.} Dialogue tasks in natural language processing (NLP) deal with understanding and generating human-like conversations between machines and users \cite{serban2018survey}. The main objective of these tasks is to enable machines to have conversations with humans in a natural way. These dialogue tasks are essential components of building effective conversational agents, which have a wide range of applications, including customer support \cite{serban2018survey, larson2022survey}.
\\

\textbf{Research works exploring GLLMs for dialogue tasks.} The research community explored GLLMs like GPT-3, GPT-3.5 and ChatGPT for various dialogue tasks like dialogue summarization \cite{chintagunta2021medically, qin2023chatgpt, prodan2022prompt} , dialogue question answering \cite{wang2023chain}, emotion dialogue understanding and generation \cite{zhao2023chatgpt}, dialogue state tracking \cite{pan2023preliminary}, dialogue generation \cite{bang2023multitask}, and dialogue discourse analysis \cite{fan2023uncovering}. Some of the research works explored LLMs for the evaluation of dialogue tasks \cite{huynh2023understanding}. Most of the research works focused on general domain and English language datasets, except a few research works which focused on the medical domain \cite{chintagunta2021medically} and languages like Chinese \cite{fan2023uncovering, wang2023chain}. Table \ref{gllms-dt} presents a summary of research works exploring GLLMs for various dialogue tasks.

Pan et al. \cite{pan2023preliminary} reported that ChatGPT exhibits better performance in dialogue state tracking compared to spoken language understanding. Further, the authors showed that the performance of ChatGPT can be enhanced by (i) using a multi-turn interactive prompt for dialogue state tracking and (ii) providing additional details like slot names, examples and descriptions for slot filling in spoken language understanding. Zhao et al. \cite{zhao2023chatgpt} explored the emotion dialogue capabilities of ChatGPT by evaluating the model on five different tasks, namely emotion recognition, emotion cause recognition, dialogue act classification (emotion dialogue understanding), empathetic response generation and emotion support generation. It is reported that ChatGPT exhibits better performances in emotion dialogue generation compared to emotion dialogue understanding. Chintagunta et al. \cite{chintagunta2021medically} showed that the in-house model trained on GPT-3 generated summaries achieves performances comparable to when trained on human-generated summaries. Further, the in-house model trained on mixed summaries (human-generated and GPT-3 generated) achieves better performances than those trained on either one of the summaries.

\begin{table*}[h!]
\begin{center}
{\renewcommand{\arraystretch}{1.5}% for the vertical padding
\begin{tabular}{|p{0.6cm}|p{2.2cm}|p{2cm}|p{1cm}|p{4cm}|p{3.1cm}|p{1.1cm}|}
\hline

\scriptsize{\textbf{Paper}} & \scriptsize{\textbf{Task(s)}} & \scriptsize{\textbf{GLLMs Explored}} & \scriptsize{\textbf{Prompt Settings}} & \scriptsize{\textbf{Domain(s)}} & \scriptsize{\textbf{Language(s)}} & \scriptsize{\textbf{SOTA Results}} \\ \hline

\scriptsize{\cite{sun2023chatgpt}} & \scriptsize{Passage Re-ranking} & \scriptsize{ GPT-3, GPT-3.5, ChatGPT, GPT-4} & \scriptsize{ZS, FS} & \scriptsize{General, News, Healthcare, Scientific Literature} & \scriptsize{English, Ten Low Resource Languages} & \scriptsize{Yes} \\ \hline 

\scriptsize{\cite{ziems2023large}} & \scriptsize{Document Retrieval} & \scriptsize{GPT-3.5} & \scriptsize{ZS, FS} & \scriptsize{General} & \scriptsize{English} & \scriptsize{Yes}	\\ \hline 

\end{tabular}}
\end{center}
\caption{ \label{gllms-ir}  Summary of research works exploring GLLMs for information retrieval tasks. Here ZS represents zero-shot, and FS represents few-shot.} 
\end{table*}

Prodan et al. \cite{prodan2022prompt} proposed a scoring system to choose the best examples for dialogue summarizing using few-shot GPT-3. The proposed scoring system enhances the quality of generated summaries with an 11\% reduction in failures. Huynh et al. \cite{huynh2023understanding} studied the impact of various aspects influencing the performance of LLMs as Dialog evaluators. The authors reported that the performance as a dialogue evaluator largely depends on the diversity and relevance of the datasets used for instruction tuning. Fan et al. \cite{fan2023uncovering} investigated the effectiveness of ChatGPT for dialogue discourse analysis by evaluating its performance on three tasks, namely topic segmentation, discourse parsing and discourse relation recognition. ChatGPT’s performance is promising in the case of topic segmentation, and CoT prompting enhances the performance.  Wang et al. \cite{ wang2023chain} proposed a novel approach based on explicit CoT prompting and demonstration selection to answer dialogue questions in few-shot settings.

\subsection{Information Retrieval}

Information retrieval (IR) involves accessing and retrieving relevant information from large volumes of data. Here, the main objective is to provide users with the most relevant information by matching their queries to the content of documents and ranking them based on relevance \cite{anand2022explainable}. The process includes indexing, query formulation, search and retrieval, ranking, and presentation. Information retrieval is utilized in a wide range of fields, such as web search engines, digital libraries, e-commerce, healthcare, and scientific research \cite{anand2022explainable}. It plays a vital role in facilitating efficient and effective access to information in the modern digital era. Table \ref{gllms-ir} presents a summary of research works exploring GLLMs for information retrieval.

Sun et al. \cite{sun2023chatgpt} explored the effectiveness of GPT-3 family models like GPT-3, GPT-3.5, ChatGPT and GPT-4 for passage re-ranking in information retrieval. The results are promising as GPT-4 outperforms SOTA models like monoT5-3B \cite{nogueira2020document} on multiple benchmarks. Moreover, the compact model trained on ChatGPT-generated data demonstrates superior performance compared to the monoT5-3B model when evaluated on the MS MARCO dataset in BEIR \cite{thakur2021beir} benchmark. The existing approaches for document retrieval employ dual dense encoders, which encode query and document independently, resulting in shallow interaction between query and document \cite{zhao2022dense}. To overcome this drawback,  Ziems et al. \cite{ziems2023large} proposed a novel approach which involves generating URLs using LLMs for document retrieval. The authors reported that document retrieval by generating URLs outperforms existing approaches.

\subsection{Recommendation Systems}

\textbf{Overview.} Recommendation systems aim to reduce information overload and enhance the user experience by making relevant recommendations related to products or content based on user preferences and behaviour \cite{adomavicius2005toward}. In recent times, recommendation systems have gained immense popularity and are extensively utilized across a range of fields, such as entertainment, e-commerce,  social media etc. For example, popular platforms like YouTube and Netflix use recommendation systems to suggest relevant videos and platforms like Amazon use recommendation systems to suggest relevant products to the user \cite{peng2022survey}. The commonly used approaches for recommendation systems are based on collaborative filtering \cite{rezaimehr2021survey}, content-based \cite{xie2023rethinking} and knowledge-based \cite{dong2020interactive}. The performance of traditional recommendation systems is limited by a number of issues like cold-start problem, poor generalization across domains and lack of explainability \cite{gao2023chat, zhu2021cross}. 

To overcome these drawbacks in traditional recommendation systems, recent works explored GPT-3 family large language models for various tasks in recommendation systems like next item prediction \cite{wang2023zero}, rating prediction \cite{gao2023chat, zhiyuli2023bookgpt}, top-k predictions \cite{gao2023chat}, direct recommendation \cite{liu2023chatgpt}, sequence recommendation \cite{liu2023chatgpt} and generating explanations \cite{liu2023chatgpt}. The evaluation is done in a variety of domains like movies \cite{wang2023zero, dai2023uncovering, gao2023chat, kang2023llms, zhang2023chatgpt, hou2023large}, news \cite{dai2023uncovering}, books \cite{dai2023uncovering, kang2023llms, zhiyuli2023bookgpt}, music \cite{dai2023uncovering, zhang2023chatgpt}, social media \cite{mysore2023large}, beauty \cite{liu2023chatgpt}, and games \cite{hou2023large}. Table \ref{gllms-rs} presents a summary of research works exploring GLLMs for recommendation systems. 

\begin{table*}[h!]
\begin{center}
{\renewcommand{\arraystretch}{1.5}% for the vertical padding
\begin{tabular}{|p{0.6cm}|p{2.2cm}|p{2cm}|p{3.5cm}|p{1.3cm}|p{1.1cm}|}
\hline

\scriptsize{\textbf{Paper}}  & \scriptsize{\textbf{GLLMs Explored}} & \scriptsize{\textbf{Prompt Settings}} & \scriptsize{\textbf{Domain(s)}} & \scriptsize{\textbf{Language(s)}} & \scriptsize{\textbf{SOTA Results}} \\ \hline

\scriptsize{\cite{wang2023zero}} & \scriptsize{GPT-3.5} & \scriptsize{ZS} & \scriptsize{Movies} & \scriptsize{English} & \scriptsize{ No} \\ \hline 

\scriptsize{\cite{dai2023uncovering}} & \scriptsize{GPT-3.5, ChatGPT} & \scriptsize{ZS, FS} & \scriptsize{News, Books, Movies, Music} & \scriptsize{English} & \scriptsize{No}	 \\ \hline 

\scriptsize{\cite{gao2023chat}} & \scriptsize{GPT-3.5, ChatGPT} & \scriptsize{ZS} & \scriptsize{Movies} & \scriptsize{English} & \scriptsize{No}	 \\ \hline 

\scriptsize{\cite{mysore2023large}} & \scriptsize{InstructGPT} & \scriptsize{FS} & \scriptsize{Social Media} & \scriptsize{ English} & \scriptsize{No}	 \\ \hline 

\scriptsize{\cite{kang2023llms}} & \scriptsize{GPT-3.5, ChatGPT} & \scriptsize{	ZS, FS} & \scriptsize{Movies, Books} & \scriptsize{English} & \scriptsize{No}	 \\ \hline 

\scriptsize{\cite{zhang2023chatgpt}} & \scriptsize{ChatGPT} & \scriptsize{ZS} & \scriptsize{Music, Movies} & \scriptsize{ English} & \scriptsize{No} 	 \\ \hline 

\scriptsize{\cite{liu2023chatgpt}} & \scriptsize{ChatGPT} & \scriptsize{ZS, FS} & \scriptsize{Beauty} & \scriptsize{English} & \scriptsize{Yes}	 \\ \hline 

\scriptsize{\cite{ hou2023large}} & \scriptsize{ChatGPT} & \scriptsize{	ZS} & \scriptsize{Movies, Games} & \scriptsize{ English } & \scriptsize{No}	 \\ \hline 

\scriptsize{\cite{zhiyuli2023bookgpt}} & \scriptsize{ChatGPT} & \scriptsize{ZS, FS} & \scriptsize{Books} & \scriptsize{ English} & \scriptsize{No}	 \\ \hline 

\end{tabular}}
\end{center}
\caption{ \label{gllms-rs}  Summary of research works exploring GLLMs for recommendation systems. Here ZS represents zero-shot, and FS represents few-shot.} 
\end{table*}

\textbf{Research works exploring GLLMs for recommendation systems.} Wang et al. \cite{wang2023zero} proposed a novel prompting strategy called “Next-Item Recommendation (NIR)” to recommend movies using GLLMs. The proposed prompting strategy involves a three-step process to capture the user's preferences, choose representative movies they have watched in the past, and provide a ranked list of ten recommended movies. Dai et al. \cite{dai2023uncovering} reported that ChatGPT outperforms other GLLMs and is more effective with pair-wise and list-wise ranking compared to point-wise ranking. When it comes to balancing cost and performance, ChatGPT with list-wise ranking outperforms both point-wise and pair-wise ranking approaches. ChatGPT demonstrates the potential for providing explanations for recommendations and addressing the challenges of the cold start problem. Gao et al. \cite{gao2023chat} proposed Chat-REC, which leverages GLLMs to build conversational recommendation systems. The authors reported that Chat-REC performs well in tasks like top-k recommendations and zero-shot rating prediction. Moreover, Chat-REC enhances the conversational recommendation systems by making them more interactive and providing clear explanations.

Mysore et al. \cite{mysore2023large} explored GLLMs like InstructGPT to generate synthetic data, and the experiment results showed that narrative-driven recommendation models trained on augmented datasets outperform LLM baselines and other approaches. Kang et al. \cite{ kang2023llms}  evaluated GLLMs like GPT-3.5 and ChatGPT on user rating prediction in zero and few-shot settings. Based on the experimental findings on datasets from movies and book domains, the authors reported that traditional models that have access to user interaction data perform better than GLLMs. Zhang et al. \cite{zhang2023chatgpt} introduced FaiRLLM, a new benchmark having eight sensitive attributes from domains like movies and music, to investigate the fairness of GLLM recommendations. The authors reported that GLLM-based recommendation systems are not fair to certain sensitive attributes. 

Liu et al. \cite{liu2023chatgpt} evaluated the performance of ChatGPT in five recommendation tasks, which include predicting ratings, direct recommendation, sequence recommendation, generating explanations, and summarizing reviews. Based on the evaluation of Amazon beauty datasets, the authors reported that (i) ChatGPT is much better in rating prediction compared to other tasks like direct and sequence recommendation. and (ii) ChatGPT achieves new SOTA results in generating explanations based on human evaluation. Hou et al. \cite{hou2023large} demonstrated that GLLMs possess strong potential for zero-shot ranking tasks, showcasing performance that is comparable to or even superior to traditional recommendation models. Here, the authors designed the prompts in a way that important information like candidate items, sequential interaction history and ranking instruction is included.  Zhiyuli \cite{zhiyuli2023bookgpt} proposed BookGPT, a novel framework which leverages GLLMs like ChatGPT for book recommendation. Specifically, the performance of BookGPT is evaluated on three sub-tasks, namely the book rating task, book summary recommendation task and user rating recommendation task. The performance of BookGPT is promising in all three sub-tasks, and the performance increases with an increase in prompt examples. 

\begin{table*}[h!]
\begin{center}
{\renewcommand{\arraystretch}{1.5}% for the vertical padding
\begin{tabular}{|p{0.6cm}|p{2cm}|p{5.5cm}|p{1.2cm}|p{2.5cm}|p{2cm}|}
\hline

\scriptsize{\textbf{Paper}} & \scriptsize{\textbf{GLLMs Explored}} & \scriptsize{\textbf{Task(s)}} & \scriptsize{\textbf{Prompt Settings}} & \scriptsize{\textbf{Language(s)}} & \scriptsize{\textbf{SOTA Results}} \\ \hline

 \scriptsize{\cite{xia2023keep}} & \scriptsize{ChatGPT} & \scriptsize{Code Repair} & \scriptsize{ZS, FS} & \scriptsize{Java} & \scriptsize{Yes} \\ \hline 

\scriptsize{\cite{cheshkov2023evaluation}} & \scriptsize{GPT-3, ChatGPT} & \scriptsize{Code Vulnerability Detection} & \scriptsize{ZS} & \scriptsize{Java} & \scriptsize{No} \\ \hline 

\scriptsize{\cite{yeticstiren2023evaluating}} & \scriptsize{ChatGPT} & \scriptsize{Code Generation} & \scriptsize{ZS} & \scriptsize{Python} & \scriptsize{No} \\ \hline 

\scriptsize{\cite{li2023finding}} & \scriptsize{ChatGPT} & \scriptsize{Finding Failure-Inducing Test Cases} & \scriptsize{ZS} & \scriptsize{Python} & \scriptsize{Yes} \\ \hline 

\scriptsize{\cite{liu2023improving}} & \scriptsize{ ChatGPT} & \scriptsize{Code Generation} & \scriptsize{ZS} & \scriptsize{Java, C\#} & \scriptsize{No} \\ \hline 

\scriptsize{\cite{poldrack2023ai}} & \scriptsize{GPT-4} & \scriptsize{Code Generation, Code Refactoring, Test Case Generation} & \scriptsize{ZS} & \scriptsize{Python} & \scriptsize{No} \\ \hline 

\scriptsize{\cite{liu2023your}} & \scriptsize{ChatGPT, GPT-4} & \scriptsize{Code Generation} & \scriptsize{ZS} & \scriptsize{Python} & \scriptsize{No} \\ \hline 

\scriptsize{\cite{chen2023gptutor}} & \scriptsize{ ChatGPT} & \scriptsize{Code Explanation Generation} & \scriptsize{ZS} & \scriptsize{Python} & \scriptsize{No} \\ \hline 

\scriptsize{\cite{nascimento2023comparing}} & \scriptsize{ ChatGPT} & \scriptsize{Code Generation} & \scriptsize{ZS} & \scriptsize{C++} & \scriptsize{No} \\ \hline 

\scriptsize{\cite{khan2022automatic}} & \scriptsize{Codex} & \scriptsize{Code Documentation Generation} & \scriptsize{ZS, FS} & \scriptsize{Java, Python, PHP, GO, Ruby, JS} & \scriptsize{Yes} \\ \hline 

\scriptsize{\cite{leinonen2023comparing}} & \scriptsize{ 	GPT-3} & \scriptsize{Code Explanation Generation} & \scriptsize{ZS} & \scriptsize{C} & \scriptsize{No} \\ \hline 

\scriptsize{\cite{li2023think}} & \scriptsize{ChatGPT} & \scriptsize{Code Generation} & \scriptsize{ZS} & \scriptsize{Python} & \scriptsize{Yes} \\ \hline 

\scriptsize{\cite{prenner2021automatic}} & \scriptsize{ Codex} & \scriptsize{Automatic Code Repair} & \scriptsize{ZS, FS} & \scriptsize{Python, Java} & \scriptsize{No} \\ \hline 

\scriptsize{\cite{Siddiq2023ExploringTE}} & \scriptsize{Codex, ChatGPT} & \scriptsize{Unit Test Generation} & \scriptsize{ZS} & \scriptsize{Java} & \scriptsize{No} \\ \hline 

\scriptsize{\cite{tian2023chatgpt}} & \scriptsize{ChatGPT} & \scriptsize{	Code Generation, APR, Code Explanation Generation} & \scriptsize{ZS} & \scriptsize{Python} & \scriptsize{No} \\ \hline 

\scriptsize{\cite{Geng2023AnES}} & \scriptsize{Codex} & \scriptsize{Code Documentation Generation} & \scriptsize{ZS, FS} & \scriptsize{Java} & \scriptsize{Yes} \\ \hline 

\scriptsize{\cite{kang2023explainable}} & \scriptsize{Codex, ChatGPT} & \scriptsize{Automate Program Repair} & \scriptsize{ZS} & \scriptsize{Python, Java} & \scriptsize{No} \\ \hline 

\scriptsize{\cite{Kashefi2023ChatGPTFP}} & \scriptsize{ChatGPT} & \scriptsize{Code Generation} & \scriptsize{ZS} & \scriptsize{C, C++, Python, Julia, MATLAB} & \scriptsize{No} \\ \hline 

\scriptsize{\cite{destefanis2023preliminary}} & \scriptsize{GPT-3.5} & \scriptsize{Code Generation} & \scriptsize{ZS} & \scriptsize{Java} & \scriptsize{No} \\ \hline 

\scriptsize{\cite{Yuan2023NoMM}} & \scriptsize{ChatGPT} & \scriptsize{Unit Test Generation} & \scriptsize{ZS} & \scriptsize{Java} & \scriptsize{No} \\ \hline 

\scriptsize{\cite{Phung2023GenerativeAF}} & \scriptsize{ChatGPT, GPT-4} & \scriptsize{Code Repair, Code Completion, Code Explanation Generation, Coding Hints Generation} & \scriptsize{ZS} & \scriptsize{Python} & \scriptsize{No} \\ \hline 

\end{tabular}}
\end{center}
\caption{ \label{gllms-coding}  Summary of research works exploring GLLMs for various coding tasks. Here ZS represents zero-shot, and FS represents few-shot.} 
\end{table*}

\subsection{Coding Tasks}
\textbf{Overview.} Software engineering is a discipline which deals with designing, developing, testing, and maintaining software systems \cite{hou2023largesurvey}. To create software systems, software engineers use a variety of programming languages, development tools, and technologies. To aid software engineers and enhance their productivity, the research community focused on automating a number of coding tasks like code generation from natural language descriptions, code repair, code explanation generation, code hints generation, code completion, code document generation, test cases generation, code vulnerability detection, code refactoring, etc. The evolution of pre-trained source code models has paved the way for achieving cutting-edge results across coding tasks [455]. Some of the popular pretrained source code models are CodeBERT \cite{feng2020codebert}, CodeGPT \cite{lu2021codexglue}, CoTexT \cite{phan2021cotext}, GraphCodeBERT \cite{guo2020graphcodebert}, CodeT5 \cite{wang2021codet5}, CodeT5+ \cite{wang2023codet5+}, PLBART \cite{ahmad2021unified}, PyCodeGPT \cite{zan2022cert} etc.  Inspired by the success of GLLMs in NLP tasks, the research community focused on assessing the performances of these models in coding tasks also.

\textbf{Research works exploring GLLMs for various coding tasks.} The research community explored GLLMs for coding tasks across various languages like Java \cite{xia2023keep, cheshkov2023evaluation, liu2023improving, khan2022automatic, prenner2021automatic, Siddiq2023ExploringTE, Geng2023AnES, kang2023explainable, destefanis2023preliminary, Yuan2023NoMM}, Python \cite{yeticstiren2023evaluating, li2023finding, poldrack2023ai, liu2023your, chen2023gptutor, khan2022automatic, li2023think,  prenner2021automatic, tian2023chatgpt, kang2023explainable, Kashefi2023ChatGPTFP, Phung2023GenerativeAF}, PHP \cite{khan2022automatic}, GO  \cite{khan2022automatic}, Ruby \cite{khan2022automatic}, JavaScript \cite{khan2022automatic}, C \cite{leinonen2023comparing, Kashefi2023ChatGPTFP}, C++ \cite{nascimento2023comparing, Kashefi2023ChatGPTFP}, Julia \cite{Kashefi2023ChatGPTFP}, and MATLAB \cite{Kashefi2023ChatGPTFP}. Most of the research works focused on Python and Java languages, while a few research works focused on other languages like GO, PHP, GO, Ruby, JavaScript, C, C++, Julia and MATLAB.  The assessment is done in zero and few-shot settings using mostly direct prompts. Table \ref{gllms-coding} presents a summary of research works exploring GLLMs for various coding tasks.

Some of the research works \cite{yeticstiren2023evaluating, liu2023your, nascimento2023comparing, Kashefi2023ChatGPTFP, destefanis2023preliminary} explored GLLMs for code generation task. Yeticstiren et al. \cite{yeticstiren2023evaluating}  compared various AI-assisted code generation tools like ChatGPT, Amazon’s Code Whisperer and Github’s Copilot on the Human Eval \cite{chen2021evaluating} dataset. ChatGPT outperforms other tools by generating correct code 65.2\% of the time, while the other tools generate correct code for a maximum of 46.3\% of the time only.  The test cases in existing datasets for code generation evaluation are limited in terms of quality and quantity. So, Liu et al. \cite{liu2023your}  proposed EvaPlus, a new framework for automatic test case generation using ChatGPT and the traditional mutation approach. The authors use EvaPlus to develop HumanEvalPlus on the top of the HumanEval \cite{chen2021evaluating} dataset. The authors reported that HumanEvalPlus can detect a lot of incorrectly generated code that was previously undetected. Nascimento et al. \cite{nascimento2023comparing}  compared the quality of code generated by ChatGPT and software developers for competitive coding problems on the LeetCode platform using various evaluation metrics. The authors reported that ChatGPT exhibits better performance compared to novice programmers but is outperformed by experienced programmers. Kashefi et al. \cite{Kashefi2023ChatGPTFP} explored how effective ChatGPT is for generating code for numerical methods in five different programming languages: C, C++, Python, MATLAB and Julia. The authors observed that the results are promising but have some limitations which require further investigation. Destefanis et al. \cite{destefanis2023preliminary} assessed the code generation ability of LLMs like Bard and GPT-3.5 by evaluating their performances in generating Java language code given the natural language descriptions. The authors observed that GPT-3.5 outperforms the Bard model by a large margin of more than 37\%.

Some of the research works \cite{prenner2021automatic, tian2023chatgpt, kang2023explainable, Phung2023GenerativeAF} explored GLLMs for code repair task. Prenner et al. \cite{prenner2021automatic} explored the Codex model for automatic program repair in Python and Java programming languages. The authors observed that the performance of Codex is comparable to state-of-the-art methods. Moreover, the Codex model is slightly better at fixing errors in Python language compared to Java language. Kang et al. \cite{kang2023explainable} developed AutoSD, a novel framework for automatic program repair using GLLMs. The authors reported that the evaluation on three standard datasets showed that the proposed framework is on par with the baselines. 

Unit tests generated using traditional approaches suffer from low readability \cite{Yuan2023NoMM}. To address this drawback, some of the research works \cite{Siddiq2023ExploringTE, Yuan2023NoMM} explored GLLMs for test case generation. Siddiq et al. \cite{Siddiq2023ExploringTE} evaluated models like Codex and ChatGPT for unit test generation for Java code. Experiment results showed that Codex performs better with 80\% coverage for the HumanEval dataset. However, both models perform poorly in the case of the SF110 benchmark, with less than 2\% coverage. Yuan et al. \cite{Yuan2023NoMM} designed a ChatGPT-based unit test generation framework called “Chat-Tester”. The iterative test refiner helps Chat-Tester to generate better unit tests compared to vanilla ChatGPT.

In all the above discussed research works, the performance of GLLMs in various coding tasks is promising but still lags behind SOTA results. Some of the research works \cite{xia2023keep, li2023finding, khan2022automatic, li2023think, Geng2023AnES}  demonstrated that GLLMs can achieve SOTA results in coding tasks. Xia et al. \cite{xia2023keep} proposed ChatRepair, an automatic program repair tool based on ChatGPT. ChatRepair achieves remarkable performance, surpassing all the existing methods. It successfully resolves 114 and 48 bugs on Defects4j 1.2 and 2.0 \cite{just2014defects4j}, respectively, outperforming the previous best by 15 and 17 bugs, respectively. Khan et al. \cite{khan2022automatic} explored Codex, GPT-3 family model pretrained on natural and programming languages to automate code documentation generation. The evaluation results on six programming languages showed that Codex, with just one example, outperforms existing approaches by a large margin of 11.2\%. Geng et al. \cite{Geng2023AnES} explored Codex for code document generation and demonstrated that few-shot in-context learning with systematic demonstration selection helps the GPT-3 model to achieve new SOTA results on two standard datasets related to Java language.

Some of the research works \cite{li2023finding, liu2023improving, li2023think} explored advanced prompting like CoT, brainstorming, differential prompting, etc., for coding tasks. Liu et al. \cite{liu2023improving} evaluated the code generation capabilities of ChatGPT by evaluating its performances on text-to-code and code-to-code generation tasks on CodeXGLUE  \cite{lu2021codexglue} datasets. The authors observed that advanced prompting strategies like CoT enhance the code generation capabilities of models like ChatGPT. Li et al. \cite{li2023think} proposed Brainstorm, a new framework for code generation. Brainstorm involves three steps: brainstorming to generate diverse thoughts, thoughts selection to select the best thought using a ranking model and writing code to generate the code based on the problem statement and the best thought. The authors reported that the proposed framework helps ChatGPT to increase its performance by more than 50\% and achieve new SOTA results on the CodeContests \cite{li2022competition} benchmark. Li et al. \cite{ li2023finding} showed that directly using ChatGPT to find failure-inducing test cases results in poor performances. So, the authors proposed a new prompting strategy called “Differential Prompting”, which enables ChatGPT to achieve new SOTA results on the Quixbugs dataset \cite{lin2017quixbugs}. Differential Prompting involves program intention inference followed by two more steps: program generation and differential testing.

\subsection{Multimodal AI Tasks}
\textbf{Overview.} Traditional AI systems are designed to handle data from a single modality such as text, image,  audio or video. As real-world data is often multi-modal, researchers focused on developing multi-modal AI systems which can leverage input data from multiple modalities to generate more accurate results. Multi-modal AI systems leverage techniques from different areas of AI, like natural language processing, computer vision, speech processing etc., to process multi-modal input data effectively \cite{sundar2022multimodal, xu2023multimodal}. Multi-Modal AI systems can perform a variety of understanding and generation tasks like visual question answering \cite{shao2023prompting, lin2022revive, yang2022empirical, gui2022kat}, text-to-image generation \cite{ lu2023llmscore, zhu2023collaborative, zhang2023controllable}, text-to-video generation \cite{hong2023large}, text-to-speech synthesis \cite{huang2023audiogpt}, speech-to-text synthesis \cite{huang2023audiogpt}, image captioning \cite{ranjit2023retrieval} etc. 

\textbf{Research works exploring GLLMs for Multimodal AI tasks.} After the huge success of LLMs in natural language generation and understanding tasks, the research community recently explored GPT-3 family models in multi-modal understanding and generation tasks in various combinations like image+language \cite{kalakonda2022action, wu2023visual, shao2023prompting, yang2023mm, ranjit2023retrieval, lin2022revive, lu2023llmscore, zhu2023collaborative, li2023prompt, hakimov2023images, yang2022empirical, feng2023layoutgpt, zhang2023controllable, fan2023improving, li2023llava, gui2022kat}, video+language \cite{bhattacharya2023video, hong2023large}, audio+language \cite{mei2023wavcaps, zhang2023speechgpt}. Most of the research works focused on general domain datasets, which some of the research works focused on specific domains like healthcare \cite{ranjit2023retrieval, li2023llava}. Table \ref{gllms-mm} presents a brief summary of research works exploring GLLMs for various multimodal AI tasks.

\begin{table*}[h!]
\begin{center}
{\renewcommand{\arraystretch}{1.5}% for the vertical padding
\begin{tabular}{|p{0.6cm}|p{2cm}|p{5.5cm}|p{1.2cm}|p{2.5cm}|p{2cm}|}
\hline

\scriptsize{\textbf{Paper}} & \scriptsize{\textbf{GLLMs Explored}} & \scriptsize{\textbf{Task(s)}} & \scriptsize{\textbf{Prompt Settings}} & \scriptsize{\textbf{Multimodality}} & \scriptsize{\textbf{Domain}} \\ \hline

\scriptsize{\cite{ kalakonda2022action}} & \scriptsize{GPT-3} & \scriptsize{Text-based Action Generation} & \scriptsize{ZS} & \scriptsize{Image + Language} & \scriptsize{General}  \\ \hline 

 \scriptsize{\cite{wu2023visual}} & \scriptsize{ChatGPT} & \scriptsize{Twenty Two Vision Language Tasks} & \scriptsize{ZS} & \scriptsize{Image + Language} & \scriptsize{General}  \\ \hline 

 \scriptsize{\cite{shao2023prompting}} & \scriptsize{GPT-3} & \scriptsize{Knowledge-based Visual Question Answering} & \scriptsize{FS} & \scriptsize{Image + Language} & \scriptsize{General} \\ \hline 

\scriptsize{\cite{mei2023wavcaps}} & \scriptsize{ChatGPT} & \scriptsize{	Audio Labelling} & \scriptsize{ZS} & \scriptsize{Audio + Language} & \scriptsize{General} \\ \hline 

\scriptsize{\cite{yang2023mm}} & \scriptsize{ChatGPT} & \scriptsize{Multi-Image Reasoning, Multi-hop Document Understanding, Open-World Concept Understanding, Video Summarization} & \scriptsize{ZS} & \scriptsize{Image + Language} & \scriptsize{General} \\ \hline 

 \scriptsize{\cite{ranjit2023retrieval}} & \scriptsize{GPT-3.5, ChatGPT, GPT-4} & \scriptsize{Chest X-Ray Report Generation} & \scriptsize{ZS} & \scriptsize{Image + Language} & \scriptsize{Healthcare} \\ \hline 

\scriptsize{\cite{lin2022revive}} & \scriptsize{GPT-3} & \scriptsize{Knowledge-based Visual Question Answering} & \scriptsize{FS} & \scriptsize{Image + Language} & \scriptsize{General} \\ \hline 

\scriptsize{\cite{bhattacharya2023video}} & \scriptsize{GPT-3.5} & \scriptsize{Five Video Understanding Tasks} & \scriptsize{ZS} & \scriptsize{Video + Language} & \scriptsize{General} \\ \hline 

\scriptsize{\cite{zhang2023speechgpt}} & \scriptsize{GPT-4} & \scriptsize{Generate Instructions} & \scriptsize{ZS} & \scriptsize{Audio + Language} & \scriptsize{General} \\ \hline 

\scriptsize{\cite{lu2023llmscore}} & \scriptsize{GPT-3.5, GPT-4} & \scriptsize{Evaluator for Text-to-Image Generation} & \scriptsize{ZS} & \scriptsize{Image + Language} & \scriptsize{General} \\ \hline 

\scriptsize{\cite{zhu2023collaborative}} & \scriptsize{GPT-3, GPT-3.5} & \scriptsize{Editing in Text-to-Image Generation} & \scriptsize{FS} & \scriptsize{Image + Language} & \scriptsize{General} \\ \hline 

\scriptsize{\cite{li2023prompt}} & \scriptsize{ChatGPT} & \scriptsize{Multimodal Named Entity Recognition} & \scriptsize{FS} & \scriptsize{Image + Language} & \scriptsize{General} \\ \hline 

\scriptsize{\cite{hakimov2023images}} & \scriptsize{GPT-3} & \scriptsize{Five vision language tasks (four classification tasks and one question answering task) } & \scriptsize{	FS} & \scriptsize{Image + Language} & \scriptsize{General} \\ \hline 

\scriptsize{\cite{hong2023large}} & \scriptsize{GPT-4} & \scriptsize{Text-to-Video Generation} & \scriptsize{ZS} & \scriptsize{Video + Language} & \scriptsize{General} \\ \hline 

\scriptsize{\cite{yang2022empirical}} & \scriptsize{GPT-3} & \scriptsize{Knowledge-based Visual Question Answering} & \scriptsize{FS} & \scriptsize{Image + Language} & \scriptsize{General} \\ \hline 

\scriptsize{\cite{feng2023layoutgpt}} & \scriptsize{GPT-3.5, ChatGPT, GPT-4} & \scriptsize{Layout Generation} & \scriptsize{FS} & \scriptsize{Image + Language} & \scriptsize{General} \\ \hline 

\scriptsize{\cite{zhao2023chatbridge}} & \scriptsize{ChatGPT, GPT-4} & \scriptsize{Multimodal tasks covering text, video, audio and images} & \scriptsize{ZS} & \scriptsize{Multimodal covering text, video, audio and images} & \scriptsize{General} \\ \hline 

\scriptsize{\cite{zhang2023controllable}} & \scriptsize{GPT-3.5, ChatGPT, GPT-4} & \scriptsize{Controlled Text-to-Image Generation} & \scriptsize{ZS} & \scriptsize{Image + Language} & \scriptsize{General} \\ \hline

\scriptsize{\cite{fan2023improving}} & \scriptsize{ChatGPT} & \scriptsize{Paraphrasing} & \scriptsize{ZS} & \scriptsize{Image + Language} & \scriptsize{General} \\ \hline 

\scriptsize{\cite{huang2023audiogpt}} & \scriptsize{ChatGPT} & \scriptsize{Audio Understanding and Generation Tasks} & \scriptsize{ZS} & \scriptsize{Multimodal covering text,  audio and images} & \scriptsize{General} \\ \hline 

\scriptsize{\cite{li2023llava}} & \scriptsize{GPT-4} & \scriptsize{Generate Instruction Tuning Dataset} & \scriptsize{FS} & \scriptsize{Image + Language} & \scriptsize{Healthcare} \\ \hline

\scriptsize{\cite{gui2022kat}} & \scriptsize{GPT-3} & \scriptsize{Knowledge-based Visual Question Answering} & \scriptsize{FS} & \scriptsize{Image + Language} & \scriptsize{General} \\ \hline 

\end{tabular}}
\end{center}
\caption{ \label{gllms-mm}  Summary of research works exploring GLLMs for various multimodal AI tasks. Here ZS represents zero-shot, and FS represents few-shot.} 
\end{table*}

Some of the research works developed multi-model AI systems for a specific task like action generation \cite{kalakonda2022action}, knowledge-based visual question answering \cite{shao2023prompting, lin2022revive, yang2022empirical, gui2022kat}, x-ray report generation \cite{ranjit2023retrieval}, named entity recognition \cite{li2023prompt}, text-to-video generation \cite{hong2023large}, layout generation \cite{feng2023layoutgpt}, text-to-image generation \cite{zhang2023controllable}. Kalakonda et al. \cite{kalakonda2022action} proposed GPT-3 based plug-and-play framework called Action-GPT for text-based action generation. Here, the authors generated multiple detailed body movement descriptions from the action phrases and then used them to generate actions. Shao et al. \cite{shao2023prompting} proposed Prophet, which avoids using an external knowledge base by using GPT-3 as an implicit knowledge base and includes vanilla visual question answering to provide answer heuristics to GPT-3. The answer heuristics, along with caption and question information, provide rich task-specific information to the GPT-3 model, which results in much better performances. Ranjit et al. \cite{ranjit2023retrieval} proposed automatic x-ray report generation based on contrastively pretrained vision-language encoder and GPT-3 family models like GPT-3.5, ChatGPT and GPT-4. The contrastively pretrained encoder is used to encode input x-ray image into image vector embedding based on which the most similar sentences from the radiology report corpus are retrieved. The retrieved similar sentences form the context and allow LLM to generate a quality X-Ray report. Li et al. \cite{li2023prompt} proposed PGIM, a two-stage approach which utilizes ChatGPT as an implicit knowledge base for multi-modal NER task. In the first stage, ChatGPT, when prompted with text descriptions of the image, generates the auxiliary knowledge. In the second stage, the downstream model receives the raw text and ChatGPT-generated auxiliary knowledge as input. The authors reported that the proposed approach outperforms existing SOTA approaches based on text-text and text-image paradigms.

Hong et al. \cite{hong2023large} proposed DirecT2V for text-to-video generation, which leverages GPT-4 model as a frame-level director. Here, the GPT-4 model generates descriptions for each frame based on a single prompt, and then the Text-to-Image model is used to generate frames based on these descriptions.  Feng et al. \cite{feng2023layoutgpt} developed LayoutGPT, which leverages LLM and Layout-to-Image models to generate 2D and 3D planning layouts from text descriptions. Zhang et al. \cite{zhang2023controllable} proposed “Control-GPT” based on LLMs and diffusion models for controllable text-to-image generation. Here, GPT-4 generates sketches based on Tikz code based on the text instructions, and then diffusion model generates realistic images with generated sketches and the text instructions as input. Here, the generated sketches help diffusion models to get a better idea about spatial relationships.

Some of the research works focused on developing multi-model AI systems which can handle multiple tasks \cite{wu2023visual, yang2023mm, bhattacharya2023video, hakimov2023images, zhao2023chatbridge, huang2023audiogpt}. As ChatGPT is trained on one data modality i.e., text data, ChatGPT can only handle text inputs and training models from scratch for vision-language tasks, is not a feasible option as it involves huge computation. So, Wu et al. \cite{wu2023visual} developed Visual ChatGPT based on ChatGPT and various visual foundation models to handle 22 vision language tasks.  Bhattacharya et al. \cite{bhattacharya2023video} proposed a novel three-stage approach to handle five video understanding tasks. The proposed approach involves transforming video into text stories and then using this text content for video understanding tasks. Hakimov et al. \cite{hakimov2023images} explored GPT-3 model for five vision language tasks, including four classifications and one question answering. Here the model is prompted with text description of the input image along with other elements like task instruction and similar examples. Huang et al. \cite{huang2023audiogpt} proposed AudioGPT, which allows ChatGPT to handle multiple audio understanding and generation tasks with the help of audio foundation models. 

Some of the research works explored GPT-3 family models for other tasks like data labelling \cite{mei2023wavcaps}, generating instructions \cite{zhang2023speechgpt}, data generation \cite{fan2023improving}, prompt editing \cite{zhu2023collaborative} and evaluation \cite{lu2023llmscore} while developing multimodal AI systems. Mei et al. \cite{mei2023wavcaps} used ChatGPT to rewrite those noisy audio captions and developed WavCaps, an audio captions dataset of 400k instances. The authors reported that the models trained on WavCaps datasets achieve new SOTA results. Zhang et al. \cite{zhang2023speechgpt} developed SpeechGPT and then do cross-modal instruction tuning to enhance its multi-model instruction following ability. Here, the authors use GPT-4 to generate the instructions for diverse tasks.  Fan et al. \cite{fan2023improving} proposed LaCLIP (Language augmented Contrastive Language-Image Pretraining), an extended version of CLIP which applies data augmentation to both text and image data to ensure that the model gets exposed to diversified texts during training. Here the data augmentation is performed using the open-source LLaMA model in few-shot settings, and the examples for LLaMA ICL are generated using ChatGPT. Zhu et al. \cite{zhu2023collaborative} explored GPT-3 and GPT-3.5 models for prompt editing in text-to-image generation. The authors observed a potential reduction of 20-30\% in the remaining edits required by implementing the prompt edits suggested by GPT-3 family models. Lu et al. \cite{lu2023llmscore} proposed LLMScore, a new metric which can effectively capture both image and object-level compositionality for text-to-image generation evaluation.

\begin{table*}[h!]
\begin{center}
{\renewcommand{\arraystretch}{1.5}% for the vertical padding
\begin{tabular}{|p{0.6cm}|p{5cm}|p{2.5cm}|p{1.2cm}|p{2.5cm}|p{2cm}|}
\hline

\scriptsize{\textbf{Paper}} & \scriptsize{\textbf{Task(s)}} & \scriptsize{\textbf{GLLMs Explored}} & \scriptsize{\textbf{Prompt Settings}} & \scriptsize{\textbf{Language(s)}} \\ \hline

\scriptsize{\cite{zheng2023can}}	& \scriptsize{Neural Architecture Search} & \scriptsize{GPT-4} & \scriptsize{ZS} & \scriptsize{English} \\ \hline 

\scriptsize{\cite{shen2023hugginggpt}} & \scriptsize{Multiple AI tasks in language, speech and vision areas} & \scriptsize{GPT-3.5, ChatGPT, GPT-4} & \scriptsize{FS} & \scriptsize{English} \\ \hline 

\scriptsize{\cite{ zhang2023mlcopilot}} & \scriptsize{Machine Learning Tasks} & \scriptsize{GPT-3.5} & \scriptsize{FS} & \scriptsize{English} \\ \hline 

\scriptsize{ \cite{zhang2023automl}}	& \scriptsize{Machine Learning  Tasks} & \scriptsize{GPT-4} & \scriptsize{FS} & \scriptsize{English} \\ \hline 

\end{tabular}}
\end{center}
\caption{ \label{gllms-ml}  Summary of research works exploring GLLMs to automate machine learning tasks. Here ZS represents zero-shot, and FS represents few-shot.} 
\end{table*}

\subsection{Machine Learning Tasks}
\textbf{Overview.} Machine learning (ML) is an area of artificial intelligence (AI) that deals with the development of algorithms that can learn from data and make decisions \cite{zhang2023mlcopilot}. Even though machine learning algorithms are successfully used in various real-world applications, creating an effective ML solution for a new task can be difficult due to the numerous design choices involved. In recent times, AutoML has evolved as a solution to reduce the human effort involved in designing ML solutions \cite{hutter2019automated}. However, AutoML algorithms suffer from various drawbacks \cite{zhang2023mlcopilot}, like (i) the requirement of multiple rounds of trial-and-error, resulting in significant time consumption, (ii) starting the search for a new task from scratch, ignoring past experience gained from the previous tasks and (iii) many AutoML methods lack interpretability because of their black-box nature.  

\begin{table*}[h!]
\begin{center}
{\renewcommand{\arraystretch}{1.5}% for the vertical padding
\begin{tabular}{|p{0.6cm}|p{5cm}|p{2.5cm}|p{1.2cm}|p{2.5cm}|p{2cm}|} 
\hline

\scriptsize{\textbf{Paper}} & \scriptsize{\textbf{Task(s)}} & \scriptsize{\textbf{GLLMs Explored}} & \scriptsize{\textbf{Prompt Settings}} & \scriptsize{\textbf{Language(s)}} & \scriptsize{\textbf{SOTA Results}} \\ \hline

\scriptsize{\cite{olmo2021gpt3}} & \scriptsize{Plan Extraction}	& \scriptsize{GPT-3} & \scriptsize{FS} & \scriptsize{English} & \scriptsize{	Yes}  \\ \hline

\scriptsize{\cite{zhang2023large}} & \scriptsize{Planning in Human-Robot Interaction} & \scriptsize{GPT-3.5} & \scriptsize{ZS} & \scriptsize{English} & \scriptsize{No} \\ \hline 

\scriptsize{\cite{xie2023translating}} & \scriptsize{Plan Extraction} & \scriptsize{GPT-3.5} & \scriptsize{	FS} & \scriptsize{English} & \scriptsize{	No}  \\ \hline 

\scriptsize{\cite{hu2023chain}} & \scriptsize{Planning} & \scriptsize{InstructGPT, ChatGPT} & \scriptsize{FS} & \scriptsize{English} & \scriptsize{	No} \\ \hline 
						
\end{tabular}}
\end{center}
\caption{ \label{gllms-pl}  Summary of research works exploring GLLMs for planning. Here ZS represents zero-shot, and FS represents few-shot.} 
\end{table*}

\textbf{Research works exploring GLLMs to automate machine learning tasks.} Inspired by the success of GLLMs in other tasks, the research community explored GLLMs as an alternative to AutoML to automate machine learning tasks \cite{zheng2023can, shen2023hugginggpt, zhang2023mlcopilot, zhang2023automl}. Table \ref{gllms-ml} presents a summary of research works exploring GLLMs to automate machine learning tasks. Zheng et al. \cite{zheng2023can} explored how effective is GPT-4 for neural architecture search, i.e., designing optimal neural network configurations. The proposed approach involves two steps, namely (i) GPT-4 generates the optimal neural architecture based on the given problem statement, (ii) the generated configuration is evaluated, and for further refinement, the evaluation results along with the problem statement are passed to the model. This two-step process is repeated for a certain number of iterations to achieve the optimal configuration. Shen et al. \cite{shen2023hugginggpt} proposed HuggingGPT to solve AI tasks with the help of GLLMs like ChatGPT and models in AI communities like Hugging Face. HuggingGPT involves four steps, namely task planning, model selection, task execution and response generation. The authors reported that HuggingGPT achieves promising results in solving AI tasks in language, vision and speech.

Zhang et al. \cite{zhang2023mlcopilot} proposed MLCopilot, which leverages the power of GLLMs to solve machine learning tasks. MLCopilot works in two stages, namely offline and online. The offline stage involves creating an experience pool from which GLLM is used to retrieve relevant knowledge. The online stage involves retrieving relevant examples from the experience pool, and then GLLM generates results based on the task description, relevant examples and knowledge. Zhang et al. \cite{zhang2023automl} proposed AutoML-GPT, which leverages the advanced GPT-4 GLLM to automatic machine learning tasks and reduces human efforts in building machine learning models. AutoML-GPT involves two stages. The first stage involves composing a prompt paragraph based on the model and data cards. The second stage involves performing the four crucial steps from data processing to training log prediction.

\subsection{ Planning} 
\textbf{Overview.} Many important industries, like finance and banking, often involve repetitive sequential tasks. These workflows, despite their significance, are typically not fully automated or formally defined. Recently, due to strong reasoning capabilities, the research community explored GLLMs for planning. Some of the research works \cite{zhang2023large, hu2023chain} directly used LLMs for planning, while some of them \cite{olmo2021gpt3, xie2023translating} explored LLMs for planning extraction, which can then be used by automated systems. 

\textbf{Research works exploring GLLMs for planning.} Table \ref{gllms-pl} presents a summary of research works exploring GLLMs for planning. Human models are crucial in facilitating human-robot interaction (HRI), as they empower robots to plan their behaviour based on the impact of their actions on individuals. As it is difficult to craft good human labels,  Zhang et al. \cite{zhang2023large} used the GPT-3.5 model (i)  as zero-shot human models and also (ii) for planning in trust-related scenarios. Hu et al. \cite{hu2023chain} proposed a novel prompting strategy called “Chain of Symbol” prompting to elicit better the planning abilities of large language models like InstructGPT and ChatGPT.  Unlike CoT prompting, which uses natural language descriptions to represent complex environments, CoS prompting uses condensed symbols to represent them in intermediate reasoning steps. The authors reported that CoS prompting outperforms CoT prompting in both performance and efficiency.

There are usually natural language documents that describe the procedures for the company's employees. Plan extraction methods offer the opportunity to extract structured plans from these natural language descriptions of workflows [93, 95]. These extracted plans can then be used by automated systems. Olmo et al. \cite{olmo2021gpt3} explored the GPT-3 model for plan extraction in few-shot settings from the natural language descriptions of workflows and showed that GPT-3 model outperforms existing SOTA models in some cases.  Xie et al. \cite{xie2023translating} explored GPT-3.5 models to extract plans from natural language descriptions. The authors reported that the models are poor planners on their own, which is in line with the existing works \cite{valmeekam2022large, collins2022structured, mahowald2023dissociating} and are better at extracting plans from natural language. However, these models are sensitive to prompts and also struggle in the case of tasks involving spatial or numerical reasoning.

\section{Performance of GLLMs in Specific Domains}
\label{section-6}
Apart from the general domain, natural language processing is also explored in specific domains like healthcare, finance, legal, social media, etc.  Analyzing domain-specific texts is more challenging because of domain-specific terminology and abbreviations, complex language structures, etc.  In domains like healthcare, finance and legal,  domain experts use many words and abbreviations that are specific to the domain and not commonly found in general domain texts. In domains like social media, the texts are mostly authored by the general public using informal language and slang words. Moreover, social media texts are noisy, with many misspelt words, emojis, irregular grammar and abbreviations \cite{kalyan2020medical, kalyan2020target}. 

\begin{table*}[h!]
\begin{center}
{\renewcommand{\arraystretch}{1.5}% for the vertical padding
\begin{tabular}{|p{0.6cm}|p{2cm}|p{6cm}|p{1.2cm}|p{2cm}|p{2cm}|} 
\hline

\scriptsize{\textbf{Paper}} & \scriptsize{\textbf{GLLMs Explored}} & \scriptsize{\textbf{Task(s)}} & \scriptsize{\textbf{Prompt Settings}} & \scriptsize{\textbf{Language(s)}} & \scriptsize{\textbf{Outperforms Domain-Specific Models}}\\ \hline

\scriptsize{\cite{holmes2023evaluating}} & \scriptsize{ChatGPT, GPT-4} & \scriptsize{Question Answering} & \scriptsize{ZS} & \scriptsize{English}  & \scriptsize{-} \\ \hline 

\scriptsize{\cite{liu2023deid}} & \scriptsize{ChatGPT, GPT-4} & \scriptsize{Text De-identification} & \scriptsize{ZS} & \scriptsize{English} & \scriptsize{Yes} \\ \hline 

\scriptsize{\cite{giorgi2023wanglab}} & \scriptsize{GPT-4} & \scriptsize{Dialogue Summarization}  & \scriptsize{FS} & \scriptsize{English} & \scriptsize{Yes} \\ \hline 

\scriptsize{\cite{Nori2023CapabilitiesOG}} & \scriptsize{GPT-3.5, ChatGPT, GPT-4}  & \scriptsize{Question Answering} & \scriptsize{ZS, FS} & \scriptsize{English} & \scriptsize{Yes} \\ \hline 

\scriptsize{\cite{chen2023large}} & \scriptsize{GPT-3.5, GPT-4} & \scriptsize{Named Entity Recognition, Relation Extraction, Document Classification and Semantic Similarity} & \scriptsize{ZS, FS} & \scriptsize{English} & \scriptsize{Yes} \\ \hline 

\scriptsize{\cite{Tanaka2023PerformanceOG}} & \scriptsize{GPT-3.5, ChatGPT} & \scriptsize{Question Answering} & \scriptsize{ZS} & \scriptsize{Japanese} & \scriptsize{-} \\ \hline 

\scriptsize{\cite{liu2023benchmarking}} & \scriptsize{GPT-3.5, GPT-4} & \scriptsize{Question Answering, Reasoning} & \scriptsize{ZS} & \scriptsize{Chinese} & \scriptsize{Yes} \\ \hline 

\scriptsize{\cite{yang2023data}} & \scriptsize{GPT-3} & \scriptsize{Text Simplification} & \scriptsize{FS} & \scriptsize{English} & \scriptsize{-} \\ \hline 

\scriptsize{\cite{gutierrez2022thinking}} & \scriptsize{GPT-3} & \scriptsize{Entity Extraction, Relation Classification} & \scriptsize{FS} & \scriptsize{English} & \scriptsize{No} \\ \hline 

\scriptsize{[\cite{wu2023exploring}} & \scriptsize{ChatGPT, GPT-4} & \scriptsize{Natural Language Inference} & \scriptsize{ZS, FS} & \scriptsize{English} & \scriptsize{-} \\ \hline 

\scriptsize{\cite{ma2023impressiongpt}} & \scriptsize{ ChatGPT} & \scriptsize{Text Summarization} & \scriptsize{FS} & \scriptsize{English} & \scriptsize{Yes} \\ \hline 

\scriptsize{\cite{wang2023large}} & \scriptsize{ GPT3.5, GPT4} & \scriptsize{Natural Language Inference, Document Classification} & \scriptsize{ZS, FS} & \scriptsize{English} & \scriptsize{-} \\ \hline 

\scriptsize{\cite{kasai2023evaluating}} & \scriptsize{GPT-3, ChatGPT, GPT-4} & \scriptsize{Question Answering} & \scriptsize{FS} & \scriptsize{Japanese} & \scriptsize{-} \\ \hline 

\scriptsize{\cite{moradi2021gpt}} & \scriptsize{GPT-3} & \scriptsize{Natural Language Inference, Relation Classification, Semantic Similarity, Question Answering, Text Classification} & \scriptsize{FS} & \scriptsize{English} & \scriptsize{No} \\ \hline 

\scriptsize{\cite{jeblick2022chatgpt}} & \scriptsize{ChatGPT} & \scriptsize{Text Simplification} & \scriptsize{ZS} & \scriptsize{English} & \scriptsize{-} \\ \hline 

\scriptsize{\cite{tang2023gersteinlab}} & \scriptsize{GPT-3, GPT-4} & \scriptsize{Dialogue Summarization} & \scriptsize{FS} & \scriptsize{English} & \scriptsize{-} \\ \hline 

\scriptsize{\cite{agrawal2022large}} & \scriptsize{GPT-3} & \scriptsize{Clinical Sense Disambiguation, Biomedical Evidence Extraction, Coreference Resolution, Medication Status Extraction, Medication Attribute Extraction} & \scriptsize{ZS, FS} & \scriptsize{English} & \scriptsize{-} \\ \hline 

\scriptsize{\cite{nair2023generating}} & \scriptsize{ GPT-3} & \scriptsize{Dialogue Summarization} & \scriptsize{ZS, FS} & \scriptsize{English} & \scriptsize{-} \\ \hline 

\scriptsize{\cite{shaib2023summarizing}} & \scriptsize{GPT-3} & \scriptsize{Text Summarization} & \scriptsize{ZS,FS} & \scriptsize{English} & \scriptsize{-} \\ \hline 

\scriptsize{\cite{xu2023medgpteval}} & \scriptsize{ ChatGPT} & \scriptsize{Multi-Turn Medical Dialogue} & \scriptsize{ZS} & \scriptsize{Chinese} & \scriptsize{No} \\ \hline 

\scriptsize{\cite{singhal2023towards}} & \scriptsize{GPT-4} & \scriptsize{Question Answering} & \scriptsize{FS} & \scriptsize{English} & \scriptsize{No} \\ \hline 

\scriptsize{\cite{wang2023chatgptmed}} & \scriptsize{ChatGPT} & \scriptsize{Question Answering} & \scriptsize{ZS} & \scriptsize{Chinese} & \scriptsize{-} \\ \hline 

\scriptsize{\cite{carpenter2023using}} & \scriptsize{GPT-3} & \scriptsize{Synonym Generation} & \scriptsize{ZS} & \scriptsize{English} & \scriptsize{-} \\ \hline 

\scriptsize{\cite{hernandez2023we}} & \scriptsize{GPT-3} & \scriptsize{Natural Language Inference, Question Answering, Text Classification} & \scriptsize{ZS} & \scriptsize{English} & \scriptsize{No} \\ \hline 

\scriptsize{\cite{rao2023assessing}} & \scriptsize{ChatGPT} & \scriptsize{Clinical Decision Support} & \scriptsize{ZS} & \scriptsize{English} & \scriptsize{-} \\ \hline 

\scriptsize{\cite{kung2023performance}} & \scriptsize{ChatGPT} & \scriptsize{Question Answering} & \scriptsize{ZS} & \scriptsize{	English} & \scriptsize{-} \\ \hline 

\scriptsize{\cite{hulman2023chatgpt}} & \scriptsize{ChatGPT} & \scriptsize{Question Answering} & \scriptsize{ZS} & \scriptsize{English} & \scriptsize{-} \\ \hline 

\scriptsize{\cite{hirosawa2023diagnostic}} & \scriptsize{ ChatGPT} & \scriptsize{Diagnosis Lists Generation} & \scriptsize{ZS} & \scriptsize{English} & \scriptsize{-} \\ \hline 

\scriptsize{\cite{liu2023assessing}} & \scriptsize{ ChatGPT} & \scriptsize{Clinical Decision Support} & \scriptsize{ZS} & \scriptsize{English} & \scriptsize{-} \\ \hline 

\scriptsize{\cite{gilson2023does}} & \scriptsize{GPT-3, GPT-3.5, ChatGPT} & \scriptsize{Question Answering} & \scriptsize{ZS} & \scriptsize{English} & \scriptsize{-} \\ \hline 

\scriptsize{\cite{antaki2023evaluating}} & \scriptsize{ChatGPT} & \scriptsize{Question Answering} & \scriptsize{ZS} & \scriptsize{English} & \scriptsize{-} \\ \hline 

\scriptsize{\cite{lyu2023translating}} & \scriptsize{ ChatGPT, GPT-4} & \scriptsize{Text Simplification} & \scriptsize{ZS} & \scriptsize{English} & \scriptsize{-} \\ \hline 

\end{tabular}}
\end{center}
\caption{ \label{gllms-healthcare}  Summary of research works exploring GLLMs for various NLP tasks in the healthcare domain. Here ZS represents zero-shot, and FS represents few-shot. Here '-' represents there is no comparison between GLLMs and domain-specific pretrained language models in the paper.} 
\end{table*}

Inspired by the success of pretrained language models like BERT, RoBERTa, ELECTRA, DeBERTa and T5  in the general domain, these models are also explored for domain-specific NLP tasks \cite{kalyan2021ammus}. However, the performance of general domain models is limited as these models are pretrained on general domain texts \cite{yang2020finbert, lee2020biobert}, and fine-tuning alone cannot provide enough domain knowledge \cite{kalyan2021ammus}. So, the research community focused on developing domain-specific pretrained language models either by continual pretraining or pretraining from scratch \cite{kalyan2021ammus, kalyan2022ammu}. Currently, domain-specific pretrained language models achieve state-of-the-art results in most tasks in specific domains like healthcare, finance, legal, social media, etc. 

GPT-3 family large language models achieve impressive performances in most NLP tasks in zero and few-shot settings in the general domain. Surprisingly, these models outperform fine-tuned pretrained language models in some tasks and achieve state-of-the-art results \cite{sun2023text, xu2023unleash,  wan2023gpt, ma2023large}. Inspired by the massive success of GLLMs in the general domain, the research community explored GLLMs in specific domains to assess how good these models are in domain-specific NLP tasks. Moreover, an extensive evaluation of these models in domain-specific tasks helps to arrive at valuable insights that will guide the research community to improve the performance further and increase the usage of these models in domain-specific NLP tasks.

\subsection{Healthcare Domain}
The recent works explored GLLMs for a variety of clinical NLP tasks like question answering \cite{holmes2023evaluating, Nori2023CapabilitiesOG, Tanaka2023PerformanceOG, liu2023benchmarking, kasai2023evaluating, moradi2021gpt, singhal2023towards, wang2023chatgptmed, hernandez2023we, kung2023performance, hulman2023chatgpt, gilson2023does, antaki2023evaluating},  text de-identification \cite{liu2023deid}, dialogue summarization \cite{giorgi2023wanglab, tang2023gersteinlab, nair2023generating}, named entity recognition \cite{chen2023large, gutierrez2022thinking}, relation extraction \cite{chen2023large}, text classification \cite{chen2023large, wang2023large, moradi2021gpt, hernandez2023we}, semantic similarity \cite{chen2023large, moradi2021gpt}, text simplification \cite{yang2023data, jeblick2022chatgpt, lyu2023translating}, relation classification \cite{gutierrez2022thinking, moradi2021gpt}, text summarization \cite{ma2023impressiongpt, shaib2023summarizing}, natural language inference \cite{wu2023exploring, wang2023large, moradi2021gpt, hernandez2023we}, word sense disambiguation \cite{agrawal2022large}, biomedical evidence extraction \cite{agrawal2022large}, coreference resolution \cite{agrawal2022large}, medical status extraction \cite{agrawal2022large}, medical attribute extraction \cite{agrawal2022large}, synonym generation \cite{carpenter2023using}, clinical decision support \cite{rao2023assessing, liu2023assessing} and diagnostic lists generation  \cite{hirosawa2023diagnostic}. Most of the research focused on English datasets, except a few focused on other languages like Japanese \cite{Tanaka2023PerformanceOG, kasai2023evaluating} and Chinese \cite{liu2023benchmarking, xu2023medgpteval, wang2023chatgptmed}. Table \ref{gllms-healthcare} presents a summary of research works exploring GLLMs for various NLP tasks in the healthcare domain.

Lyu et al. \cite{lyu2023translating} investigated the performance of ChatGPT and GPT-4 models in the healthcare domain, specifically the radiology area, by evaluating their ability to simplify the content in radiology reports. Experiment results showed that (i) GPT-4 performs better than ChatGPT. and (ii) optimized prompt with detailed instructions improves the performance for both models by a good margin. Antaki et al. \cite{antaki2023evaluating} evaluated the effectiveness of ChatGPT in answering Opthalmology questions. The test set consists of both easy and moderate-level questions. Experiment results showed that ChatGPT achieves an average accuracy of 49.25\%. Specifically, ChatGPT is able to answer the questions with good accuracy in general medicine. However,  its performance in specific sub-areas of Opthalmology is worst. Gilson et al. \cite{gilson2023does} evaluated GLLMs like GPT-3, GPT-3.5, and ChatGPT model in answering the medical questions in Step 1 and Step 2 exams of USMLE. Experiment results showed that ChatGPT outperforms the other two models by a good margin. Rao et al. \cite{rao2023assessing} demonstrated that ChatGPT performs better in the final diagnosis than in the initial diagnosis. This is because ChatGPT has access to more clinical data during the final diagnosis than the initial one. 

Carpenter et al. \cite{carpenter2023using} demonstrated that GPT-3 can be used for the synonym generation for drugs of abuse. The authors query GPT-3 repeatedly for each drug to generate multiple synonyms, which are later filtered. The generated synonyms are then used to build a lexicon that is helpful for pharmacovigilance on social media platforms. 
Inspired by the success of the GPT-3 model for text summarization in the general domain, Shaib et al. \cite{shaib2023summarizing} explored the GPT-3 model for summarizing biomedical documents. Experiment results revealed that (i) GPT-3 performance is promising in the case of single document summarization and (ii) GPT-3 struggles to summarize the content from multiple biomedical documents. Nair et al. \cite{nair2023generating} proposed a novel approach called “MEDSUM-ENT”, a multi-stage framework for clinical dialogue summarization. The proposed method leverages the GPT-3 model through multiple intermediate calls to extract medical entities from the conversations. In the final step of summarization, the extracted entities,  task instructions and in-context examples help the GPT-3 model to generate high-quality summaries. Based on the evaluation of radiology reports simplified by ChatGPT, Jeblick et al. \cite{jeblick2022chatgpt} reported that  ChatGPT-generated simplified radiology reports are not potentially harmful, complete and factually correct. However, further analysis reveals that some simplified reports contain factually incorrect sentences, potentially harmful paragraphs and a lack of essential medical findings.

\begin{table*}[h!]
\begin{center}
{\renewcommand{\arraystretch}{1.5}% for the vertical padding
\begin{tabular}{|p{0.6cm}|p{2cm}|p{6cm}|p{1.2cm}|p{2cm}|p{2cm}|} 
\hline

\scriptsize{\textbf{Paper}} & \scriptsize{\textbf{GLLMs Explored}} & \scriptsize{\textbf{Task(s)}} & \scriptsize{\textbf{Prompt Settings}} & \scriptsize{\textbf{Language(s)}} & \scriptsize{\textbf{Outperforms Domain-Specific Models}}\\ \hline

\scriptsize{\cite{ yu2022legal}} & \scriptsize{ GPT-3} & \scriptsize{Natural Language Inference} & \scriptsize{ZS, FS} & \scriptsize{English} & \scriptsize{-} \\ \hline 

\scriptsize{\cite{bommarito2022gpt}} & \scriptsize{GPT-3.5} & \scriptsize{Question Answering} & \scriptsize{ZS} & \scriptsize{English} & \scriptsize{-} \\ \hline 

\scriptsize{\cite{nguyen2023brief}} & \scriptsize{GPT-3} & \scriptsize{Question Answering, Text Generation} & \scriptsize{ZS} & \scriptsize{English} & \scriptsize{-} \\ \hline 

\scriptsize{\cite{chalkidis2023chatgpt}} & \scriptsize{ChatGPT} & \scriptsize{Text Classification} & \scriptsize{ZS, FS} & \scriptsize{English} & \scriptsize{No} \\ \hline 

\scriptsize{\cite{choi2023chatgpt}} & \scriptsize{ChatGPT} & \scriptsize{Question Answering, Text Generation} & \scriptsize{ZS} & \scriptsize{English} & \scriptsize{-} \\ \hline 

\end{tabular}}
\end{center}
\caption{ \label{gllms-legal}  Summary of research works exploring GLLMs for various NLP tasks in the legal domain. Here ZS represents zero-shot, and FS represents few-shot. Here '-' represents there is no comparison between GLLMs and domain-specific pretrained language models in the paper.} 
\end{table*}

Hirosawa et al. \cite{hirosawa2023diagnostic} investigated the effectiveness of ChatGPT for clinical diagnosis by evaluating its ability to generate accurate diagnosis lists for clinical vignettes with common chief complaints. Experimental results showed that ChatGPT can generate diagnosis lists with good accuracy. However, the accuracy rate of ChatGPT is still less than the accuracy rate of physicians. Wang et al. \cite{wang2023chatgptmed} evaluated the performance of the ChatGPT model in answering medical questions in the Chinese language. Here, ChatGPT is prompted with questions in both English and Chinese to avoid language barriers. Experimental results show that the performance of ChatGPT is much lower than the average performance of the medical students. For example, ChatGPT correctly answers 45.8\% of questions, while the average answering rate of medical students is 67.9\% in 2021.

Some of the research works demonstrated that domain-specific pretrained language models outperform GLLMs. Hernandez et al. \cite{hernandez2023we} compared the performance of the GPT-3 model with the performances of general and domain-specific pretrained language models on three healthcare NLP tasks: natural language inference, question answering and text classification. Experiment results showed that domain-specific pretrained language models achieve better results even though they are much smaller than GPT-3. Xu et al. \cite{xu2023medgpteval} introduced MedGPTEval, a benchmark to assess large language models in the healthcare domain. An extensive evaluation showed that domain-specific Chinese LLM outperforms general-purpose models like ChatGPT and ERNINE Bot. Singhal et al. \cite{singhal2023towards} introduced MedPaLM2, a healthcare domain-specific LLM obtained by domain-specific finetuning of the PaLM2 \cite{anil2023palm}  model. Experiment results showed that  MedPaLM2  outperforms few-shot GPT-4 and achieves new state-of-the-art results on the MultiMedQA benchmark. Moradi et al. \cite{moradi2021gpt} investigated the performances of BioBERT and GPT-3 in few-shot settings on five biomedical NLP tasks: text classification, natural language inference, question answering, relation extraction and semantic similarity. The authors observed that BioBERT and GPT-3 models underperform the model fine-tuned using full training data. Moreover, the BioBERT model outperforms GPT-3 in few-shot settings even though the BioBERT model is 514 times smaller than GPT-3.

Some research works showed that GLLMs can outperform domain-specific pretrained language models.  Ma et al. \cite{ma2023impressiongpt} proposed ImpressionGPT, a novel approach for summarizing radiology reports using ChatGPT. The proposed method involves dynamic prompt construction and iterative optimization to enhance the performance of ChatGPT further. Evaluation on two standard datasets showed that the proposed framework achieves new SOTA results outperforming fine-tuned models like ChestXrayBERT \cite{cai2021chestxraybert}. Liu et al. \cite{liu2023benchmarking}   introduced CMExam, a dataset with 60k+ multiple-choice medical questions in the Chinese language and evaluated GLLMs like GPT-3.5 and GPT-4 on answer prediction and answer reasoning tasks. The authors observed that GPT-4 achieves the best results for both tasks, outperforming GPT-3.5 and medical domain-specific Chinese LLMs like Huatuo [352] and DoctorGLM \cite{xiong2023doctorglm}.  Chen et al. \cite{chen2023large} explored GLLMs like GPT-3.5 and GPT-4 on eight datasets spanning four tasks in zero and few-shot settings. The authors observed that fine-tuned PubMedBERT outperforms both the GLLMs in all the biomedical tasks except question answering. In the case of biomedical question answering, GPT-4 outperforms the fine-tuned PubMedBERT model by a large margin of 17%. 

Giorgi et al. \cite{giorgi2023wanglab} explored models like Longformer Encoder-Decoder (LED) \cite{beltagy2020longformer} based on supervised fine-tuning and GLLMs like GPT-4 based on few-shot ICL for clinical dialogue summarization as a part of MEDIQA-Chat 2023 \cite{abacha2023overview} shared task. Here, the authors used Instructor \cite{su2022one} to select the most similar examples for few-shot ICL. Experiment results based on automatic metrics like BERTScore and ROUGE demonstrated that GPT-4 not only outperforms the LED model but also achieves first rank in the shared task. For medical text de-identification, Liu et al. \cite{liu2023deid} proposed a novel approach called “DeID-GPT”, a two-step approach based on GLLMs. In the first step, HIPAA identifiers are included in the prompt. In the second step, GLLM receives the prompt and the medical record based on which the model generates the de-identified medical record having the personal information masked.  The authors observed that GPT-4 outperforms not only ChatGPT but also fine-tuned models based on BERT, RoBERTa and ClinicalBERT.

\subsection{Legal Domain}
The recent works explored GLLMs for a variety of legal NLP tasks like natural language inference \cite{yu2022legal}, question answering \cite{bommarito2022gpt, lan2023chinese, choi2023chatgpt}, text generation \cite{nguyen2023brief, choi2023chatgpt}  and text classification \cite{chalkidis2023chatgpt}. Table \ref{gllms-legal} presents a summary of research works exploring GLLMs for various NLP tasks in the legal domain.  Bommarito et al. \cite{bommarito2022gpt} evaluated the performance of the GPT3.5 model in the legal domain by evaluating its ability to answer bar exam questions.  The model answers the questions correctly at a rate of 50\%, which is 25\% more than the random guess baseline. However, the model performance is almost 18\% less than the human performance, and overall model performance is below the passing threshold. Nguyen et al. \cite{nguyen2023brief}  presented LawGPT 1.0, the first-ever chatbot model based on GPT-3 for the legal domain. The GPT-3 model is pretrained on mostly generic corpus, so it lacks domain-specific knowledge. To add domain-specific knowledge, LawGPT is developed by fine-tuning the GPT-3 model on the law corpus. Experimental results showed that LawGPT 1.0 performs on par with existing legal assistants. 

\begin{table*}[h!]
\begin{center}
{\renewcommand{\arraystretch}{1.5}% for the vertical padding
\begin{tabular}{|p{0.6cm}|p{2cm}|p{6cm}|p{1.2cm}|p{2cm}|p{2cm}|} 
\hline

\scriptsize{\textbf{Paper}} & \scriptsize{\textbf{GLLMs Explored}} & \scriptsize{\textbf{Task(s)}} & \scriptsize{\textbf{Prompt Settings}} & \scriptsize{\textbf{Language(s)}} & \scriptsize{\textbf{Outperforms Domain-Specific Models}}\\ \hline

 \scriptsize{\cite{li2023chatgpt}} & \scriptsize{ ChatGPT, GPT-4} & \scriptsize{News Headlines Classification, Financial Sentiment Analysis, Named Entity Recognition, Question Answering} & \scriptsize{ZS} & \scriptsize{English} & \scriptsize{Yes} \\ \hline 
 
\scriptsize{\cite{fatouros2023transforming}} & \scriptsize{ChatGPT} & \scriptsize{Sentiment Analysis} & \scriptsize{ZS} & \scriptsize{English} & \scriptsize{Yes} \\ \hline 

\scriptsize{\cite{leippold2023sentiment}} & \scriptsize{GPT-3} & \scriptsize{Sentiment Analysis} & \scriptsize{ZS} & \scriptsize{English} & \scriptsize{No} \\ \hline 

\scriptsize{\cite{wiriyathammabhum2022promptshots}} & \scriptsize{GPT-3.5} & \scriptsize{Pairwise Ranking} & \scriptsize{FS} & \scriptsize{Chinese} & \scriptsize{-} \\ \hline 

\scriptsize{\cite{shah2023zero}} & \scriptsize{	ChatGPT} & \scriptsize{Sentiment Analysis, Claim Detection, Named Entity Recognition} & \scriptsize{ZS} & \scriptsize{English} & \scriptsize{No} \\ \hline 

\scriptsize{\cite{zhang2023fineval}} & \scriptsize{ChatGPT, GPT-4} & \scriptsize{Question Answering} & \scriptsize{ZS, FS} & \scriptsize{Chinese} & \scriptsize{-} \\ \hline 

\scriptsize{\cite{rajpoot2023gpt}} & \scriptsize{ChatGPT, GPT-4} & \scriptsize{Relation Extraction} & \scriptsize{FS} & \scriptsize{English} & \scriptsize{-} \\ \hline 

\scriptsize{\cite{lan2023chinese}} & \scriptsize{ChatGPT} & \scriptsize{Sentiment Analysis} & \scriptsize{ZS} & \scriptsize{Chinese} & \scriptsize{-} \\ \hline 

\scriptsize{\cite{loukas2023breaking}} & \scriptsize{GPT-3.5, GPT-4} & \scriptsize{Text Classification} & \scriptsize{ZS, FS} & \scriptsize{English} & \scriptsize{-} \\ \hline 

\end{tabular}}
\end{center}
\caption{ \label{gllms-finance}  Summary of research works exploring GLLMs for various NLP tasks in the finance domain. Here ZS represents zero-shot, and FS represents few-shot. Here '-' represents there is no comparison between GLLMs and domain-specific pretrained language models in the paper.} 
\end{table*}

Chalkidis et al. \cite{chalkidis2023chatgpt} investigated how effective ChatGPT is for legal text classification by evaluating the model performance on the LexGLUE \cite{chalkidis2022lexglue} benchmark, which consists of seven legal text classification datasets. The evaluation is performed in both zero and few-shot settings. Experiment results showed that ChatGPT performs poorly on legal text classification datasets. Choi et al. \cite{choi2023chatgpt} demonstrated that the performance of ChatGPT is just above the passing threshold, i.e., equivalent to a C+ grade student. The authors found that advanced prompts like CoT \cite{wei2022chain} and Ranking prompts performed worse or the same as simple prompts for multiple-choice questions.  For essay writing, the authors used carefully crafted simple prompts by including specific instructions at the end of the prompt.

\subsection{Finance Domain}
The recent works explored GLLMs for a variety of finance NLP tasks like text classification \cite{li2023chatgpt, loukas2023breaking}, sentiment analysis \cite{li2023chatgpt, leippold2023sentiment, shah2023zero, lan2023chinese}, named entity recognition \cite{li2023chatgpt, shah2023zero}, question answering \cite{li2023chatgpt, zhang2023fineval}, pairwise ranking \cite{wiriyathammabhum2022promptshots}, claim detection \cite{shah2023zero} and relation extraction \cite{rajpoot2023gpt}.  Table \ref{gllms-finance} presents a summary of research works exploring GLLMs for various NLP tasks in the finance domain. 

Li et al. \cite{li2023chatgpt} compared the performances of general LLMs like ChatGPT and GPT-4 in the finance domain with domain-specific models like BloombergGPT \cite{wu2023bloomberggpt} and small fine-tuned models like FinBERT \cite{araci2019finbert} and FinQANet \cite{chen2021finqa}. The evaluation is done on five different datasets related to four financial NLP tasks: news headlines classification, sentiment analysis, entity extraction, and question answering. The ChatGPT and GPT4 models do well in question-answering task but lag behind in tasks requiring domain-specific knowledge like entity extraction and sentiment analysis. Fatouros et al. \cite{fatouros2023transforming}  evaluated the effectiveness of ChatGPT for financial sentiment analysis by assessing its performance on the forex-related news headlines dataset. Experiment results showed that ChatGPT outperforms the domain-specific FinBERT \cite{liu2021finbert} model by a large margin of 35\% and also exhibits a high correlation with market returns.  

Leippold et al. \cite{leippold2023sentiment} explored GPT-3 for financial sentiment analysis and to generate adversarial attacks. Experiment results showed that FinBERT outperforms keyword-based approaches and the few-shot GPT-3 model in financial sentiment analysis. To study the robustness of FinBERT-based and keyword-based approaches, the authors explored GPT-3 to generate adversarial attacks. The main advantage of GPT-3 over existing adversarial attack-generating methods is that the model makes more subtle changes to the instances such that they are not noticeable to humans but still can fool the models. Wiriyathammabhum et al. \cite{wiriyathammabhum2022promptshots} explored instruction fine-tuned T5 and GPT-3.5 models to evaluate investments-related social media posts in Chinese. The task involves two subtasks, namely pairwise ranking and unsupervised ranking.  Experiment results showed that the few-shot prompted GPT-3.5 model outperforms the instruction fine-tuned T5 model and the few-shot prompted GPT-3.5 model with English-translated social media posts.

Shah et al. \cite{shah2023zero} compared the performance of ChatGPT with the performance of fine-tuned pretrained language models for three different financial NLP tasks: claim detection, sentiment analysis and named entity recognition. The authors observed that fine-tuned models outperform ChatGPT, but ChatGPT performs much better than some open-source LLMs.   Zhang et al. \cite{zhang2023fineval} introduced FinEval, a new benchmark to evaluate the financial domain of knowledge of LLMs in the Chinese language.  FinEval includes 4,661 multiple-choice questions in Chinese language from four different categories spanning 34 academic subjects. Experiment results showed that GPT-4 achieves around 70\% accuracy and outperforms all other LLMs, including ChatGPT and Chinese LLMs. 

Rajpoot et al. \cite{rajpoot2023gpt} assessed the effectiveness of  ChatGPT and GPT-4 for financial relation extraction in few-shot settings. As the choice of examples is crucial in few-shot ICL, the authors explored learning free and learning-based retriever for example selection. The authors observed that GPT-4 outperforms ChatGPT by a decent margin, and the learning-based retriever performs better than the learning-free retriever.

\section{Multilingual Performance of GLLMs}
\label{section-7}

\begin{table*}[h!]
\begin{center}
{\renewcommand{\arraystretch}{1.5}% for the vertical padding
\begin{tabular}{|p{0.6cm}|p{1.5cm}|p{6.5cm}|p{1.2cm}|p{3cm}|p{1.5cm}|} 
\hline
\scriptsize{\textbf{Paper}} & \scriptsize{\textbf{GLLMs explored}} & \scriptsize{\textbf{Task(s)}} & \scriptsize{\textbf{Prompt Settings}} & \scriptsize{\textbf{Language(s)}} & \scriptsize{\textbf{Domain(s)}} \\ \hline

\scriptsize{\cite{lai2023chatgpt}} & \scriptsize{ ChatGPT} & \scriptsize{	PoS Tagging, Entity Extraction, Relation Extraction, Natural Language Inference, Question Answering, Text Summarization, Common Sense Reasoning} & \scriptsize{ZS} & \scriptsize{37 Languages} & \scriptsize{	General} \\ \hline 

\scriptsize{\cite{fang2023chatgptgrammar}} & \scriptsize{ChatGPT} & \scriptsize{Grammar Error Correction} & \scriptsize{ZS, FS} & \scriptsize{English, German, Chinese} & \scriptsize{General} \\ \hline 

\scriptsize{\cite{armengol2022multilingual}} & \scriptsize{ GPT-3} & \scriptsize{Question Answering, Natural Language Generation, Text Summarization} & \scriptsize{ZS} & \scriptsize{German, Spanish, Russian, Turkish, Catalan} & \scriptsize{General} \\ \hline 

\scriptsize{\cite{ahuja2023mega}} & \scriptsize{ GPT-3.5, ChatGPT,  GPT-4} & \scriptsize{Natural Language Inference, Paraphrase Identification, Commonsense Reasoning, Question Answering, Parts of Speech Tagging, Sentiment Analysis, Text Summarization} & \scriptsize{ZS} & \scriptsize{70 languages} & \scriptsize{	General} \\ \hline 

\scriptsize{\cite{bang2023multitask}} & \scriptsize{ ChatGPT} & \scriptsize{Sentiment Analysis, Language Identification, Machine Translation} & \scriptsize{ZS} & \scriptsize{Multiple language including low resource languages like Sudanese, Javanese etc.} & \scriptsize{	General} \\ \hline 

\scriptsize{\cite{ kuzman2023chatgpt}} & \scriptsize{ ChatGPT} & \scriptsize{Genre Identification} & \scriptsize{ZS} & \scriptsize{English, Slovenian} & \scriptsize{General} \\ \hline 

\scriptsize{\cite{zhang2023don}} & \scriptsize{ ChatGPT} & \scriptsize{Question Answering, Reasoning} & \scriptsize{ZS} & \scriptsize{Six languages including Chinese, German and French} & \scriptsize{General} \\ \hline 

\scriptsize{\cite{das2023evaluating}} & \scriptsize{ChatGPT} & \scriptsize{Hate Speech Detection} & \scriptsize{ZS} & \scriptsize{Eleven languages including Hindi, Arabic and Italian} & \scriptsize{Social Media} \\ \hline 

\scriptsize{\cite{hada2023large}} & \scriptsize{ GPT-4} & \scriptsize{Three Text Generation Tasks} & \scriptsize{ZS} & \scriptsize{Ten languages including Chinese and Japanese.} & \scriptsize{General} \\ \hline 

\scriptsize{\cite{ leong2023bhasa}} & \scriptsize{ ChatGPT, GPT-4} & \scriptsize{Question Answering, Sentiment Analysis, Text Summarization, Named Entity Recognition, Toxicity Detection, Machine Translation, Natural Language Inference, Casual Reasoning} & \scriptsize{ZS, FS} & \scriptsize{Indonesian, Vietnamese, Thai, Tamil} & \scriptsize{General, Social Media, News} \\ \hline 

\end{tabular}}
\end{center}
\caption{ \label{gllms-multilingual}  Summary of research works exploring GLLMs for NLP tasks in multilingual settings. Here, ZS represents zero-shot, and FS represents few-shot. } 
\end{table*}

\textbf{Overview.} GLLMs are pretrained over large volumes of text data from multiple languages. For example, the corpus used to pretrain the GPT-3 model includes text from around 90 languages, and the percentage of English text is more than 90\% \cite{brown2020language, ahuja2023mega}. In the beginning, most of the research focused on assessing the performance of GLLMs on English datasets only. However, it is essential to evaluate these models on datasets from non-English languages, especially low-resource languages, to know how effective GLLMs are for non-English languages, and the insights gained from the comprehensive evaluation help to further improve these models towards non-English languages.

\textbf{Research works exploring GLLMs in multilingual settings.} Recently, some of the research works focused on evaluating GLLMs across various non-English languages. The evaluation is done on various tasks like parts of speech tagging \cite{lai2023chatgpt, ahuja2023mega}, named entity recognition \cite{lai2023chatgpt, leong2023bhasa}, relation extraction \cite{lai2023chatgpt}, natural language inference \cite{lai2023chatgpt,ahuja2023mega, leong2023bhasa}, question answering \cite{lai2023chatgpt, armengol2022multilingual, ahuja2023mega, zhang2023don, leong2023bhasa}, text summarization \cite{lai2023chatgpt, armengol2022multilingual, ahuja2023mega, leong2023bhasa}, commonsense reasoning \cite{lai2023chatgpt, ahuja2023mega}, grammar error correction \cite{fang2023chatgptgrammar}, text generation \cite{armengol2022multilingual, hada2023large}, paraphrase identification \cite{ahuja2023mega}, sentiment analysis \cite{ahuja2023mega, bang2023multitask, leong2023bhasa}, language identification \cite{bang2023multitask}, machine translation \cite{bang2023multitask, leong2023bhasa}, genre identification \cite{kuzman2023chatgpt}, hate speech detection \cite{das2023evaluating} and toxicity detection \cite{leong2023bhasa}.  Most of the research focused on general domain datasets, except a few focused on other domains like social media \cite{das2023evaluating, leong2023bhasa} and news \cite{leong2023bhasa}. Table \ref{gllms-multilingual} presents a summary of research works exploring GLLMs for NLP tasks in multilingual settings. 

Bang et al. \cite{bang2023multitask} presented an extensive multilingual evaluation of ChatGPT across three tasks: sentiment analysis, language identification and machine translation. When compared to English, the performance of ChatGPT degrades in the case of low-resource languages, particularly in the case of languages with non-Latin scripts. Das et al. \cite{das2023evaluating}  assessed the effectiveness of ChatGPT for emoji-based hate speech detection in multilingual settings. The authors reported that ChatGPT exhibits good performance but tends to misclassify abusive content as hate speech for non-English languages in the case of non-protected groups. Moreover, Armengol et al. \cite{armengol2022multilingual} reported that the performance of GPT-3 can be improved in the case of low-resource languages with optimized tokenization. 

The focus of existing benchmarks like HELM \cite{bommasani2023holistic} and BIG-Bench \cite{srivastava2023beyond} is on the English language. So, some of the research works focused on introducing new benchmarks to facilitate a systematic and comprehensive evaluation of the multilingual performance of GLLMs \cite{ahuja2023mega,leong2023bhasa }. For example, Ahuja et al. \cite{ahuja2023mega} presented MEGA, a comprehensive evaluation benchmarking having 16 datasets covering 70 languages. Based on the evaluation of GLLMs like GPT-3.5, ChatGPT and GPT-4, the authors reported that  GLLMs perform well in the case of languages with Latin scripts, and the performance is worst in the case of low-resource languages with non-Latin scripts across tasks. One of the possible reasons for this is the quality of tokenization. Similarly, Leong et al. \cite{leong2023bhasa} introduced BHASA, a benchmark to evaluate the performance of LLMs in four Southeast Asian languages. The benchmark consists of 20 datasets covering eight NLP tasks. The authors reported that (i) GPT-4 achieves better results compared to ChatGPT, and (ii) overall, the performance on some of the tasks is promising, with a lot of room for improvement in other tasks.

\begin{table*}[h!]
\begin{center}
{\renewcommand{\arraystretch}{1.5}% for the vertical padding
\begin{tabular}{|p{0.6cm}|p{2cm}|p{4.5cm}|p{1cm}|p{1.5cm}|p{1.5cm}|p{1.5cm}|} 
\hline
\scriptsize{\textbf{Paper}} & \scriptsize{\textbf{GLLMs Explored}} & \scriptsize{\textbf{Task(s)}} & \scriptsize{\textbf{Prompt Settings}} & \scriptsize{\textbf{Domain(s)}} & \scriptsize{\textbf{Language(s)}} & \scriptsize{\textbf{Outperforms Human Annotators}}\\ \hline

\scriptsize{\cite{gilardi2023chatgpt}} & \scriptsize{ChatGPT} & \scriptsize{	Stance, Relevance, Frame and Topics Detection} & \scriptsize{ZS} & \scriptsize{Social Media, News} & \scriptsize{English} & \scriptsize{Yes} \\ \hline 

\scriptsize{\cite{he2023annollm}} & \scriptsize{ GPT-3.5} & \scriptsize{Three Binary Text Classification Tasks} & \scriptsize{ZS, FS} & \scriptsize{General} & \scriptsize{	English} & \scriptsize{Yes} \\ \hline 

\scriptsize{\cite{tornberg2023chatgpt}} & \scriptsize{GPT-4} & \scriptsize{Political Tweets Classification} & \scriptsize{ZS} & \scriptsize{Social Media} & \scriptsize{English} & \scriptsize{Yes} \\ \hline 

\scriptsize{\cite{zhu2023can}} & \scriptsize{ ChatGPT} & \scriptsize{Stance Detection, Sentiment Analysis, Hate Speech Detection, Bot Detection} & \scriptsize{ZS} & \scriptsize{Social Media} & \scriptsize{English} & \scriptsize{No}  \\ \hline 

\scriptsize{\cite{li2023hot}} & \scriptsize{ ChatGPT} & \scriptsize{Detection of Hateful, Toxic and Offensive Comments} & \scriptsize{ZS} & \scriptsize{Social Media} & \scriptsize{English} & \scriptsize{No} \\ \hline 

\scriptsize{\cite{gu2023distilling}} & \scriptsize{ GPT-3.5, GPT-4} & \scriptsize{Adverse Drug Reaction Extraction} & \scriptsize{ZS, FS} & \scriptsize{Healthcare} & \scriptsize{English} & \scriptsize{-}   \\ \hline 

\scriptsize{\cite{wang2021want}} & \scriptsize{GPT-3} & \scriptsize{Text Entailment, Topic Classification, Sentiment Analysis, Answer Type Classification, Question Generation, Text Generation} & \scriptsize{ZS} & \scriptsize{General} & \scriptsize{English} & \scriptsize{-}  \\ \hline 

\scriptsize{\cite{ding2022gpt}} & \scriptsize{GPT-3} & \scriptsize{Sentiment Analysis, Relation Extraction, Named Entity Recognition} & \scriptsize{FS} & \scriptsize{General} & \scriptsize{English} & \scriptsize{-} 	\\ \hline 

\scriptsize{\cite{meoni2023large}} & \scriptsize{GPT-3.5} & \scriptsize{Named Entity Recognition} & \scriptsize{ZS} & \scriptsize{Healthcare} & \scriptsize{English, French, Spanish, Italian, Basque} & \scriptsize{-}  \\ \hline 

\scriptsize{\cite{xu2023inheritsumm}} & \scriptsize{ GPT-3.5} & \scriptsize{Text Summarization} & \scriptsize{ZS, FS} & \scriptsize{General} & \scriptsize{English} & \scriptsize{-} \\ \hline 

\scriptsize{\cite{alizadeh2023open}} & \scriptsize{ChatGPT} & \scriptsize{Detection of Stance, Topics, Relevance, General Frame and Policy Frame} & \scriptsize{ZS,FS} & \scriptsize{Social Media, News} & \scriptsize{English} & \scriptsize{Yes} \\ \hline 

\scriptsize{\cite{yang2023data}} & \scriptsize{GPT-3} & \scriptsize{Radiology Text Simplification} & \scriptsize{FS} & \scriptsize{Healthcare} & \scriptsize{English} & \scriptsize{-} \\ \hline 

\end{tabular}}
\end{center}
\caption{ \label{gllms-label}  Summary of research works exploring GLLMs for data labelling. Here, '-' represents that the paper doesn't include a comparison between GLLMs and human annotators. } 
\end{table*}

Some of the existing works demonstrated that using prompts in English improves the performance of GLLMs in the case of non-English languages \cite{lai2023chatgpt, kuzman2023chatgpt}. For example, Lai et al. \cite{lai2023chatgpt} performed a comprehensive evaluation of the multilingual abilities of ChatGPT on seven tasks covering more than 30 languages ranging from high-resource to extremely low-resource languages. The experiment results confirmed the bias of ChatGPT towards the English language, i.e., the performance is better for English compared to other languages and prompts in the English language can enhance the performance for non-English languages. The possible reason for the bias of GLLMs towards the English language is that GLLMs are trained mostly on English text corpus; hence, these models can better understand the prompt if it is in English \cite{kuzman2023chatgpt}.

Some of the research works investigated how GLLMs exhibit multilingual capabilities \cite{zhang2023don}  and how effective GLLM-based evaluators are in scaling up evaluation in multilingual settings \cite{hada2023large}. Zhang et al. \cite{zhang2023don} proposed a novel back translation prompting approach to systematically study how ChatGPT exhibit multilingual capabilities, although these models are largely pretrained on the English text corpus. The authors demonstrated that ChatGPT does translation in multilingual settings. Moreover, the multilingual performance of GLLMs is good only in the case of tasks which can translated. Hada et al. \cite{hada2023large} assessed the effectiveness of GPT-4 as an evaluator for natural language generation tasks in multilingual settings. The authors reported that GPT-4 tends to favour high scores and should be used carefully.

\section{Data Labelling and Data Augmentation Abilities of GLLMs}
\label{section-8}
\subsection{Data Labelling}

\textbf{Overview.} Large language models, specifically GLLMs, have achieved impressive performances in most of the NLP tasks, highlighting the huge potential of these models. However, large model size, high latency, high inference costs, proprietary access (in the case of GLLMs) and confidentiality concerns (in the case of sensitive domains like medical \cite{meoni2023large}) have become bottlenecks for the practical use of these models. Because of these bottlenecks, in environments with constrained resources or confidentiality constraints,  pretrained language models are preferred over  GLLMs as these models are much smaller in size and also more efficient compared to GLLMs \cite{thapa2023humans}. For example, BERT  base and large models contain just 110M and 340M parameters, while the GPT-3 model contains 175B parameters. Moreover, it is reported that GLLMs are trailing the SOTA models, with 4\% to 70\% lower performance when evaluated across a set of 25 diverse natural language processing tasks \cite{kocon2023chatgpt}.

The performance of fine-tuned pretrained language models is largely determined by the quality as well as the quantity of labelled data. Human-annotated data is considered the gold standard \cite{murthy2019twitsenti, van2021validity}, and we have two strategies for this \cite{gilardi2023chatgpt, tornberg2023chatgpt}. The first one is using trained expert coders like students and research assistants, and the second one is using crowd workers from online platforms like Amazon Mechanical Turk. Although human-labelled data is considered the gold standard, the human annotation process is expensive, laborious and time-consuming. The second strategy, i.e., using crowd workers, is comparatively less expensive, but there is a growing concern regarding the degrading annotation quality of crowd workers \cite{chmielewski2020mturk}. Moreover, the annotation quality varies with annotators, and hence it is consistent. To address the challenges associated with the human annotation process, there is a growing interest in the NLP research community to leverage the extraordinary generative abilities of GLLMs to make the data annotation process less expensive, faster and consistent. Similar to the human annotation process, GLLMs are provided with detailed instructions along with some labelled examples to label the data. 

\textbf{Research exploring GLLMs for data labelling. }The research community explored GLLMs for data labelling in a variety of NLP tasks like stance detection \cite{gilardi2023chatgpt, zhu2023can}, political tweets classification \cite{tornberg2023chatgpt}, sentiment analysis \cite{zhu2023can, wang2021want, ding2022gpt}, hate speech detection \cite{zhu2023can, li2023hot}, bot detection \cite{zhu2023can}, toxic comments detection \cite{li2023hot}, offensive comments detection \cite{li2023hot}, adverse drug reaction extraction \cite{gu2023distilling}, text entailment \cite{wang2021want}, topic classification \cite{wang2021want}, text generation \cite{wang2021want}, answer type classification \cite{wang2021want}, question generation \cite{wang2021want}, relation extraction \cite{ding2022gpt}, named entity recognition \cite{ding2022gpt, meoni2023large}, text summarization \cite{xu2023inheritsumm}, radiology text simplification \cite{yang2023data} etc. Most of the research works focused on English datasets, except a few research works focused on other languages like French \cite{meoni2023large}, Spanish \cite{meoni2023large}, Italian \cite{meoni2023large} and Basque \cite{meoni2023large}.  Table \ref{gllms-label} presents a summary of research works exploring GLLMs for data labelling.

Gu et al. \cite{gu2023distilling} labelled sentences from PubMed abstracts using the GPT-3.5 model and then fine-tuned the PubMedBERT model for adverse drug reaction extraction. Experiment results showed that (i) PubMedBERT achieves results comparable to the SOTA model and (ii) PubMedBERT outperforms the GPT-3.5 and GPT-4 models by large margins of 6 and 5 points in F1 score, respectively.  Based on the evaluation of multiple NLU and NLG tasks, Wang et al. \cite{wang2021want}   demonstrated that GPT-3 labelled data can result in a 50 to 96\% reduction in labelling expenses. Moreover, pretrained language models fine-tuned on GPT-3 labelled data outperform the few-shot GPT-3 model in both NLU and NLG tasks. Further, the authors proposed an approach based on active learning to make use of both human and GPT-3 labels, which further enhances the performance of the fine-tuned models. Meoni et al. \cite{meoni2023large}  investigated the effectiveness of GPT-3.5 labelled data and dictionary-based labelled data in fine-tuning pretrained language models to extract clinical entities in multiple languages like English, Spanish, Basque, Italian and French. The authors reported that (i) the performance of GPT-3.5 labelled data is on par with dictionary-based labelled data, and (ii) combining annotations from both approaches further enhances the results. Xu et al. \cite{xu2023inheritsumm} proposed InhertiSumm, a novel approach for training small text summarization models like ZCode++ \cite{he2022z} using GPT-3.5 generated summaries. The authors showed that the ZCode++ model with just 390M parameters trained using GPT-3.5 generated summaries performs on par with GPT-3.5 in zero and few-shot settings.

Zhu et al. \cite{zhu2023can} investigated how effective ChatGPT is for labelling data for social computing tasks. Based on the evaluation of five datasets spanning over tasks like stance detection, hate speech detection, bot detection and sentiment analysis, the authors reported that ChatGPT achieves an average accuracy of 60.9. Li et al. \cite{li2023hot}   investigated the ability of ChatGPT to label hateful, offensive and toxic comments and compared the performances with MTurk annotations. The authors observed that ChatGPT performance is promising as it is able to label 80\% of comments correctly. Moreover, the performance of ChatGPT is more consistent for non-harmful comments than harmful comments.

Some of the research works \cite{gilardi2023chatgpt, he2023annollm, tornberg2023chatgpt, alizadeh2023open} showed that GLLMs as data annotators can outperform human annotators. Gilardi et al. \cite{gilardi2023chatgpt} investigated the effectiveness of ChatGPT as an annotator in zero-shot settings for four text classification tasks involving tweets and news articles. The authors reported that ChatGPT is more effective than MTurk crowd-workers as (i) ChatGPT achieves 25 points more than crowd-workers in terms of accuracy, (ii) ChatGPT is approximately 30 times cheaper, and (iii) intercoder agreement of ChatGPT is more than crowd-workers. He et al. \cite{he2023annollm} proposed a novel approach called “explain then annotate” to enhance the performance of GLLMs as text data annotators. The proposed approach involves two steps: (i) GLLM generates explanations for the demonstrations and then (ii) annotates the data by leveraging annotation guidelines, demonstrations and explanations through CoT prompting. Evaluation on three binary text classification tasks revealed that GPT-3.5 outperforms crowd-workers on one task and matches the performance of crowd-workers on the other two tasks. Tornberg et al. \cite{tornberg2023chatgpt} demonstrated that zero-shot GPT-4 outperforms human annotators in labelling political English tweets. Further analysis demonstrated that GPT-4 possesses the ability to accurately label tweets that involve logical reasoning from contextual information.  Alizadeh et al. \cite{alizadeh2023open} compared the performances of GLLMs like ChatGPT, open-source LLMs like FLAN \cite{chung2022scaling} and MTurk annotators in labelling data (tweets and news articles) for five text classification tasks. The authors reported that ChatGPT achieves the best results, outperforming both open-source LLMs and MTurk annotators. One promising observation here is that open-source LLMs outperform MTurk annotators, and the performance is comparable to ChatGPT.

\begin{table*}[h!]
\begin{center}
{\renewcommand{\arraystretch}{1.5}% for the vertical padding
\begin{tabular}{|p{0.6cm}|p{2cm}|p{4.5cm}|p{1cm}|p{2.5cm}|p{1.5cm}|} 
\hline

\scriptsize{\textbf{Paper}} & \scriptsize{\textbf{GLLMs Explored}} & \scriptsize{\textbf{Task(s)}} & \scriptsize{\textbf{Prompt Settings}} & \scriptsize{\textbf{Domain(s)}} & \scriptsize{\textbf{Language(s)}} \\ \hline

\scriptsize{\cite{cegin2023chatgpt}} & \scriptsize{ ChatGPT} & \scriptsize{Intent Classification} & \scriptsize{ZS} & \scriptsize{General} & \scriptsize{English} \\ \hline 

\scriptsize{\cite{oh2023data}} & \scriptsize{ChatGPT} & \scriptsize{Machine Translation} & \scriptsize{ZS} & \scriptsize{General} & \scriptsize{Korean, German}  \\ \hline 

\scriptsize{\cite{sharma2022systematic}} & \scriptsize{GPT-3} & \scriptsize{Named Entity Recognition} & \scriptsize{ZS} & \scriptsize{News, Social Media, General, Healthcare} & \scriptsize{English} \\ \hline 

\scriptsize{\cite{guo2023dr}} & \scriptsize{ ChatGPT, GPT-4} & \scriptsize{Question Answering} & \scriptsize{ZS} & \scriptsize{Healthcare} & \scriptsize{English} \\ \hline 

\scriptsize{\cite{abaskohi2023lm}} & \scriptsize{GPT-3} & \scriptsize{Text Classification} & \scriptsize{FS} & \scriptsize{General} & \scriptsize{English} \\ \hline 

\scriptsize{\cite{sarker2023medical}} & \scriptsize{ ChatGPT} & \scriptsize{Medical Event Classification, Medication Identification} & \scriptsize{ZS} & \scriptsize{Healthcare} & \scriptsize{English} \\ \hline 

\scriptsize{\cite{parikh2023exploring}} & \scriptsize{GPT-3} & \scriptsize{Intent Classification} & \scriptsize{ZS} & \scriptsize{Social Media} & \scriptsize{English} \\ \hline 

\scriptsize{\cite{dai2023auggpt}} & \scriptsize{ChatGPT} & \scriptsize{Text Classification} & \scriptsize{ZS} & \scriptsize{General, Healthcare} & \scriptsize{English} \\ \hline 

\scriptsize{\cite{fang2023chatgpt}} & \scriptsize{ChatGPT} & \scriptsize{Open Intent Detection} & \scriptsize{ZS} & \scriptsize{General} & \scriptsize{English} \\ \hline 

\end{tabular}}
\end{center}
\caption{ \label{gllms-aug}  Summary of research works exploring GLLMs for paraphrasing-based data augmentation.} 
\end{table*}

\subsection{Data Augmentation}

\textbf{Overview.} The performance of downstream task-specific models is determined by the quality as well as the quantity of labelled data. Fine-tuning the pretrained language models on a small amount of labelled data will result in overfitting \cite{kalyan2021ammus} and, subsequently, poor performances. However, it is not feasible all the time to label a large number of instances as the annotation process is expensive. So, the research community focused on alternative approaches like data augmentation to increase the size of training sets in a relatively inexpensive way \cite{shorten2019survey, li2022data, liu2020survey, feng2021survey, bayer2022survey}. The data augmentation approaches focus on generating additional training instances either by making small changes to the existing instances or creating new instances with a distribution similar to the existing instances. 

Data augmentation is initially explored in the area of computer vision \cite{shorten2019survey}  and then explored in natural language processing \cite{li2022data, liu2020survey, feng2021survey, bayer2022survey}. When compared to computer vision, text data augmentation is more challenging because of the discrete nature of text. Data augmentation can be done at character, word and sentence levels. Character-level data augmentation approaches involve random deletion, addition, exchange or insertion of characters \cite{belinkov2018synthetic, coulombe2018text}.  For example, in the case of keyboard augmentation, a random character is replaced with its neighbour based on the QWERTY layout \cite{belinkov2018synthetic}. Similar to character-level data augmentation, word-level data augmentation approaches involve deletion, replacement, exchange or insertion of words at random positions \cite{wei2019eda, wang2015s}. Sentence-level approaches like back translation and paraphrasing generate augmented instances by rewriting the sentence \cite{sennrich2016improving, mallikarjuna2022question}. Overall, the main drawbacks of existing data augmentation approaches are (i) lack of sufficient diversity in the augmented instances and (ii)  often struggle to guarantee the accurate labelling of the augmented data \cite{dai2023auggpt}. To address these drawbacks, the research community focused on leveraging the exceptional generating abilities of GLLMs for data augmentation to ensure sufficient diversity and correct labelling in the augmented data. 

\subsubsection{Paraphrasing}
\textbf{Research works exploring GLLMs for paraphrasing-based data augmentation.} The research community explored GLLMs for paraphrasing in various NLP tasks like intent classification \cite{cegin2023chatgpt, parikh2023exploring, fang2023chatgpt}, machine translation \cite{oh2023data}, named entity recognition \cite{sharma2022systematic}, question answering \cite{guo2023dr}, medical event classification \cite{sarker2023medical}, medication identification \cite{sarker2023medical} etc. GLLM-based paraphrasing is explored in multiple domains like general \cite{cegin2023chatgpt, oh2023data, sharma2022systematic, abaskohi2023lm, dai2023auggpt, fang2023chatgpt}, news \cite{sharma2022systematic}, social media \cite{sharma2022systematic, parikh2023exploring} and healthcare \cite{sharma2022systematic, guo2023dr, sarker2023medical, dai2023auggpt}. Table \ref{gllms-aug} presents a summary of research works exploring GLLMs for paraphrasing-based data augmentation.

\begin{table*}[h!]
\begin{center}
{\renewcommand{\arraystretch}{1.5}% for the vertical padding
\begin{tabular}{|p{0.6cm}|p{2cm}|p{4cm}|p{1cm}|p{3.6cm}|p{2cm}|} 
\hline
\scriptsize{\textbf{Paper}} & \scriptsize{\textbf{GLLMs Explored}} & \scriptsize{\textbf{Task(s)}} & \scriptsize{\textbf{Prompt Settings}} & \scriptsize{\textbf{Domain(s)}} & \scriptsize{\textbf{Language(s)}} \\ \hline

\scriptsize{\cite{zhan2023socialdial}} & \scriptsize{ ChatGPT} & \scriptsize{Text Classification} & \scriptsize{ZS} & \scriptsize{Social Media} & \scriptsize{Chinese}  \\ \hline 

\scriptsize{\cite{wang2023umass_bionlp}} & \scriptsize{ ChatGPT} & \scriptsize{ Note2Dialogue Generation} & \scriptsize{ ZS} & \scriptsize{ Healthcare} & \scriptsize{ English} \\ \hline 

\scriptsize{\cite{Gunasekar2023TextbooksAA}} & \scriptsize{ GPT-3.5} & \scriptsize{ Training Phi-1 LLM} & \scriptsize{ ZS} & \scriptsize{ Programming} & \scriptsize{ English} \\ \hline 

\scriptsize{\cite{Whitehouse2023LLMpoweredDA}} & \scriptsize{ChatGPT, GPT-4} & \scriptsize{ Cross-lingual Common Sense Reasoning} & \scriptsize{ FS} & \scriptsize{ General } & \scriptsize{ Multiple Languages} \\ \hline 

\scriptsize{\cite{ hartvigsen2022toxigen}} & \scriptsize{ GPT-3} & \scriptsize{ Hate Speech Detection} & \scriptsize{ FS} & \scriptsize{ Social Media} & \scriptsize{ English} \\ \hline 

\scriptsize{ \cite{markov2023holistic}} & \scriptsize{GPT-3} & \scriptsize{ Undesired Context Detection} & \scriptsize{ ZS, FS} & \scriptsize{ Social Media} & \scriptsize{ English} \\ \hline 

\scriptsize{ \cite{Guo2023DrLI}} & \scriptsize{ ChatGPT, GPT-4} & \scriptsize{ Question Answering} & \scriptsize{ ZS} & \scriptsize{ Healthcare} & \scriptsize{ English} \\ \hline 

\scriptsize{ \cite{parikh2023exploring}} & \scriptsize{GPT-3} & \scriptsize{ Intent Classification} & \scriptsize{ ZS} & \scriptsize{ General} & \scriptsize{ English } \\ \hline 

\scriptsize{\cite{eldan2023tinystories}} & \scriptsize{  GPT-3.5, GPT-4} & \scriptsize{ Training Smaller LLMs} & \scriptsize{ ZS} & \scriptsize{ General} & \scriptsize{ English}  \\ \hline 

\scriptsize{ \cite{xu2023unleash}} & \scriptsize{ GPT-3.5} & \scriptsize{ Relation Extraction} & \scriptsize{ FS} & \scriptsize{ General,  Scientific Literature} & \scriptsize{ English} \\ \hline 

\scriptsize{ \cite{liu2023logicot}} & \scriptsize{ GPT-4} & \scriptsize{ CoT Instruction Tuning} & \scriptsize{ FS} & \scriptsize{ General} & \scriptsize{ English}  \\ \hline 

\scriptsize{ \cite{peng2023instruction}} & \scriptsize{GPT-4} & \scriptsize{ Instruction Tuning} & \scriptsize{ ZS} & \scriptsize{ General} & \scriptsize{ English, Chinese} \\ \hline 

\scriptsize{ \cite{malkiel2023gpt}} & \scriptsize{ GPT-3} & \scriptsize{ Call segmentation, Topic extraction} & \scriptsize{ ZS} & \scriptsize{ Dialogue} & \scriptsize{ English} \\ \hline 

\scriptsize{\cite{wahle2022large}} & \scriptsize{GPT-3} & \scriptsize{Paraphrase Detection} & \scriptsize{ZS} & \scriptsize{General,  Scientific Literature} & \scriptsize{English} \\ \hline 

\scriptsize{\cite{michail2023uzh_clyp}} & \scriptsize{ ChatGPT} & \scriptsize{Tweet Intimacy Prediction} & \scriptsize{FS} & \scriptsize{Social Media} & \scriptsize{Multiple Languages} \\ \hline 

\scriptsize{\cite{tang2023does}} & \scriptsize{ ChatGPT} & \scriptsize{Named Entity Recognition, Relation Classification} & \scriptsize{ZS} & \scriptsize{Healthcare} & \scriptsize{English} \\ \hline 

\scriptsize{\cite{yu2023large}} & \scriptsize{ChatGPT} & \scriptsize{Topic Classification} & \scriptsize{ZS} & \scriptsize{News, Social Media} & \scriptsize{English} \\ \hline 

\scriptsize{\cite{yang2023neural}} & \scriptsize{ChatGPT} & \scriptsize{Neural Machine Translation} & \scriptsize{ZS} & \scriptsize{General} & \scriptsize{Multiple Languages} \\ \hline 

\scriptsize{\cite{zhao2023robut}} & \scriptsize{GPT-3, Codex} & \scriptsize{Table Question Answering} & \scriptsize{ZS} & \scriptsize{General} & \scriptsize{English} \\ \hline 

\scriptsize{\cite{xu2023instructscore}} & \scriptsize{GPT-4} & \scriptsize{Text Generation Evaluation} & \scriptsize{ZS} & \scriptsize{General} & \scriptsize{Multiple Languages} \\ \hline 

\end{tabular}}
\end{center}
\caption{ \label{gllms-gen}  Summary of research works exploring GLLMs for data generation-based data augmentation. Here ZS represents zero-shot and FS represents few-shot. } 
\end{table*}

Cegin et al. \cite{cegin2023chatgpt} compared the quality of paraphrases generated by ChatGPT and crowd workers for intent classification. The authors reported that (i) ChatGPT generates more diversified paraphrases compared to crowd-workers and (ii) the robustness of models fine-tuned on ChatGPT is comparable to the models fine-tuned on crowd-workers generated paraphrases. Oh et al. \cite{oh2023data} explored ChatGPT-based data augmentation to generate additional training instances to fine-tune the mBART-50 model \cite{tang2020multilingual} for machine translation involving Korean-German language pairs. Here, the authors explored three different prompting strategies, out of which the storytelling prompting approach achieves the best results and improves the BLUE score by 0.68. Here, the storytelling prompting approach involves generating a three-sentence story based on the source sentence and then translating each of these sentences into the target language. Abaskohi et al.  \cite{abaskohi2023lm}  proposed a novel approach based on prompt-based tuning and contrastive learning to fine-tune pretrained language models for text classification. As contrastive learning requires data augmentation, the authors explored models like GPT-3 and OPT-175B \cite{zhang2022opt} for paraphrasing. Experiment results showed that GPT-3 based paraphrasing outperforms existing data augmentation approaches like back translation \cite{sugiyama2019data} and easy data augmentation \cite{wei2019eda}.

To overcome the problem of limited training instances for EHR analysis, sarker et al. \cite{sarker2023medical}   explored ChatGPT to generate additional training instances through paraphrasing. Experiments on medication event classification and medical identification tasks revealed that fine-tuning the pretrained language models on ChatGPT augmented training set enhances the performance.  Dai et al. \cite{dai2023auggpt} proposed AugGPT, a ChatGPT-based approach to generate additional training instances by paraphrasing existing training instances for few-shot classification. Experiments on general and medical domain text classification datasets revealed that AugGPT outperforms all the existing data augmentation approaches by a good margin. Further analysis showed that AugGPT generates more diversified instances while preserving the original labels.

Paraphrasing-based data augmentation for entity extraction is challenging because of the difficulty in preserving span-level labels. Sharma et al. \cite{sharma2022systematic} explored GPT-3 models, back translation and PEGASUS-based paraphraser for synthetic data generation using paraphrasing.  The authors observed that the larger GPT-3 variant with inline annotations achieves the best results for entity extraction across datasets from multiple domains.

\subsubsection{Data Generation}
\textbf{Research works exploring GLLMs for data generation-based data augmentation.} The research community explored GLLMs for data generation-based data augmentation in various NLP tasks like dialogue generation \cite{wang2023umass_bionlp}, training smaller LLMs \cite{Gunasekar2023TextbooksAA, eldan2023tinystories}, common sense reasoning \cite{Whitehouse2023LLMpoweredDA}, hate speech detection \cite{hartvigsen2022toxigen}, undesired content detection \cite{markov2023holistic}, question answering \cite{Guo2023DrLI, zhao2023robut}, intent classification \cite{parikh2023exploring},  relation extraction \cite{xu2023unleash, tang2023does}, instruction tuning \cite{liu2023logicot, peng2023instruction}, paraphrase detection \cite{wahle2022large}, tweet intimacy prediction \cite{michail2023uzh_clyp}, named entity recognition \cite{tang2023does}, machine translation \cite{yang2023neural} etc. GLLM-based data generation for data augmentation is explored in multiple domains like general \cite{Whitehouse2023LLMpoweredDA, parikh2023exploring, eldan2023tinystories, xu2023unleash, liu2023logicot, peng2023instruction, wahle2022large, yang2023neural, zhao2023robut, xu2023instructscore}, social media \cite{zhan2023socialdial, hartvigsen2022toxigen, markov2023holistic, michail2023uzh_clyp, yu2023large}, news \cite{yu2023large}, scientific literature  \cite{xu2023unleash, wahle2022large}, healthcare \cite{wang2023umass_bionlp, Guo2023DrLI, tang2023does}, dialogue \cite{malkiel2023gpt}, programming \cite{Gunasekar2023TextbooksAA} etc. Table \ref{gllms-gen} presents a summary of research works exploring GLLMs for data generation-based data augmentation.

Some of the research works explored GLLMs for data generation-based data augmentation in various text classification tasks 
 \cite{zhan2023socialdial, hartvigsen2022toxigen, markov2023holistic, parikh2023exploring, michail2023uzh_clyp, yu2023large}. For example, Hartvigsen et al. \cite{hartvigsen2022toxigen}  used GPT-3 with demonstration-based prompting to create a large-scale synthetic dataset for the detection of implicit hate speech. Here, the authors explored a variant of constrained beam search to ensure subtle toxicity in the generated examples. Michail et al. \cite{michail2023uzh_clyp}  investigated the effectiveness of ChatGPT-generated synthetic data to fine-tune multilingual models for tweet intimacy prediction in the case of languages with no labelled instances.  Here, ChatGPT is prompted with instructions and examples from a high-resource language and asked to generate new examples in the target language.  Most of the existing research works use simple prompts for data generation, limiting the diversity of the generated synthetic data. To address this, Yu et al. \cite{yu2023large}  proposed a novel approach that leverages attributed prompts for data generation to increase the diversity in the generated data. Based on the evaluation on four topic classification datasets, the authors observed that (i) the proposed approach enhances the model performance and (ii) reduces the querying cost of ChatGPT by a large margin.

Some of the research works explored GLLMs for data generation-based data augmentation in various information extraction tasks like relation extraction \cite{xu2023unleash},  relation classification \cite{tang2023does} and named entity recognition \cite{tang2023does}. Xu et al. \cite{xu2023unleash} evaluated how effective is the GPT-3.5 model for relation classification. To address the data scarcity problem in few-shot settings, the authors used the GPT-3.5 model to generate additional data. The prompt used for data generation consists of instance descriptions along with some example instances. Tang et al. \cite{tang2023does} used ChatGPT in zero-shot settings to generate synthetic data for tasks like named entity recognition and relation classification in the healthcare domain. The authors showed that the model fine-tuned on this synthetic data outperforms zero-shot ChatGPT by a large margin in both tasks. 

Some of the research works explored GLLMs for data generation in LLM development stages, like LLM pretraining \cite{Gunasekar2023TextbooksAA, eldan2023tinystories} and instruction tuning \cite{liu2023logicot, peng2023instruction}. Gunasekar et al. \cite{Gunasekar2023TextbooksAA} trained Phi-1, a code LLM using GPT-3.5 generated synthetic textbook and code data. Here, the training corpus includes 1B tokens of GPT-3.5 generated Python textbook and code data along with 6B tokens of code data from the web. Eldan et al. \cite{eldan2023tinystories} explored GLLMs like  GPT-3.5 and GPT-4 models to generate TinyStories, a synthetic dataset of stories with only the words understood by typical 3 to 4-year-old kids. The authors demonstrated that the GLLM generated dataset can be used to train smaller LLMs, which can generate coherent and consistent stories with near-perfect grammar. Instruction tuning requires large human-annotated datasets, which are often difficult to obtain. Stanford Alpaca \footnote{https://crfm.stanford.edu/2023/03/13/alpaca.html} and Vicuna \footnote{https://lmsys.org/blog/2023-03-30-vicuna/} showed the effectiveness of synthetic instruction tuning datasets generated using GPT-3.5 and ChatGPT, respectively. Inspired by the success of these models, Peng et al. \cite{peng2023instruction} explored advanced models like GPT-4 to generate instruction-tuning datasets in English and Chinese languages. The experiment results showed that GPT-4 generated instruction tuning datasets further enhance the zero-shot performance of LLaMA models. Liu  et al. \cite{liu2023logicot} used GPT-4 to generate LogiCoT, a synthetic dataset of CoT rationales. This dataset can be used for instruction tuning the LLMs to enhance their logical reasoning abilities.

\section{Detecting GLLM Generated Text}
\label{section-9}

\textbf{Overview.} GLLMs demonstrated extraordinary human-like capabilities to understand user queries, follow the instructions and then answer the user queries with high-quality content. Apart from responding to user queries, these models can also generate news articles, research papers, code and essays with human-like fluency. With the ability to generate text with human-like fluency, these models are widely adopted in a variety of real-world applications like writing assistants,  coding assistants, chatbots, etc \cite{Mireshghallah2023SmallerLM}.  Although there is a lot of excitement about GLLMs and their applications in recent times, there are also growing concerns regarding the potential misuse of these models for illegal activities \cite{Guo2023HowCI}, such as fake news on social media platforms \cite{hacker2023regulating, de2023chatgpt}, fake reviews on e-commerce websites \cite{Mitrovic2023ChatGPTOH},  fake research papers \cite{gao2023comparing}, academic fraud \cite{cotton2023chatting}, etc. For example, these models can be easily used by malicious users to create fake news \cite{hacker2023regulating, de2023chatgpt} and propagate on social platforms at a large scale to exaggerate or manipulate the facts to get an undue advantage, especially during political campaigns.  Similarly, students can use these models to write their assignments or generate code for their projects \cite{cotton2023chatting}, and GLLM generated fake research papers \cite{gao2023comparing} can have a serious impact on the scientific community as these papers are written without conducting any experiments. 

There is a strong need for the development of approaches to detect GLLM generated text, as there are growing concerns regarding the misuse of GLLMs. Such approaches help to distinguish the GLLM generated text from human-generated text and verify the source as well as the authenticity of the information. However, detecting GLLM generated text is more challenging as models like ChatGPT and GPT-4 can generate content with human-like fluency. 

\textbf{Research exploring the detection of GLLM generated text.} To avoid misuse and ensure the safe use of these models, the research community focused on developing approaches to identify the GLLM generated text accurately. The recent research works explored the detection of GLLM generated text in multiple domains like scientific literature \cite{theocharopoulos2023detection, zaitsu2023distinguishing, yu2023cheat, yang2023dna}, academic \cite{liu2023argugpt, orenstrakh2023detecting}, healthcare \cite{liao2023differentiate, zhan2023g3detector, yang2023dna}, news \cite{clark2021all}, legal \cite{zhan2023g3detector, Guo2023HowCI}, social media \cite{yang2023dna, Mitrovic2023ChatGPTOH}, Finance \cite{Guo2023HowCI} etc. Most of the research works focused on the English language, while a few research works focused on other languages like Japanese \cite{zaitsu2023distinguishing}, German \cite{yang2023dna} and Spanish \cite{orenstrakh2023detecting}. Table \ref{gllms-gllmtext} presents a summary of research works exploring the detection of GLLM generated text.

\begin{table*}[h!]
\begin{center}
{\renewcommand{\arraystretch}{1.5}% for the vertical padding
\begin{tabular}{|p{0.6cm}|p{2cm}|p{4.5cm}|p{1.5cm}|p{1.5cm}|p{1.5cm}|p{1.5cm}|} 
\hline
\scriptsize{\textbf{Paper}} & \scriptsize{\textbf{Detect}} & \scriptsize{\textbf{Approach}} & \scriptsize{\textbf{Satisfactory Performance}} & \scriptsize{\textbf{Training Free}} & \scriptsize{\textbf{Domain(s)}} & \scriptsize{\textbf{Language(s)}}\\ \hline

\scriptsize{\cite{pegoraro2023chatgpt}} & \scriptsize{ChatGPT generated text} & \scriptsize{Evaluate multiple online tools} & \scriptsize{No} & \scriptsize{-} & \scriptsize{Multiple domains} & \scriptsize{English} \\ \hline 

\scriptsize{\cite{theocharopoulos2023detection}} & \scriptsize{GPT-3 generated text} & \scriptsize{Classifiers based on machine learning models like LR, SVM and deep learning models like LSTM and BERT} & \scriptsize{Yes} & \scriptsize{No} & \scriptsize{Scientific Literature} & \scriptsize{English}  \\ \hline 

\scriptsize{\cite{zaitsu2023distinguishing}} & \scriptsize{ChatGPT and GPT-4 generated text} & \scriptsize{Classifier based on random forest and stylometric features} & \scriptsize{Yes} & \scriptsize{No} & \scriptsize{Scientific Literature} & \scriptsize{Japanese} \\ \hline

\scriptsize{\cite{liu2023argugpt}} & \scriptsize{GPT-3 and ChatGPT generated text} & \scriptsize{Classifier based on models like SVM and RoBERTa} & \scriptsize{Yes} & \scriptsize{No} & \scriptsize{Academic} & \scriptsize{English} \\ \hline

\scriptsize{\cite{yu2023cheat}} & \scriptsize{ChatGPT generated text} & \scriptsize{Classifier based on models like RoBERTa} & \scriptsize{No} & \scriptsize{No} & \scriptsize{Scientific Literature} & \scriptsize{English} \\ \hline

\scriptsize{\cite{liao2023differentiate}} & \scriptsize{ ChatGPT generated text} & \scriptsize{Classifier based on models like BERT} & \scriptsize{Yes} & \scriptsize{No} & \scriptsize{Healthcare} & \scriptsize{English} \\ \hline

\scriptsize{\cite{orenstrakh2023detecting}} & \scriptsize{ChatGPT generated text} & \scriptsize{Evaluate multiple online tools} & \scriptsize{Yes} & \scriptsize{-} & \scriptsize{Academic} & \scriptsize{English, Spanish} \\ \hline

\scriptsize{\cite{clark2021all}} & \scriptsize{GPT-3 generated text} & \scriptsize{Evaluate human evaluators} & \scriptsize{No} & \scriptsize{-} & \scriptsize{Stories, News, Recipies} & \scriptsize{English} \\ \hline

\scriptsize{\cite{zhan2023g3detector}} & \scriptsize{ChatGPT and GPT-4 generated text} & \scriptsize{Classifier based on models like BERT and RoBERTa} & \scriptsize{Yes} & \scriptsize{No} & \scriptsize{Law, Medical, Dialogue, General} & \scriptsize{English} \\ \hline

\scriptsize{\cite{yang2023dna}} & \scriptsize{GPT-3.5, ChatGPT and GPT-4 generated text} & \scriptsize{Training free divergent N-gram Analysis} & \scriptsize{Yes} & \scriptsize{Yes} & \scriptsize{Healthcare, Social Media, Scientific Literature} & \scriptsize{English, German} \\ \hline

\scriptsize{\cite{shi2023red}} & \scriptsize{ChatGPT generated text} & \scriptsize{Evaluate the robustness of existing detectors} & \scriptsize{No} & \scriptsize{-} & \scriptsize{General} & \scriptsize{English}  \\ \hline

\scriptsize{\cite{khalil2023will}} & \scriptsize{ ChatGPT generated text} & \scriptsize{Evaluate existing plagiarism tools} & \scriptsize{No} & \scriptsize{-} & \scriptsize{General} & \scriptsize{English} \\ \hline

\scriptsize{\cite{he2023mgtbench}} & \scriptsize{ChatGPT generated text} & \scriptsize{	Propose benchmark and evaluate existing detectors} & \scriptsize{Yes} & \scriptsize{-} & \scriptsize{General} & \scriptsize{English} \\ \hline 

\scriptsize{\cite{Mitrovic2023ChatGPTOH}} & \scriptsize{ ChatGPT generated text} & \scriptsize{Propose novel approach based on DistilBERT and SHAP to detect and explain} & \scriptsize{Yes} & \scriptsize{No} & \scriptsize{Social Media} & \scriptsize{English}  \\ \hline

\scriptsize{\cite{Guo2023HowCI}} & \scriptsize{ ChatGPT generated text} & \scriptsize{Introduce new dataset and evaluate multiple existing detection models} & \scriptsize{Yes} & \scriptsize{-} & \scriptsize{General, Finance, Healthcare, Legal , Psychology} & \scriptsize{English} \\ \hline 

\scriptsize{\cite{Wang2023BotOH}} & \scriptsize{GPT-3 and ChatGPT-based bots} & \scriptsize{Propose FLAIR to detect online GPT-3 and ChatGPT-based bots} & \scriptsize{	Yes} & \scriptsize{Yes} & \scriptsize{General} & \scriptsize{English} \\ \hline

\scriptsize{\cite{ Chen2023GPTSentinelDH}} & \scriptsize{ ChatGPT generated text} & \scriptsize{Classifiers based on models like RoBERTa and T5} & \scriptsize{Yes} & \scriptsize{No} & \scriptsize{General} & \scriptsize{English} \\ \hline

\scriptsize{\cite{Mireshghallah2023SmallerLM}} & \scriptsize{ChatGPT generated text} & \scriptsize{Propose a zero-shot approach based on local optimality} & \scriptsize{Yes} & \scriptsize{Yes} & \scriptsize{General} & \scriptsize{English} \\ \hline

\scriptsize{\cite{Yu2023GPTPT}} & \scriptsize{ChatGPT generated text} & \scriptsize{Propose an approach based on Siamese Network and binary classifier} & \scriptsize{Yes} & \scriptsize{No} & \scriptsize{General} & \scriptsize{English} \\ \hline

\scriptsize{\cite{Yang2023IsCI}} & \scriptsize{ChatGPT polished text} & \scriptsize{Trains classifier and polish ratio models to detect and explain} & \scriptsize{Yes} & \scriptsize{No} & \scriptsize{General} & \scriptsize{English} \\ \hline 

\scriptsize{\cite{krishna2023paraphrasing}} & \scriptsize{GPT-3.5 generated text} & \scriptsize{Evaluate robustness using paraphrase attacks} & \scriptsize{No} & \scriptsize{-} & \scriptsize{General} & \scriptsize{English} \\ \hline

\end{tabular}}
\end{center}
\caption{ \label{gllms-gllmtext}  Summary of research works exploring the detection of GLLM generated text.} 
\end{table*}

Some of the research works focused on assessing the effectiveness of the existing machine-generated text detection tools to detect GLLM generated text. A number of online tools are available, ranging from simple classifiers based on logistic regression to advanced classifiers based on pretrained language models to detect ChatGPT-generated text. To assess the effectiveness of these tools, Pegoraro et al. \cite{pegoraro2023chatgpt}  introduced a dataset having ChatGPT-generated responses for questions from various domains like finance, medicine, etc., and user-generated responses from social media platforms. The comprehensive evaluation showed that the maximum success rate of these tools is less than 50\% only, which leaves a lot of room for improvement. Orenstrakh et al. \cite{orenstrakh2023detecting} evaluated the effectiveness of eight popular detectors using three metrics, namely resilience, false positives and accuracy. The authors observed that CopyLeaks, GPTKit and GLTR  achieve the best results for the metrics accuracy, false positives and resilience. However, all these detectors struggle with non-English languages and paraphrased LLM-generated text. There is a lack of comprehensive evaluation benchmark to detect machine-generated text as the existing approaches use different models, datasets and settings. To address this, He et al. \cite{he2023mgtbench}  proposed MGTBench, the first machine-generated text detection benchmark. Evaluation on this benchmark showed that, except for the ChatGPT detector \cite{Guo2023HowCI} and LM detector \cite{ippolito2020automatic}, the performance of other detectors is not satisfactory. Guo et al. \cite{Guo2023HowCI}  introduced the HC3 dataset, having human-authored and ChatGPT-generated responses to questions from multiple domains like legal, healthcare, finance, psychology, etc. The performance of existing detection approaches on the HC3 dataset is just satisfactory, and linguistic analysis showed that human-authored answers are short in length but use a large vocabulary compared to ChatGPT-generated answers. 

Some of the research works focused on developing approaches based on trained classifier models to detect GLLM generated text.  Theocharopoulos et al. \cite{theocharopoulos2023detection} evaluated the effectiveness of classifiers based on models like logistic regression, support vector machine, LSTM, and BERT to identify GPT-3 generated scientific abstracts. The LSTM-based classifier with word2vec embeddings achieves an accuracy of more than 98\% and outperforms other classifiers. Zaitsu et al. \cite{zaitsu2023distinguishing} observed that LLM-generated texts differ significantly from human-written texts in terms of stylometric features.  The authors demonstrated that random forest trained with different stylometric features can identify the LLM-generated Japanese text with 100\% accuracy. Liu et al. \cite{liu2023argugpt} reported that fine-tuned RoBERTa model achieves an accuracy of more than 90\% on the AruGPT dataset of human-written and GLLM generated argumentative essays. Moreover, linguistic analysis revealed that GLLM generated texts tend to be more complex syntactically, while human-generated texts are lexically more complex. To facilitate the development of a ChatGPT-written abstract detector, Yu et al. \cite{yu2023cheat}  introduced CHEAT, a large dataset of ChatGPT and human-written abstracts. Based on the evaluation of multiple existing approaches like ZeroGPT, OpenAI detector, ChatGPT-detector-roberta \cite{Guo2023HowCI} and ChatGPT-qa-detector-roberta \cite{Guo2023HowCI}, the authors reported that performance is away from satisfactory and the human involvement further increases the detection difficulty. Zhan et al. \cite{zhan2023g3detector} treated the detection of LLM generated as a binary classification problem and proposed a novel approach based on fine-tuned RoBERTa model. The authors reported that the proposed approach exhibits good performance and also has the ability to detect the text generated using a detection evasion technique. Mitrovic et al. \cite{Mitrovic2023ChatGPTOH}  proposed a novel approach based on DistilBERT \cite{sanh2019distilbert} and SHAP \cite{lundberg2017unified} to detect the machine-generated text and explain the reasoning. The proposed approach achieves an accuracy of 79\%, and based on the explanations, the authors observed that ChatGPT-generated text maintains a polite tone, lacks specific details and generally refrains from expressing emotions.

Chen et al. \cite{Chen2023GPTSentinelDH} introduced OpenGPTText, which includes ChatGPT-generated paraphrased text. The authors reported that fine-tuned classifiers based on models like RoBERTa and T5 can achieve impressive results in detecting ChatGPT-generated text with an accuracy of more than 97\%.  Yu et al. \cite{Yu2023GPTPT} introduced GPT-Pat, a novel approach based on ChatGPT, a Siamese network and binary classifier, to detect machine-generated text effectively. The proposed approach enhances the SOTA accuracy by more than 12\% and also exhibits better robustness to attacks like re-translation and text polishing. Yang et al. \cite{Yang2023IsCI} focused on detecting GLLM-polished text, which is more challenging and useful in real-world applications. The proposed approach involves training a classification model to identify the machine-generated text and a polish ratio (regression) model to explain the ChatGPT involvement. A Polish ratio of 0.2 indicates ChatGPT involvement and a value of more than 0.6 represents the text is entirely ChatGPT generated.

\begin{table*}[h!]
\begin{center}
{\renewcommand{\arraystretch}{1.5}% for the vertical padding
\begin{tabular}{|p{0.6cm}|p{2cm}|p{4.8cm}|p{1.2cm}|p{2.4cm}|p{1.5cm}|p{1.5cm}|} 
\hline

\scriptsize{\textbf{Paper}} & \scriptsize{\textbf{GLLMs Explored}} & \scriptsize{\textbf{Task(s)}} & \scriptsize{\textbf{Prompt Settings}} & \scriptsize{\textbf{Robustness}} & \scriptsize{\textbf{Domain(s)}} & \scriptsize{\textbf{Language(s)}}\\ \hline

\scriptsize{\cite{chen2023robust}} & \scriptsize{ GPT-3, GPT-3.5} & \scriptsize{Nine NLU Tasks} & \scriptsize{ZS, FS} & \scriptsize{Adversarial Input} & \scriptsize{General} & \scriptsize{English} \\ \hline 

\scriptsize{\cite{wang2023robustness}} & \scriptsize{ GPT-3.5, ChatGPT} & \scriptsize{Four NLU Tasks, Machine Translation} & \scriptsize{ZS} & \scriptsize{Out of Distribution} & \scriptsize{General, Medical} & \scriptsize{English} \\ \hline 

\scriptsize{\cite{zhuo2023robustness}} & \scriptsize{ Codex} & \scriptsize{Semantic Parsing} & \scriptsize{ZS, FS} & \scriptsize{Adversarial Input} & \scriptsize{Programming} & \scriptsize{English} \\ \hline 

\scriptsize{\cite{zhu2023promptbench}} & \scriptsize{ChatGPT} & \scriptsize{Eight Tasks including Four NLU tasks} & \scriptsize{ZS, FS} & \scriptsize{Adversarial Prompt}	& \scriptsize{General} & \scriptsize{English} \\ \hline 

\scriptsize{\cite{shirafuji2023exploring}} & \scriptsize{Codex, InstructGPT, ChatGPT} & \scriptsize{Code Generation} & \scriptsize{ZS} & \scriptsize{Adversarial Prompt} & \scriptsize{Programming} & \scriptsize{English} \\ \hline 

\scriptsize{\cite{zhao2023robut}} & \scriptsize{GPT-3, Codex} & \scriptsize{Table Question Answering} & \scriptsize{FS} & \scriptsize{Adversarial Input} & \scriptsize{General} & \scriptsize{English} \\ \hline 

\scriptsize{\cite{han2023information}} & \scriptsize{ChatGPT} & \scriptsize{Fourteen IE Tasks} & \scriptsize{ZS, FS} & \scriptsize{Adversarial Prompt} & \scriptsize{General} & \scriptsize{English} \\ \hline 

\scriptsize{\cite{liu2023evaluating}} & \scriptsize{ChatGPT, GPT-4} & \scriptsize{Question Answering} & \scriptsize{ZS, FS} & \scriptsize{Out-of-Distribution} & \scriptsize{General} & \scriptsize{English} \\ \hline 

\scriptsize{\cite{liu2023comprehensive}} & \scriptsize{ChatGPT} & \scriptsize{Text-to-SQL Generation} & \scriptsize{ZS} & \scriptsize{Adversarial Input} & \scriptsize{General} & \scriptsize{English} \\ \hline 
							
\end{tabular}}
\end{center}
\caption{ \label{gllms-robust}  Summary of research works exploring GLLMs robustness to out-of-distribution instances, adversarial prompts and adversarial inputs. Here ZS represents zero-shot, and FS represents few-shot.} 
\end{table*}

Training-based approaches to detect LLM-generated text have limited flexibility, especially when used for new domains \cite{yang2023dna}. To overcome this drawback, some of the research works focused on developing training-free approaches to detect GLLM generated text. Yang et al. \cite{yang2023dna} proposed DNA-GPT, a training-free approach based on divergent n-gram analysis. With the proposed approach, the authors achieved SOTA results on both English and German datasets. Wang et al. \cite{Wang2023BotOH} proposed a novel framework called FLAIR to detect LLM-based bots with a single question in an effective way.  The results showed that the proposed approach is effective and a good alternative to existing CAPTCHA-based approaches. Mireshghallah et al. \cite{Mireshghallah2023SmallerLM} investigated whether models other than the generator can be used to identify machine-generated text. In general, smaller models serve as more effective universal text detectors. These models exhibit better accuracy in identifying text produced by both small and larger models. For example, OPT-125M achieves better results compared to the GPT-J 6B model in detecting ChatGPT-generated text. 

Some of the research works focused on assessing the robustness of machine-generated text detectors towards different attacks. Shi et al. \cite{shi2023red} evaluated the robustness of existing detectors using attacks like synonym word replacement and writing style modification. The authors implemented both attacks using LLMs. The results showed that the existing detectors are not robust to the attacks, which emphasizes the need for more robust and reliable detectors to detect and avoid the misuse of LLMs. Krishna et al. \cite{krishna2023paraphrasing} showed that existing detectors like OpenAI detector,  GPTZero and DetectGPT \cite{mitchell2023detectgpt} are not robust to paraphrase attacks. For example, paraphrase attacks result in a drop of more than 65\% accuracy in the case of DetectGPT. 

Some of the research works focused on assessing the effectiveness of humans in identifying GLLM generated text. For example, Clark et al. \cite{clark2021all} observed that non-expert evaluators are unable to differentiate GPT-3 generated text from human-authored text in three different domains, namely news, recipes and stories. The reason for this is the evaluators arrived at their decisions based on surface-level features without considering the advanced text generation capabilities of the GPT-3 model.

\section{Robustness of GLLMs}
\label{section-10}

\textbf{Overview.} GPT-3 family large language models achieve impressive performances in zero and few-shot settings in many NLP tasks. In some tasks like text classification \cite{sun2023text}, relation extraction \cite{wan2023gpt}, etc. GLLMs without any explicit fine-tuning outperform state-of-the-art fine-tuned models. For example, Sun et al. \cite{sun2023text} demonstrated that InstructGPT, with the advanced prompting strategy, achieves SOTA results using just 16 examples on four text classification datasets. Similarly, Wan et al. \cite{wan2023gpt} achieved SOTA results in relation extraction with the GPT-RE framework. However, to increase the reliability of these models in real-world applications, especially in critical domains like medicine, it is essential to systematically study the robustness of these models in various scenarios. Adversarial robustness refers to the model’s ability to maintain good performance even in the case of deliberately crafted instances \cite{goyal2022survey, qiu2022adversarial}. These instances are called adversarial instances and are carefully designed by making subtle changes in the original inputs to deceive the model. Out-of-distribution (OOD) instances refer to examples that differ significantly from the data distribution used to train the model \cite{shen2021towards}. These instances fall outside the range of the model's training data and present challenges to the model's performance and generalization ability. Some of the recent research works focused on evaluating the robustness of GLLMs to out-of-distribution instances \cite{wang2023robustness, liu2023evaluating}, adversarial prompts \cite{zhu2023promptbench, shirafuji2023exploring, han2023information} and adversarial inputs \cite{chen2023robust, zhuo2023robustness, zhao2023robut, liu2023comprehensive}  in one or more natural language processing tasks. Table \ref{gllms-robust} presents a summary of research works assessing GLLMs robustness to out-of-distribution instances, adversarial prompts and adversarial inputs.

\textbf{Research works exploring GLLMs robustness.} Some of the research works evaluated the robustness of GLLMs in specific tasks like semantic parsing \cite{zhuo2023robustness}, code generation \cite{shirafuji2023exploring}, table question answering \cite{zhao2023robut}, multi-choice question answering \cite{liu2023evaluating} and text-to-SQL generation \cite{liu2023comprehensive}. Zhuo et al. \cite{zhuo2023robustness} reported that Codex-based semantic parsers are not robust to adversarial examples, and the robustness can be enhanced using few-shot in-context learning. Shirafuji et al. \cite{shirafuji2023exploring} studied the robustness of GPT-3 family models like Codex, InstructGPT, and ChatGPT to adversarial prompts in code generation task. The authors observed that InstructGPT and ChatGPT exhibit better robustness compared to Codex. However, there is much room for improvement, indicating that quality code generation requires well-designed prompts. Zhao et al. \cite{zhao2023robut} proposed RobuT, a benchmark to systematically study the robustness of large language models to adversarial inputs in table question answering. The authors reported that GLLMs like GPT-3 and Codex exhibit better robustness than fine-tuned models. Moreover, the authors demonstrated that GLLM generated adversarial inputs can enhance the adversarial robustness of fine-tuned models. Liu et al. \cite{liu2023evaluating} reported that ChatGPT and GPT-4 perform well in multiple choice question answering but struggle to answer out-of-distribution questions. Liu et al. \cite{liu2023comprehensive} showed that ChatGPT exhibits impressive zero-shot performance in Text-to-SQL generation. Moreover, ChatGPT demonstrates better robustness to adversarial inputs than SOTA models in text-to-SQL generation. 

Some of the research works evaluated the GLLM robustness in multiple natural language understanding and generation tasks \cite{chen2023robust, wang2023robustness, zhu2023promptbench, han2023information}. Chen et al. \cite{chen2023robust}  assessed the robustness of GPT-3 and GPT-3.5 models on 21 datasets covering nine natural language understanding tasks. Here the authors used adversarial text transformations from TextFlint \cite{wang2021textflint}. The authors observed that the models are robust in tasks like machine reading comprehension and exhibit performance degradation of more than 35\% in tasks like sentiment analysis and natural language inference. Wang et al. \cite{wang2023robustness} evaluated the robustness of GPT-3.5 and ChatGPT models on adversarial and out-of-distribution (OOD) samples on nine datasets covering four NLU tasks and machine translation. The authors observed that ChatGPT exhibits good performances on adversarial and OOD samples, but still, there is much room for improvement. 

Zhu et al. \cite{zhu2023promptbench} developed PromptBench, a benchmark with more than 4k adversarial prompts to evaluate the robustness of large language models to adversarial prompts. The benchmark covers 13 datasets spanning eight tasks, including four NLU tasks. The authors observed that GLLMs are not robust to adversarial prompts. Moreover, word-level attacks are the most effective, which results in a performance drop of more than 30\%. Based on the evaluation of ChatGPT on fourteen information extraction sub-tasks, Han et al. \cite{han2023information} showed that ChatGPT is vulnerable to adversarial prompts, i.e., the performance is greatly affected by including irrelevant context in the prompt.

\section{GLLMs as Evaluators}
\label{section-11}
\textbf{Overview.} Natural language processing tasks can be broadly classified into natural language understanding (NLU) and natural language generation (NLG). NLU involves the interpretation of text, while NLG involves generating human-like text. The evaluation of NLU outputs is pretty straightforward, while the evaluation of NLG outputs is challenging because of the diversity and inherent complexity of the text \cite{chen2023exploring}. Moreover, the NLG evaluation involves assessing the generated text outputs in multiple dimensions, such as coherence, fluency, naturalness and semantic consistency. Human evaluation and automatic evaluation are two existing approaches for NLG evaluation. The human evaluation depends on competent annotators for an accurate and reliable assessment \cite{sai2022survey}.

\begin{table*}[h!]
\begin{center}
{\renewcommand{\arraystretch}{1.5}% for the vertical padding
\begin{tabular}{|p{0.6cm}|p{2cm}|p{5.3cm}|p{1.2cm}|p{1.2cm}|p{1.5cm}|p{2cm}|} 
\hline

\scriptsize{\textbf{Paper}} & \scriptsize{\textbf{GLLMs Explored}} & \scriptsize{\textbf{Task(s)}} & \scriptsize{\textbf{Prompt Settings}} & \scriptsize{\textbf{References Required}} & \scriptsize{\textbf{Domain(s)}} & \scriptsize{\textbf{Language(s)}}\\ \hline

\scriptsize{\cite{zhuo2023large}} & \scriptsize{ChatGPT} & \scriptsize{Code Generation} & \scriptsize{ZS} & \scriptsize{Optional} & \scriptsize{Programming} & \scriptsize{Five Programming Languages} \\ \hline 

\scriptsize{\cite{ lai2023multidimensional}} & \scriptsize{ChatGPT} & \scriptsize{Text Style Transfer} & \scriptsize{ZS} & \scriptsize{Yes} & \scriptsize{General} & \scriptsize{English} \\ \hline 

\scriptsize{\cite{liu2023gpteval}} & \scriptsize{ChatGPT, GPT-4} & \scriptsize{Text Summarization, Dialogue Generation} & \scriptsize{ZS} & \scriptsize{No} & \scriptsize{General} & \scriptsize{English} \\ \hline 

\scriptsize{\cite{kocmi2023large}} & \scriptsize{GPT, GPT-3.5, ChatGPT, GPT-4} & \scriptsize{Machine Translation} & \scriptsize{ZS} & \scriptsize{Optional} & \scriptsize{General} & \scriptsize{English, German, Chinese, Russian} \\ \hline 

\scriptsize{\cite{chen2023exploring}} & \scriptsize{GPT-3.5, ChatGPT} & \scriptsize{Text Summarization, Dialogue  Generation, Story Generation, Paraphrase Generation} & \scriptsize{ZS} & \scriptsize{No} & \scriptsize{General} & \scriptsize{English} \\ \hline 

\scriptsize{\cite{ lu2023error}} & \scriptsize{GPT-3.5, ChatGPT} & \scriptsize{Machine Translation} & \scriptsize{ZS, FS} & \scriptsize{Yes} & \scriptsize{General} & \scriptsize{English, Chinese, German} \\ \hline 

\scriptsize{\cite{luo2023chatgpt}} & \scriptsize{ChatGPT} & \scriptsize{Text Summarization} & \scriptsize{ZS} & \scriptsize{No} & \scriptsize{General} & \scriptsize{English} \\ \hline 

\scriptsize{\cite{shen2023large}} & \scriptsize{ChatGPT} & \scriptsize{Text Summarization} & \scriptsize{ZS} & \scriptsize{No} & \scriptsize{General} & \scriptsize{English} \\ \hline 

\scriptsize{\cite{lu2023llmscore}} & \scriptsize{GPT-4} & \scriptsize{Text-to-Image Synthesis} & \scriptsize{ZS} & \scriptsize{N/A} & \scriptsize{General} & \scriptsize{English} \\ \hline 

\scriptsize{\cite{xu2023instructscore}} & \scriptsize{GPT-4} & \scriptsize{Machine Translation} & \scriptsize{ZS} & \scriptsize{Yes} & \scriptsize{General} & \scriptsize{English, German, Russian} \\ \hline 

\scriptsize{\cite{fu2023gptscore}} & \scriptsize{GPT-3, GPT-3.5} & \scriptsize{Dialogue Generation, Machine Translation, Text Summarization, Data-to-Text Generation} & \scriptsize{ZS, FS} & \scriptsize{No} & \scriptsize{General} & \scriptsize{English, Chinese} \\ \hline 

\scriptsize{\cite{liu2023learning}} & \scriptsize{GPT-3, ChatGPT} & \scriptsize{Text Summarization} & \scriptsize{ZS} & \scriptsize{No} & \scriptsize{General} & \scriptsize{English} \\ \hline 

\scriptsize{\cite{gao2023human}} & \scriptsize{ChatGPT} & \scriptsize{Text Summarization} & \scriptsize{ZS} & \scriptsize{No} & \scriptsize{General} & \scriptsize{English} \\ \hline 

\scriptsize{\cite{ tang2023not}} & \scriptsize{GPT-3.5} & \scriptsize{Machine Translation, Text Summarization, Image Caption} & \scriptsize{ZS} & \scriptsize{Yes} & \scriptsize{General} & \scriptsize{English} \\ \hline 

\scriptsize{\cite{wang2023largeeval}} & \scriptsize{ChatGPT, GPT-4} & \scriptsize{Text Generation} & \scriptsize{ZS} & \scriptsize{No} & \scriptsize{General} 	& \scriptsize{English} \\ \hline 

\scriptsize{\cite{ jain2023multi}} & \scriptsize{GPT-3.5} & \scriptsize{Text Summarization} & \scriptsize{ZS} & \scriptsize{No} & \scriptsize{General} & \scriptsize{English} \\ \hline 

\scriptsize{\cite{wang2023chatgpt2}} & \scriptsize{ChatGPT} & \scriptsize{Text Summarization, Story Generation, Data-to-Text Generation} & \scriptsize{ZS} & \scriptsize{Optional} & \scriptsize{General} & \scriptsize{English} \\ \hline 

\scriptsize{\cite{bai2023benchmarking}} & \scriptsize{GPT-4} & \scriptsize{Open-ended Question Answering} & \scriptsize{ZS} & \scriptsize{No} & \scriptsize{General} & \scriptsize{English} \\ \hline 

\scriptsize{\cite{yang2023bigtrans}} & \scriptsize{GPT-4} & \scriptsize{Machine Translation} & \scriptsize{ZS} & \scriptsize{Yes} & \scriptsize{General} & \scriptsize{Multiple Languages} \\ \hline 

\scriptsize{\cite{zheng2023judging}} & \scriptsize{GPT-4} & \scriptsize{Open-ended Question Answering} & \scriptsize{ZS} & \scriptsize{No} & \scriptsize{General} & \scriptsize{English} \\ \hline 

\end{tabular}}
\end{center}
\caption{ \label{gllms-evaluator}  Summary of research works exploring GLLM-based evaluation for natural language generation tasks. Here ZS represents zero-shot, and FS represents few-shot.} 
\end{table*}

\textbf{Human Evaluation vs Automatic Evaluation.} Human evaluation is treated as the gold standard, but it is time-consuming, expensive, difficult to scale, inconsistent, and not reproducible \cite{chen2023exploring,wang2023largeeval}. To address the issues with human evaluation, automatic evaluation metrics are developed, which fall broadly into two categories: n-gram-based and embedding-based. N-gram-based metrics assess the quality based on the lexical overlap between the generated and reference texts. Some of the commonly used n-gram-based metrics are BLEU \cite{papineni2002bleu}, ROUGE \cite{lin2004rouge} and METEOR \cite{banerjee2005meteor}.  However, these metrics have a poor correlation with human scores because of their inability to capture semantic meaning \cite{kocmi2021ship}. Later, with the evolution of transformers and pretrained language models, the researchers developed embedding-based metrics like BERTScore \cite{zhang2019bertscore}, MoverScore \cite{zhao2019moverscore}, BARTScore \cite{yuan2021bartscore}, CodeBERTScore \cite{zhou2023codebertscore} etc. These metrics leverage the pretrained language models and assess the quality based on the semantic similarity between the generated and reference text. The main drawback of the existing automatic evaluation metrics is the requirement for references, which are difficult to obtain, especially in low-resource domains. Moreover, with just a few references, it is not possible to get an accurate and reliable assessment as few references cannot account for all the semantic variations \cite{chen2023exploring}. So, there is a strong need for automatic evaluation metrics which are reference-free.

\textbf{GLLM-based Evaluation.} Recently, with the huge success of GLLMs in most of the NLP tasks, the research community focused on developing automatic evaluation metrics based on these models. These models possess the ability of in-context learning, while instruction tuning enables these models to align themselves with human evaluation \cite{ouyang2022training}. These two abilities enable these models to imitate the behaviour of human evaluators, who typically evaluate natural language generation task outputs by understanding instructions and the given examples. The GLLM-based evaluation metrics demonstrate a strong correlation with human scores even in the absence of reference outputs \cite{liu2023gpteval,fu2023gptscore}. Table \ref{gllms-evaluator} presents a summary of research works exploring GLLM-based evaluation for various natural language generation tasks.

\textbf{Research works exploring GLLM-based evaluation.} The NLP researchers proposed various GLLM-based evaluation frameworks to evaluate the outputs of various NLG tasks like code generation \cite{zhuo2023large}, text style transfer \cite{lai2023multidimensional}, text summarization \cite{liu2023gpteval, chen2023exploring, luo2023chatgpt, shen2023large, fu2023gptscore, liu2023learning, gao2023human, tang2023not, jain2023multi, wang2023chatgpt2}, dialogue generation \cite{liu2023gpteval, chen2023exploring, fu2023gptscore}, machine translation \cite{kocmi2023large, lu2023error, xu2023instructscore, fu2023gptscore, tang2023not, yang2023bigtrans}, story generation \cite{chen2023exploring, wang2023chatgpt2}, paraphrase generation \cite{chen2023exploring}, text-to-image synthesis \cite{lu2023llmscore},  data-to-text generation \cite{ fu2023gptscore, wang2023chatgpt2}, image captioning \cite{tang2023not}, text generation \cite{wang2023largeeval}, open-ended question answering \cite{bai2023benchmarking, zheng2023judging}. Most of the research works proposed evaluation frameworks using direct prompting, while some of the research works introduced evaluation frameworks based on advanced prompting strategies like chain-of-thoughts \cite{zhuo2023large,liu2023gpteval} and error analysis prompting \cite{lu2023error}. Some of the proposed evaluation frameworks work with and without references \cite{zhuo2023large, kocmi2023large, wang2023chatgpt2}, while some of them require references \cite{lai2023multidimensional, lu2023error, xu2023instructscore, tang2023not, yang2023bigtrans}, and some don’t require any references \cite{liu2023gpteval, chen2023exploring, luo2023chatgpt, shen2023large, fu2023gptscore, liu2023learning, gao2023human, wang2023largeeval, jain2023multi, bai2023benchmarking, zheng2023judging}. 

Lai et al. \cite{lai2023multidimensional} investigated how effective ChatGPT is to evaluate text style transfer task along three dimensions: fluency, content and style. The model achieves good correlations with human judgements, and the best results are obtained by using separate prompts for each dimension evaluation. Kocmi et al. \cite{kocmi2023large}  proposed GEMBA, a GPT-based metric to assess translation output quality, with references being optional. The authors reported that GPT-3.5 and higher models are only useful for the assessment, and GPT-4 achieves the best results. Based on the evaluation of four natural language generation tasks, paraphrase generation, text summarization, story generation and dialogue response generation, Chen et al. \cite{chen2023exploring} showed that explicit score with greedy decoding strategy is the best way to assess NLG outputs using GLLMs like ChatGPT.  Luo et al. \cite{luo2023chatgpt} evaluated ChatGPT’s ability as a factual inconsistency evaluator for text summarization task. Experiment results showed that ChatGPT outperforms existing metrics on most of the datasets.  

Shen et al. \cite{shen2023large} explored how effective ChatGPT can be as a zero-shot evaluator for abstractive summarization systems using different evaluation methods like likert scaling \cite{he2022ctrlsum} and head-to-head comparisons \cite{shen2022sentbs}. Extensive analysis showed that likert scaling implemented as a multiple-choice question gives the best and most stable results.  Liu et al. \cite{liu2023learning} designed a novel approach which uses BRIO \cite{liu2022brio}, a contrastive learning-based method, to train smaller models like BART for text summarization and metrics like GPTScore \cite{fu2023gptscore} or GPTRank for evaluation. The contrastive learning training method helps the model to effectively utilize the supervision signal offered by the reference LLMs. The evaluation showed that the proposed approach helps the smaller model to outperform LLMs like GPT-3 and ChatGPT. 

Gao et al. \cite{gao2023human} evaluated ChatGPT for text summarization using various human evaluation methods and reported that (i) ChatGPT-based evaluation is both cost-effective and reproducible, unlike human evaluation, (ii) the performance of ChatGPT-based evaluation is highly dependent on the prompt design, and (iii) ChatGPT generated explanations correlates with its scores. Jain et al. \cite{jain2023multi} explored the effectiveness of the GPT-3.5 model as a multi-dimensional evaluator of text summarization. The authors reported that using in-context learning, GPT-3.5-based evaluation achieves SOTA performances on factual consistency and relevance dimensions. Based on the evaluation of five datasets covering text summarization, story generation and data-to-text generation, Wang et al. \cite{wang2023chatgpt2} reported that ChatGPT as an evaluator (i) exhibits good correlations with human scores, especially in the case of story generation task and (ii) is prompt sensitive. Bai et al. \cite{bai2023benchmarking} introduced a novel evaluation framework called Language-Model-as-an-Examiner to evaluate open-ended questions. In this framework, GLLM acts as a knowledgeable examiner, generates questions using its own knowledge and then does the reference-free evaluation. Yang et al. \cite{yang2023bigtrans} developed the BigTrans model (based on LLaMA -13B model) with a multilingual translation capacity of more than 100 languages. GPT-4 based assessment showed that BigTrans performance is on par with ChatGPT and Google translate. Zheng et al. \cite{zheng2023judging} explored GPT-4 as a judge to evaluate open-ended question answering using two newly introduced benchmarks MT-Bench and Chatbot Arena. The experiment results showed that GPT-4 achieves more than 80% agreement. 

Unlike the above-discussed research works, which used direct prompting, some of the works explored advanced prompting to offer better guidance and context for the GLLM evaluator. Zhuo et al. \cite{zhuo2023large} developed a code generation evaluation framework based on ChatGPT and demonstrated that the proposed framework outperforms CodeBERTScore \cite{zhou2023codebertscore} consistently across multiple programming languages. Moreover, the performance of the evaluation framework can be enhanced using references and zero-shot CoT prompting. Liu et al. \cite{liu2023gpteval} proposed G-EVAL, a novel framework based on GPT-4 for the assessment of natural language generation tasks. The proposed framework uses CoT prompting and a form-filling paradigm. Here, CoT prompting enhances the performance of G-EVAL by offering more guidance and context.  The performance of ChatGPT-based evaluation in segment-level machine translation is poor. To overcome this, Lu et al. \cite{lu2023error} proposed a novel prompting called Error Analysis (EA) prompting, which combines error analysis \cite{lu2022toward} and CoT prompting. The authors showed that with EA prompting, ChatGPT can assess translations at the segment level much better. 

Some of the research works explored GLLMs for the evaluation of multi-modal AI tasks \cite{lu2023llmscore}, fine-tuning open-source LLM evaluators \cite{xu2023instructscore}, and paraphrasing references to enhance existing metrics based on pretrained language models \cite{tang2023not}. For example,  Lu et al. \cite{lu2023llmscore} introduced LLMScore (based on GPT-4), a new metric which can effectively capture both image and object-level compositionality for text-to-image synthesis evaluation. Some of the research works explored these models to fine-tune open-source LLMs so that they can be used as evaluators, which makes the evaluation less expensive. For example, Xu et al. \cite{xu2023instructscore} introduced InstructScore, a novel and explainable metric based on fine-tuned LLaMA model for text generation evaluation. Here the authors use GPT-4 generated synthetic data to fine-tune the LLaMA model. InstructScore can generate an error diagnostic report having error details along with an explanation. Natural language generation evaluation using few references results in poor correlation with human judgements. To overcome this drawback, Tang et al. \cite{tang2023not} introduced Para-Ref, which leverages LLMs to increase the number of references by paraphrasing. The evaluation on three NLG tasks, text summarization, machine translation and image caption, showed that the proposed approach enhances the correlation of sixteen automatic evaluation metrics with human judgements by a good margin. 

Some of the research works focused on addressing the limitations of using GLLMs as evaluators. For example, Wang et al. \cite{wang2023largeeval} demonstrated positional bias in GLLM-based evaluation, i.e., the order of candidate responses can significantly influence the results. The authors demonstrated that the two proposed strategies, namely multiple evidence calibration and balanced position calibration, can reduce the bias and enhance the correlation with human judgements. 

\section{Future Research Directions}
\label{section-12}
\subsection{Enhance Robustness of GLLMs}
GLLMs achieved promising results across various NLP tasks in zero and few-shot settings across various NLP tasks. In some of the tasks like data labelling  \cite{gilardi2023chatgpt, he2023annollm, tornberg2023chatgpt, alizadeh2023open}, text classification \cite{sun2023text}, relation extraction \cite{wan2023gpt}, question answering \cite{yang2022empirical,bang2023multitask}, keyphrase generation \cite{song2023chatgpt}, etc., these models achieved even SOTA results. However, some of the recent research works exposed the brittleness of these models towards out-of-distribution inputs \cite{wang2023robustness, liu2023evaluating}, adversarial prompts \cite{zhu2023promptbench, shirafuji2023exploring, han2023information} and inputs \cite{chen2023robust,zhuo2023robustness,zhao2023robut,liu2023comprehensive} . For example, Liu et al. \cite{liu2023evaluating} reported that ChatGPT and GPT-4 perform well in multiple choice question answering but struggle to answer out-of-distribution questions. Similarly, Chen et al. \cite{chen2023robust} observed more than 35\% performance degradation for GPT-3 and GPT-3.5 models in tasks like sentiment analysis and natural language inference for adversarial inputs. The brittleness towards out-of-distribution and adversarial inputs makes these models unreliable and limits their practical utility, especially in sensitive domains. So, it is necessary for the research community to focus more on this research direction to make GLLMs more robust and enhance their reliability and usage. 

\subsection{Red Teaming} 
Red teaming involves an assessment to expose undesirable model behaviours like generating harmful text \cite{bhardwaj2023red,ganguli2022red,mehrabi2023flirt,perez2022red}. GLLMs trained over large volumes of text data with a simple next-word prediction objective are surprisingly good at generating text with human-like fluency. However, the other side is that these models sometimes generate harmful text. For example, Risabh et al. \cite{bhardwaj2023red} observed that GLLMs like ChatGPT and GPT-4 generate answers to more than 60\% of harmful queries. One of the possible reasons for this undesirable behaviour of GLLMs is that data used for pretraining these models includes toxic, biased and noisy text to some extent \cite{bhardwaj2023red}. This unwanted behaviour of generating harmful text raises concerns and limits the scalable deployment of these models for public use. We can expect more research in future to expose such undesirable behaviour in various scenarios and eventually enhance the safety alignment as well as the safe use of GLLMs.

\subsection{State-Of-The-Art Results Across NLP Tasks} 
In the beginning, GLLMs like GPT-3 achieved impressive performances in zero and few-shot settings across NLP tasks. Advanced GLLMs like ChatGPT and GPT-4 further pushed the results but still lag behind SOTA results achieved by pretrained language models fine-tuned based on supervised learning. Later, with the evolution of advanced prompting strategies and novel approaches, GLLMs are able to achieve SOTA results in some of the NLP tasks. For example, InstructGPT with CARP prompting strategy using just 16 examples achieves SOTA results on four text classification datasets \cite{sun2023text}. Similarly, Wan et al. \cite{wan2023gpt} achieved SOTA results in relation extraction with the novel GPT-RE framework. Yang et al. \cite{yang2022empirical} proposed a novel approach which uses GPT-3 as an implicit knowledge source and achieves SOTA results in knowledge-based visual question answering.   In future, we can expect more focus from the research community to achieve SOTA results using GLLMs in as many NLP tasks as possible, which will be treated as a further push towards artificial general intelligence. Moreover, this eliminates the painful process of labelling large amounts of data and then fine-tuning pretrained language models separately for each downstream task.

\subsection{Robust Approaches to Detect GLLM Generated Text}
The ability to generate text with human-like fluency resulted in the wide adoption of GLLMs in various real-world applications like writing assistants, coding assistants, and chatbots \cite{Mireshghallah2023SmallerLM}. There is a growing concern regarding the misuse of these models for various illegal activities \cite{Guo2023HowCI}, like fake news on social media platforms \cite{hacker2023regulating,de2023chatgpt}, fake reviews on e-commerce websites \cite{Mitrovic2023ChatGPTOH},  fake research papers \cite{gao2023comparing}, academic fraud \cite{cotton2023chatting}, etc. The performance of existing approaches like DetectGPT, ZeroGPT, OpenAI detector, ChatGPT-detector-roberta and ChatGPT-qa-detector-roberta is not satisfactory \cite{pegoraro2023chatgpt,yu2023cheat}. Moreover, the existing approaches are not robust to various attacks like paraphrasing, synonym word replacement and writing style modification \cite{shi2023red, krishna2023paraphrasing}. So, there is a great need for better approaches which can reliably detect GLLM generated text and also robust to various attacks, including paraphrasing. With reliable and robust detection approaches, the misuse of GLLMs for various illegal activities can be reduced to a great extent.

\subsection{Reduce Inference Costs}
GLLMs achieve impressive performances across NLP tasks, with SOTA results in some tasks. However, the downside of using GLLMs is the high inference costs \cite{chen2023frugalgpt,cheng2023batch}. For example, a small business is required to spend more than \$21,000 monthly to use GPT-4 for better customer support \footnote{https://neoteric.eu/blog/how-much-does-it-cost-to-use-gpt-models-gpt-3-pricing-explained}. Such high inference costs have become a burden to small and medium-sized companies. Recently, Chen et al. \cite{chen2023frugalgpt} proposed FrugalGPT, a novel framework involving multiple strategies like prompt adaptation and  LLM approximation to reduce the inference costs of GLLMs. The inference costs of GLLMs increase with the prompt size as the inference cost is computed based on the number of tokens processed. Prompt adaptation focuses on reducing the size of the prompt by using fewer but effective examples or querying the GLLMs as a batch. LLM approximation uses cache to avoid querying GLLM for similar queries,  which eventually reduces overall inference costs. Similarly, Cheng et al. \cite{cheng2023batch} proposed batch prompting, which involves GLLM inference in batches rather than processing one sample individually. The authors demonstrated that the proposed prompting strategy reduces Codex model inference cost across ten datasets with little or no degradation in the performance. Future research in this direction will result in much better approaches which 
 will further reduce the GLLM inference costs and make GLLM usage more affordable for companies. 

\subsection{Enhance Performance in Domain-Specific NLP Tasks}
Inspired by the success of GLLMs in general domain NLP tasks, the research community explored GLLMs for NLP tasks in specific domains like healthcare, legal, finance, etc. However, the performances of GLLMs in domain-specific NLP tasks are not as impressive as those achieved in general domain NLP tasks \cite{moradi2021gpt,hernandez2023we,chalkidis2023chatgpt,choi2023chatgpt,li2023chatgpt, shah2023zero}. For example, Moradi et al. \cite{moradi2021gpt} reported that the BioBERT model outperforms GPT-3 in few-shot settings even though the BioBERT model is 514 times smaller than GPT-3. Chalkidis et al.  \cite{chalkidis2023chatgpt} evaluated ChatGPT on the LexGLUE benchmark and reported that ChatGPT performs poorly on legal text classification datasets. Analyzing domain-specific texts is more challenging because of domain-specific terminology and abbreviations, complex language structures, etc.  In domains like healthcare, finance and legal,  domain experts use many words and abbreviations that are specific to the domain and not commonly found in general domain texts. There is a lot of scope to improve the performance of GLLMs in domain-specific NLP tasks, which reduces the bottleneck for the widespread adoption of these models in specific domains. 

\subsection{Handle Limited Context Length}
One of the major drawbacks of GLLMs is their limited context length \cite{li2023unlocking,kaddour2023challenges,arefeen2023leancontext}. The maximum context length of GLLMs lies in the range of 2049 tokens to 32,768 tokens\footnote{https://platform.openai.com/docs/models/overview}. This limited context length poses a challenge and becomes a bottleneck for GLLMs to handle long documents or maintain long conservations in which the number of tokens falls beyond the maximum context length. Recently, Li \cite{li2023unlocking} proposed selective context, a novel approach to effectively utilize the limited context length by filtering out the less useful content in the input text. The authors demonstrated the effectiveness of the proposed approach using the ChatGPT model for question-answering and text summarization tasks across datasets having lengthy input instances. Future research in this direction will help in the evolution of more efficient approaches which will effectively utilize the limited context length and eliminate the bottlenecks for the application of GLLMs in tasks that require processing long inputs. 

\subsection{Ensure Fair Evaluation of GLLMs}
GLLMs achieved impressive performances across NLP tasks and have received much attention recently. However, one concern regarding the evaluation of GLLMs is data contamination, which refers to the presence of test data instances of downstream tasks in the training corpus of GLLMs \cite{chang2023survey,golchin2023time,aiyappa2023can}. The problem of data contamination is more relevant in the case of GLLMs because of their proprietary nature and non-disclosure of training corpus details. Recent research works have reported the problem of data contamination in GLLMs like ChatGPT  \cite{aiyappa2023can} and GPT-4 \cite{golchin2023time}. For example, Golchin et al. \cite{golchin2023time} demonstrated that GPT-4 is contaminated with instances from text classification, natural language inference and text summarization datasets like WNLI \cite{wang2018glue}, AG News \cite{zhang2015character} and XSUM \cite{narayan2018don}.  Recently, golchin et al. \cite{golchin2023time} proposed a novel approach to detect data contamination for LLMs. Future research must focus on developing simple and effective approaches to identify data contamination and ensure fair evaluation, enhancing the reliability of impressive performances of GLLMs.

\subsection{Reduce Hallucinations}
Despite the remarkable performances of GLLMs, there is a growing concern regarding their tendency to generate factually incorrect information \cite{zhang2023siren,rawte2023survey}. This tendency to generate text that doesn’t align with existing world knowledge, deviates from the user's input or contradicts the context generated earlier is referred to as hallucination \cite{zhang2023siren}. Hallucination is a serious problem yet to be addressed fully \cite{dhuliawala2023chainofverification}, and it reduces the reliability of GLLMs, which becomes a bottleneck for the adoption of GLLMs, especially in sensitive domains like healthcare \cite{umapathi2023med}. Recently, some of the research works focused on evaluating hallucination in GLLMs \cite{umapathi2023med}, assessing the ability of GLLMs to identify hallucinations \cite{li2023halueval} and developing approaches to reduce hallucinations \cite{peng2023check}. For example,  Li et al. \cite{li2023halueval} proposed HaluEval, a novel benchmark to assess the ability of GLLMs to identify hallucinations. Peng et al. \cite{peng2023check} introduced LLM-AUGMENTER, a novel approach that reduces hallucinations in ChatGPT without impacting the quality of generated responses.  Considering the seriousness of the hallucination problem, we can expect more future research to identify and reduce hallucinations in GLLMs, which enhance their reliability and adoption across domains, including sensitive domains like healthcare. 

\subsection{Enhance the Performance of GLLMs for Non-English Languages}
The performance of GLLMs is not impressive in the case of non-English languages, especially in the case of languages with non-Latin scripts \cite{ahuja2023mega, bang2023multitask, lai2023chatgpt, kuzman2023chatgpt}. This is because GLLMs are mostly pretrained on English text. For example, more than 90\% of text in the pretraining corpus of the GPT-3 model is from the English language \cite{brown2020language, ahuja2023mega}. Some of the possible options to enhance the performance of GLLMs for non-English languages are the use of English prompts \cite{lai2023chatgpt, kuzman2023chatgpt} and optimized tokenization \cite{armengol2022multilingual}. There is a great need for better approaches to greatly enhance the performance of GLLMs for non-English languages, which increase their adoption across the globe and benefit users from non-English communities.

\section{Conclusion}
In this survey paper, we provide a comprehensive review of GPT-3 family large language models in multiple dimensions covering more than 350 recent research papers. Here, we present foundation concepts, GPT-3 family large language models and discuss the performances of these models in various downstream tasks, specific domains and multiple languages.  We also discuss data labelling, data augmentation and data generation abilities of GLLMs, the robustness of GLLMs, the effectiveness of GLLMs as evaluators, and finally, conclude with multiple insightful future research directions. Overall, this comprehensive survey paper on GPT-3 family large language models will serve as a good resource for both academic and industry people to stay updated with the latest research. 

% or


%\appendices

%\section{}

\ifCLASSOPTIONcompsoc
  \section*{Acknowledgments}
\else
  \section*{Acknowledgment}
\fi

The author would like to thank Ajit Rajasekharan for his encouragement and support.

\ifCLASSOPTIONcaptionsoff
  \newpage
\fi



%\bibliographystyle{IEEEtran}
\bibliographystyle{IEEEtran}
\bibliography{ammus.bib}
\end{document}